\documentclass{article}

\usepackage{microtype}
\usepackage{graphicx}
\usepackage{subfigure}
\usepackage{booktabs} %
\usepackage{amsmath,amssymb} %
\usepackage{tikz} %
\usetikzlibrary{arrows.meta,positioning,calc,fit,backgrounds,patterns,shapes.geometric,decorations.pathreplacing}
\usepackage[nointegrals]{wasysym}
\usepackage{fontawesome6}
\usepackage{twemojis}
\usepackage{stfloats} %

\usepackage{hyperref}

\newcommand{\method}{FlipQuant}

\usepackage[accepted]{mlsys2025}
\makeatletter
\renewcommand{\Notice@String}{Preprint.}
\renewcommand{\printAffiliationsAndNotice}[1]{%
\stepcounter{@affiliationcounter}%
{\let\thefootnote\relax\footnotetext{\hspace*{-\footnotesep}#1%
\forloop{@affilnum}{1}{\value{@affilnum} < \value{@affiliationcounter}}{
\textsuperscript{\arabic{@affilnum}}\csname @affilname\the@affilnum\endcsname%
}.
\ \\
\Notice@String
}}}
\makeatother

\mlsystitlerunning{Gradient-Guided Hardware-Native MixFP4}

\begin{document}

\twocolumn[
\mlsystitle{Gradient-Guided Hardware-Native MixFP4}

\mlsyssetsymbol{equal}{*}

\begin{mlsysauthorlist}
\mlsysauthor{Jian-Jia Chen}{equal,nycu}
\mlsysauthor{Brian J.~Chan}{equal,nycu}
\mlsysauthor{Wen-Hung Lee}{equal,nycu}
\mlsysauthor{Yi-Ting Li}{ntu}
\mlsysauthor{Po-Shen Chen}{nycu}
\mlsysauthor{Kuan-Te Kuo}{nycu}
\mlsysauthor{Hsien-Cheng Huang}{nycu}
\mlsysauthor{Cody Hao Yu}{oai}
\mlsysauthor{Chi-Chih Chang}{cornell}
\mlsysauthor{Mohamed S.~Abdelfattah}{cornell}
\mlsysauthor{Kai-Chiang Wu}{nycu}
\end{mlsysauthorlist}

\mlsysaffiliation{nycu}{National Yang Ming Chiao Tung University, Hsinchu, Taiwan}
\mlsysaffiliation{ntu}{National Taiwan University, Taipei, Taiwan}
\mlsysaffiliation{oai}{OpenAI, San Francisco, CA, USA}
\mlsysaffiliation{cornell}{Cornell University, USA}

\mlsyskeywords{Quantization, FP4, NVFP4, Mixed-Format Quantization, Blackwell, Tensor Cores}

\vskip 0.3in

\begin{abstract}
Blackwell tensor cores support native inference with 4-bit weights and
activations (W4A4), but NVFP4 leaves a perplexity (PPL) gap to
BF16. Choosing a format for each 16-element block can improve accuracy, but
current matrix multiply instructions cannot execute choices this fine.
Applying these rules at hardware-compatible tiles loses most of their
benefit. We introduce
\method{}, which selects E2M1 or E0M3 for each MMA weight tile. It keeps both
quantized weight candidates fixed and learns a binary format map from
end-to-end gradients of a bucketed KL loss to a BF16 teacher, re-evaluating
them after each update. Native kernels execute $16 \times 64$ tiles on SM120
and $256 \times 64$ tiles on SM100 with median GEMM overheads of 1.9\% and
1.7\%, respectively. Across six language models and two corpora, \method{}
at $16 \times 64$ closes 42.5\% of NVFP4's PPL gap to BF16 on average,
compared with 13.3\% for FourOverSix. Code is available at
\url{https://github.com/brian030128/flipquant}.
\end{abstract}
]

\printAffiliationsAndNotice{\mlsysEqualContribution} %

\providecommand{\pend}[1]{\textcolor{red}{#1}} %

\section{Introduction}
\label{sec:Introduction}

NVIDIA's Blackwell GPUs multiply 4-bit floating-point (FP4) operands natively
in their tensor cores, so a linear layer with 4-bit weights and activations
(W4A4) runs as a single block-scaled GEMM in the NVFP4
format~\cite{nvfp4blog}. Its E2M1 grid fits blocks with a few
large values better than evenly spread ones, and W4A4 NVFP4 models lag BF16
even with improved scaling~\cite{fouroversix}.

\begin{figure}[t]
  \centering
  \includegraphics[width=\columnwidth]{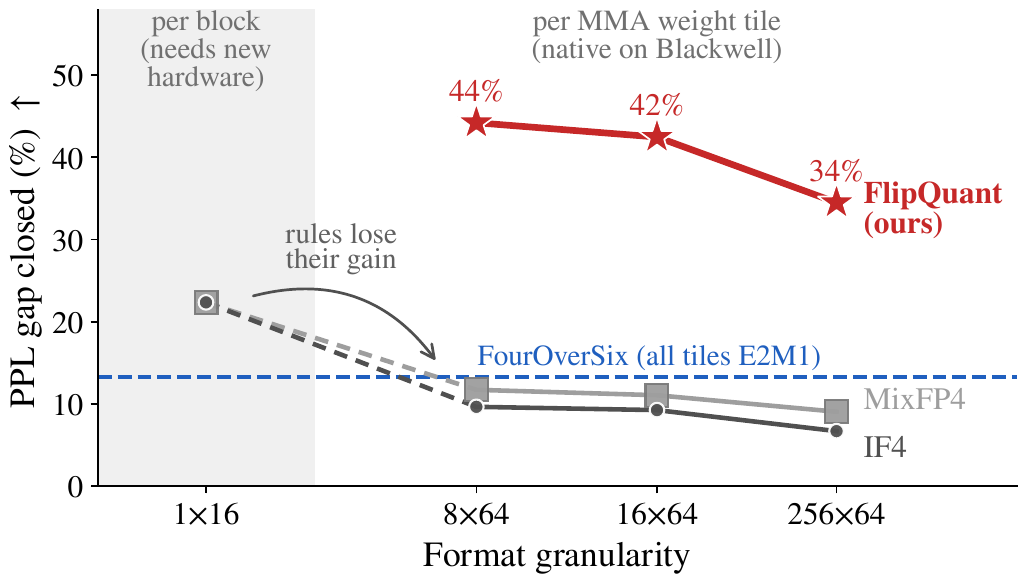}
  \caption{Format selection at hardware granularity: the share of NVFP4's
           PPL gap to BF16 closed (higher is better), averaged over WikiText-2, C4 and the six
           models of Table~\ref{tab:main-ppl}. The per-block rules IF4 and
           MixFP4 need new hardware at $1 \times 16$ and fall below FourOverSix
           per MMA weight tile, the unit Blackwell tensor cores execute;
           \method{} selects formats per tile and runs natively
           (Appendix~\ref{app:coarse}).}
  \label{fig:teaser}
\end{figure}

Recent mixed-format methods improve accuracy by choosing a 4-bit grid for
each small block~\cite{if4,mixfp4arxiv,blockdialect}. Their fine-grained
choices require modified hardware for native execution, since existing
MMA instructions assign one format to an entire operand.

Blackwell's tensor cores also support an undocumented uniform FP4 format,
E0M3~\cite{qianqian2026int4,revollllt2026e0m3}, with the same MMA throughput
as E2M1. This enables native mixed-format
execution on existing hardware, but only per instruction operand: the
finest region that can own a format is a weight tile of
$8 \times 64$ to $256 \times 64$ elements. At this granularity, per-block
rules break down (Figure~\ref{fig:teaser};
Section~\ref{sec:background:gap}). The blocks of a tile
disagree: on Mistral-7B, more than 99.999\% of the tiles contain blocks that
prefer each format. Summing their reconstruction errors weighs each block by
how much its weights move, not by how much the move matters to the output,
and the per-block rules applied at hardware tiles keep only 23--46\% of their
gain over FourOverSix on Mistral-7B; averaged over six models, they fall below it. 
Choosing a format per tile needs a signal that weighs
every block by its effect on the output.

We present \method{}, which selects the format of every hardware tile by the
end-to-end loss and does not change the weights. One backward pass gives, for
every tile, the gradient of the KL divergence to the BF16 model with respect
to the tile's format, a first-order estimate of what flipping the tile from
E2M1 to E0M3 would change. Because flips interact, this estimate holds only
for few tiles, so \method{} does not select tiles once by it: it learns a
binary flip map with a straight-through estimator and re-evaluates the
gradients after every update.

Our contributions are:
\begingroup\setlength{\leftmargini}{1.2em}
\begin{itemize}\setlength{\itemsep}{1pt}\setlength{\parsep}{0pt}
  \item \textbf{Hardware-native format tiles.} We identify the operand tile of
        Blackwell's block-scaled MMAs as the unit of native format selection,
        and show that per-block rules lose most of their gain at it
        (Section~\ref{sec:background}).
  \item \textbf{Gradient-guided format selection.}
        We introduce \method{}, which iteratively optimizes a binary
        E2M1/E0M3 format map using end-to-end distillation gradients while
        keeping the original weights and quantization candidates fixed.
        Each backward pass computes STE-based selection signals for all
        tiles without evaluating flips individually
        (Sections~\ref{sec:method} and~\ref{sec:eval:analysis}).

  \item \textbf{Efficient mixed-format kernels.}
        We identify and address distinct execution bottlenecks:
        duplicated predicated MMAs on SM120 through per-K-tile dispatch,
        and reduced arithmetic intensity from narrow MMA operands on
        SM100 through $256 \times 64$ format tiles.
        Our kernels incur median GEMM overheads of 1.9\% and 1.7\%,
        respectively. Across six models, \method{} at $16 \times 64$
        granularity closes 42.5\% of NVFP4's PPL gap to BF16,
        compared with 13.3\% for FourOverSix (Sections~\ref{sec:execution} and~\ref{sec:eval}).

\end{itemize}
\endgroup

\section{Background and Motivation}
\label{sec:background}

\begin{table*}[t]
  \caption{Block-scaled 4-bit quantization methods, by what they adapt
           (NVFP4 variants; BlockDialect builds on 32-element MX blocks). The
           granularity is the region that shares one adapted choice. Among
           the methods that adapt the element format, only \method{} runs
           natively on existing tensor cores and uses the end-to-end loss as
           its signal. }
  \label{tab:related}
  \vskip 0.1in
  \begin{center}
  \begin{footnotesize}
  \setlength{\tabcolsep}{3pt}
  \begin{tabular}{@{}llllll@{}}
    \toprule
    Method & Adapts & Choice & \begin{tabular}[b]{@{}l@{}}Granularity\end{tabular} & Signal & \begin{tabular}[b]{@{}l@{}}Tensor core support\end{tabular} \\
    \midrule
    FourOverSix~\cite{fouroversix}   & scale    & max.\ mapped to 4 or 6 & 16 elem. & local error      & native \\
    RaZeR~\cite{razer}               & codebook & $-0$ code as block value & 16 elem. & local error & emulated (two-pass GEMM) \\
    IF4~\cite{if4}                   & format   & E2M1 or INT4                   & 16 elem. & local error      & new MAC unit needed \\
    MixFP4~\cite{mixfp4arxiv}        & format   & E2M1 or E1M2                   & 16 elem. & local error      & modified core needed \\
    BlockDialect~\cite{blockdialect} & format   & one of 16 FP4 variants         & 32 elem. & local error & new MAC unit needed \\
    \textbf{\method{} (ours)}        & format   & E2M1 or E0M3                   & $r \times 64$ tile & end-to-end loss & native \\
    \bottomrule
  \end{tabular}
  \end{footnotesize}
  \end{center}
  \vskip -0.1in
\end{table*}

\subsection{Hardware-Native Block-Scaled FP4}
\label{sec:background:fp4}

Block-scaled formats split a tensor into small blocks that share one scale
and store each element as a low-bit code. The OCP microscaling (MX)
formats~\cite{mx} use 32-element blocks with a power-of-two scale.
NVFP4~\cite{nvfp4blog} uses 16-element blocks with an FP8 (E4M3) scale and
one FP32 scale per tensor; each element is a 4-bit E2M1 code with magnitudes
$\{0, 0.5, 1, 1.5, 2, 3, 4, 6\}$. Blackwell tensor cores multiply such
operands directly: a block-scaled MMA reads the codes and block scales of
both operands and accumulates in FP32, so a W4A4 linear layer runs as one
GEMM without dequantization.

Several methods target specific weaknesses of NVFP4. Its E2M1 grid jumps
from 4 to 6, so values near a block's maximum round coarsely;
FourOverSix~\cite{fouroversix} maps a block's maximum to 4 instead of 6 when
this lowers the block's error, changing only the scale's value. We build on
it and use it as our baseline. Its 16 codes also spend one on a redundant
negative zero; RaZeR~\cite{razer} repurposes this code as a per-block
special value indexed by spare bits of the block scale, for which the authors describe a two-pass NVFP4 GEMM implementation on current GPUs. Other methods change the values being
quantized: MR-GPTQ~\cite{mrgptq} combines GPTQ-style error
compensation~\cite{gptq} with block-wise Hadamard rotations,
SOAR~\cite{soar} optimizes the scales analytically, FOCUS~\cite{focus}
learns them end to end, and ARCQuant~\cite{arcquant} adds residual channels.
\method{} changes only how each tile's codes are interpreted, so it composes
with these methods (Section~\ref{sec:eval:analysis}).

\subsection{Fine-Grained Format Adaptivity}
\label{sec:background:adaptive}

Which 4-bit grid fits best depends on the values in a block: a log-spaced
grid such as E2M1 suits a block with a few large values and many small ones,
whereas a uniform grid such as INT4 suits evenly spread values.
LLM-FP4~\cite{llmfp4} searched the FP4 format and exponent bias per tensor;
recent formats choose the element format per block. IF4~\cite{if4} quantizes each
16-element block both as E2M1 and as INT4, keeps the version with the smaller
error, and records the choice in the unused sign bit of the block scale.
MixFP4~\cite{mixfp4arxiv} chooses between E2M1 and an integer-like format in
the same way and decodes both in a modified tensor core, and
BlockDialect~\cite{blockdialect} picks one of 16 FP4 variants for each
32-element MX block. Per-block choice reduces quantization error
substantially, but it assumes that the format can change every 16 or 32
elements, which today's MMA instructions do not allow: these schemes need new
arithmetic units or a software decode before a higher-precision GEMM.
Table~\ref{tab:related} summarizes how they differ from \method{}.

\subsection{Format Adaptivity under Hardware Constraints}
\label{sec:background:gap}

\paragraph{Format tiles of a block-scaled MMA.}
Because our goal is format adaptivity that runs on existing hardware, formats
must follow the granularity at which the hardware can change them. We target
both Blackwell variants: data-center SM100 (B200, GB200) and workstation SM120
(RTX PRO 6000, RTX 5090). On both, a block-scaled FP4 MMA instruction reads 64 consecutive
reduction elements of each operand---four 16-element scale blocks---and
declares one element format per operand. NVFP4 uses E2M1, but the
instructions also accept E0M3, a uniformly spaced grid with magnitudes
$\{0, 1, 2, 3, 4, 5, 6, 7\}$~\cite{qianqian2026int4,revollllt2026e0m3}. The format can change between consecutive MMAs
but not inside one. Since \method{} varies only the format of the weights, the smallest region
that can own a format is the weight operand of one MMA: 64 elements along $K$
by the instruction's extent along the output channels. On SM120,
\texttt{mma.sync} with shape \texttt{m16n8k64} gives $16 \times 64$ when the
weights are its $A$ operand and $8 \times 64$ when they are its $B$ operand,
so 64 or 32 scale blocks share one decision. On SM100, \texttt{tcgen05.mma}
accepts weight operands from 8 to 256 output channels wide, so tiles from
$8 \times 64$ to $256 \times 64$ are executable. Coarser tiles are unions of
these and are equally executable.

\paragraph{Fine-grained preferences meet coarse tiles.}
Coarsening changes the selection problem, not just its granularity. The
natural extension of per-block selection---electing each tile's format by
the reconstruction error summed over its blocks---weighs the blocks of a tile
by how much each weight moves, not by how much the move matters to the
output: it ignores the statistics of each layer's inputs and the sensitivity
of every later layer. This matters because fine-grained preferences are
genuinely mixed: when the per-block choices of the rules below are
grouped into these tiles on Mistral-7B,
more than 99.999\% of the tiles contain both formats, and the minority format
holds 37--38\% of a tile's blocks. A tile-level decision must therefore balance improvements 
in some blocks against degradations in others, and a lower aggregate error does not necessarily 
translate into better output quality.

Restricted to hardware tiles, per-block rules lose their advantage
(Figure~\ref{fig:teaser}; Appendix~\ref{app:coarse}). We run IF4~\cite{if4} and
MixFP4~\cite{mixfp4arxiv} as published: each picks E2M1 or an integer-like
format (INT4 or E1M2) for every 16-element block of the weights and the
activations by reconstruction error. We extend them to hardware tiles by
applying their weight rule per tile; their activation formats stay per block,
which favors them. On Mistral-7B (WikiText-2), both rules beat FourOverSix
at $1 \times 16$, but at $8 \times 64$, $16 \times 64$ and $256 \times 64$
both degrade significantly and keep only 23--46\% of their gain over
FourOverSix; averaged over six models, they fall below FourOverSix at every
tile. \method{}, which
changes the format of 4.1--14.8\% of the tiles, has lower perplexity than
both rules at every tile, and at $8 \times 64$ and $16 \times 64$ lower
than both rules at $1 \times 16$.

An end-to-end criterion is therefore needed. First-order estimates of the
end-to-end loss are available for every tile from one backward pass, but each
estimate assumes that all other tiles stay fixed, whereas flips interact
through shared inputs and later layers. The question we answer is how to turn
fine-grained, end-to-end evidence into coarse, hardware-executable decisions
that account for these interactions; Section~\ref{sec:eval:analysis} compares
our answer with selecting tiles once by their first-order estimates.

\section{Gradient-Guided Coarse Format Selection}
\label{sec:method}

\begin{figure*}[t]
\centering
    \includegraphics[width=0.85\textwidth]{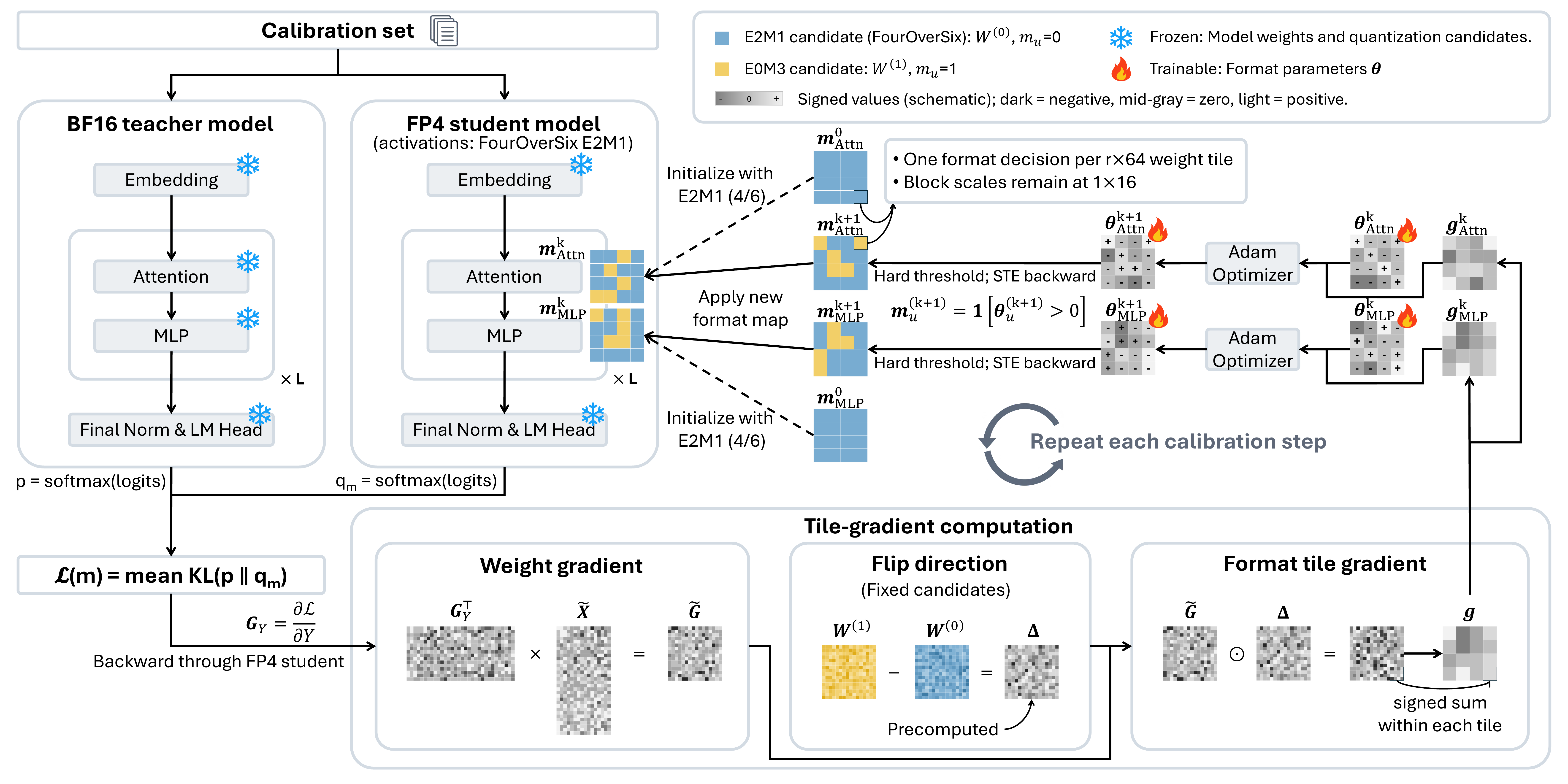}
    \caption{Overview of \method{}'s calibration. A binary map selects the
        fixed FourOverSix E2M1 or E0M3 candidate of each $r \times 64$ tile;
        scale blocks remain $1 \times 16$. Tile gradients, the sums of
        $\tilde G \odot \Delta$ over each tile, come from the bucketed KL
        divergence to a frozen BF16 teacher (Eq.~\eqref{eq:objective}). Adam
        updates only the latent parameters $\theta$, whose hard threshold,
        with a straight-through estimator, gives the next map. $k$: calibration
        step; Attn, MLP: representative linear layers.}
\label{fig:overview}
\end{figure*}

Section~\ref{sec:background} established that the hardware fixes the smallest
unit at which a weight's element format can change---one MMA weight-operand
tile---but leaves open \emph{which} format each tile should take. This section
describes how \method{} casts that choice as an optimization of a binary flip
map against the end-to-end bucketed KL divergence to the BF16 model.
We define the two candidate formats, the flip map, and the objective
(\S\ref{sec:method:formulation}), derive for every tile the gradient of the
end-to-end loss with respect to its map entry, a first-order estimate of how
the loss changes when the tile switches format
(\S\ref{sec:method:gradient}), and learn the flip map from these tile
gradients, re-evaluating them after every update
(\S\ref{sec:method:training}).
Section~\ref{sec:execution} executes the learned
map with native MMA instructions. Figure~\ref{fig:overview} gives an
overview of the calibration.

\subsection{Problem Formulation}
\label{sec:method:formulation}

\paragraph{Two fixed candidates per weight.}
Consider a linear layer $Y = XW^{\top}$, where $X \in \mathbb{R}^{n \times K}$
holds the activations of $n$ tokens and $W \in \mathbb{R}^{N \times K}$ is
the weight, with $N$ output channels and reduction dimension $K$.
We divide $W$ by one FP32 scale per tensor,
$\alpha = \max|W| / (6 \cdot 448)$, as in the standard NVFP4
recipe~\cite{nvfp4blog}. We split each row into
16-element blocks along $K$ and let $v$ denote the scaled values of one
block. Before any selection, we quantize $W$ in two ways, where $Q(\cdot)$
returns the dequantized weights:
\begin{itemize}
  \item $W^{(0)} = Q_{\mathrm{E2M1}}(W)$ is NVFP4 with FourOverSix block
        scaling~\cite{fouroversix}. The tensor scale stays $\alpha$, and each
        block chooses its scale $\max|v|/6$ or $\max|v|/4$, i.e., maps its
        maximum magnitude to 6 or 4 on the E2M1 grid
        $\{0, \pm0.5, \pm1, \pm1.5, \pm2, \pm3, \pm4, \pm6\}$, whichever
        gives the smaller squared error.
  \item $W^{(1)} = Q_{\mathrm{E0M3}}(W)$ uses the uniform grid
        $\{0, \pm1, \dots, \pm7\}$, similar to symmetric INT4, with block
        scale $\max|v|/7$.
\end{itemize}
The superscript is the value of the flip map, defined below, that selects
each candidate. Every block scale is clipped to $[2^{-9}, 448]$ and rounded
to UE4M3, the unsigned 8-bit scale format of NVFP4. Both candidates share
the tensor scale $\alpha$. E2M1 and E0M3 are the two element formats that
block-scaled FP4 MMAs can select per operand. Both encode 15 distinct values in 4-bit
codes and use the same scale formats, so the two candidates share one
storage layout and differ only in the values of their codes and block
scales; a per-tile tag determines how a tile's codes are decoded
(Section~\ref{sec:execution}).

Activations are quantized to E2M1 with FourOverSix scaling and 16-element
blocks. During calibration, each row of $X$ (one token) has its own FP32
scale, so no token's quantization depends on later tokens.

The candidates are computed once and stay fixed. The optimization never
changes the model weights or the candidates' codes and scales. It only
chooses which candidate each tile uses. 

\paragraph{Format tiles and the flip map.}
We divide the weight of every quantized linear layer into \emph{format
tiles} of $r \times 64$ elements: $r$ consecutive rows (output channels) by
64 consecutive columns (reduction elements), the executable regions of
Section~\ref{sec:background:gap}. All elements of a tile share one element
format. This paper uses $r = 8$, $16$, and $256$.

For layer $\ell$ with weight shape $N \times K$, let $\mathcal{U}_\ell$ be
its set of $\lceil N/r \rceil \cdot K/64$ tiles, and let
$P_u \in \{0,1\}^{N \times K}$ be the mask of tile $u \in \mathcal{U}_\ell$,
equal to 1 inside the tile and 0 elsewhere. The tiles of a layer do not
overlap and together cover its weight; when $r$ does not divide $N$, the
last row of tiles keeps only the remaining rows. We collect the tiles of all
layers in $\mathcal{U} = \bigcup_\ell \mathcal{U}_\ell$. The
\emph{flip map} $m \in \{0,1\}^{|\mathcal{U}|}$ assigns one bit to every
tile in the model. $m_u = 0$ keeps tile $u$ in E2M1, and $m_u = 1$ flips it
to E0M3. The effective weight of layer $\ell$ is
\begin{equation}
  \begin{aligned}
    \hat W_\ell(m) &= W^{(0)}_\ell
      + \sum_{u \in \mathcal{U}_\ell} m_u \, P_u \odot \Delta_\ell, \\
    \Delta_\ell &= W^{(1)}_\ell - W^{(0)}_\ell .
  \end{aligned}
  \label{eq:map}
\end{equation}
Inside a flipped tile, the term $P_u \odot \Delta_\ell$ turns
$W^{(0)}_\ell$ into $W^{(1)}_\ell$. All other elements keep their
$W^{(0)}_\ell$ values. Every element of $\hat W_\ell(m)$ therefore equals
the corresponding element of one of the two candidates, and no map
interpolates between them. Layer $\ell$ then computes
$Y = Q_a(X)\,\hat W_\ell(m)^{\top}$, where $Q_a$ is the activation quantizer
described above. We omit the index $\ell$ when a statement concerns a
single layer.

\paragraph{Objective.}
We choose $m$ so that the quantized model's output distribution stays close
to that of the BF16 teacher, as in knowledge
distillation~\cite{hinton2015kd}. Let $\mathcal{D}$ be a calibration set of
sequences of $T$ tokens (Section~\ref{sec:eval:setup}). For sequence $s$ and
prediction position $t$, we retain the teacher's top-1{,}000 tokens and merge
the remaining vocabulary into one tail bucket. Let $\bar p_{s,t}$ and
$\bar q_{m,s,t}$ denote the teacher and student distributions over these same
buckets, preserving their full-vocabulary probability masses. We minimize
\begin{equation}
  \mathcal{L}(m) = \frac{1}{|\mathcal{D}|} \sum_{s \in \mathcal{D}}
  \frac{1}{T-1} \sum_{t=1}^{T-1}
  \mathrm{KL}\!\left(\bar p_{s,t} \,\|\, \bar q_{m,s,t}\right).
  \label{eq:objective}
\end{equation}
Direct optimization is combinatorial: one bit per tile gives
$2^{|\mathcal{U}|}$ possible maps. A local proxy such as per-tile weight
MSE avoids this search because it scores each tile independently.
However, it ignores the layer's inputs and how later layers respond to the
layer's output (Section~\ref{sec:background:gap}). We therefore derive the
selection signal from Eq.~\eqref{eq:objective} itself
(\S\ref{sec:method:gradient}).

\subsection{Tile Gradients of the End-to-End Loss}
\label{sec:method:gradient}

For every tile, we compute the gradient of the loss in
Eq.~\eqref{eq:objective} with respect to its map entry, which estimates to
first order how the loss changes when the tile switches format. For this
derivative only, we treat $m_u$ as a real variable; the forward pass always
uses the hard map. By the chain rule,
\begin{equation}
g_u = \frac{\partial \mathcal{L}}{\partial m_u}
= \Big\langle \frac{\partial \mathcal{L}}{\partial \hat W},
\frac{\partial \hat W}{\partial m_u} \Big\rangle
= \sum_{(i,j)\in u}\tilde G_{ij}\Delta_{ij}.
\label{eq:tile-grad}
\end{equation}
The first factor, the weight gradient $\tilde G$, requires backpropagation
through quantized activations. The second, the flip direction
$P_u \odot \Delta$, follows directly from Eq.~\eqref{eq:map}. We describe the two
factors in turn.

\paragraph{The weight gradient.}
Each quantized layer computes $Y = Q_a(X)\,\hat W(m)^{\top}$, so its weight
gradient is $G_Y^{\top} Q_a(X)$, where $G_Y = \partial\mathcal{L}/\partial Y$
is the gradient of the loss with respect to the layer's output. To reach the
layer, $G_Y$ must pass backward through the activation quantizers of later
layers. These quantizers include non-differentiable rounding operations,
which would block the gradient. We therefore use an identity
straight-through estimator (STE)~\cite{bengio2013ste} in their backward pass:
\begin{equation}
  \tilde X = \mathrm{sg}\bigl(Q_a(X)\bigr) + X - \mathrm{sg}(X),
  \label{eq:act-ste}
\end{equation}
where $\mathrm{sg}$ is the stop-gradient operator. The forward value of
$\tilde X$ is $Q_a(X)$, and the backward pass treats
$\partial \tilde X / \partial X$ as the identity. With $\tilde X$ in place of
$Q_a(X)$ in every quantized layer, the first factor of
Eq.~\eqref{eq:tile-grad} is
\begin{equation}
  \tilde G = G_Y^{\top} \tilde X .
  \label{eq:weight-grad}
\end{equation}
We differentiate with respect to the effective weight $\hat W$, not the
original weight $W$. The weight quantizer, which is applied once before the
optimization, is therefore never differentiated. Because of the STE,
$\tilde G$ is an STE-based gradient rather than an exact gradient of the
quantized model.

\paragraph{The flip direction.}
By Eq.~\eqref{eq:map}, $\hat W(m)$ is linear in $m$, so the second factor
is exact: $\partial \hat W / \partial m_u = P_u \odot \Delta$. It is the
change in the weight when tile $u$ moves from E2M1 toward E0M3, and it is
zero outside the tile. The inner product in Eq.~\eqref{eq:tile-grad}
therefore reduces to a sum over the elements of tile $u$. A flip changes
$m_u$ by $+1$ for an E2M1 tile and by $-1$ for an E0M3 tile. To first
order, flipping tile $u$ thus changes the loss by $g_u$ if the tile is in
E2M1 and by $-g_u$ if it is in E0M3. An E2M1 tile with $g_u < 0$, or an
E0M3 tile with $g_u > 0$, is therefore predicted to lower the loss when
flipped. This prediction is STE-based and first-order, while a flip is a
discrete step. We therefore use $g_u$ as an optimization signal rather than
as a one-shot flip criterion (\S\ref{sec:method:training}).

\paragraph{Comparison with weight-space rules.}
The coarsened rules of Section~\ref{sec:background:gap} also score a tile by
summing over its elements, but they sum changes in weight error. For squared
error, the score is
\begin{equation}
  s_u = \sum_{(i,j)\in u}
  \Bigl[\bigl(W^{(1)}_{ij}-W_{ij}\bigr)^2 - \bigl(W^{(0)}_{ij}-W_{ij}\bigr)^2\Bigr].
  \label{eq:mse-score}
\end{equation}
Every element counts equally in $s_u$, whatever input it meets and however
much it affects the output. In Eq.~\eqref{eq:tile-grad}, each element
instead contributes $\tilde G_{ij}\Delta_{ij}$. Through $\tilde X$, the
factor $\tilde G = G_Y^{\top}\tilde X$ weights the element by the activations
of its reduction column. Through $G_Y$, it weights the element by the effect
of its output channel on every later layer and on the logits. This
difference matters because format preferences are mixed within a tile.
Section~\ref{sec:background:gap} shows that block-level preferences disagree
within virtually every tile at hardware granularity. In $g_u$, elements that
favor different formats are weighed by how much each moves the loss, not by
how much each moves the weight.

\subsection{Learning the Flip Map from Tile Gradients}
\label{sec:method:training}

We learn the flip map from the tile gradients of
\S\ref{sec:method:gradient} through a real latent parameter per tile, and
recompute $g_u$ at the current hard map after every update
(Algorithm~\ref{alg:train-map}), so each gradient reflects the flips made by
earlier updates. Unlike AdaRound~\cite{adaround} and BRECQ~\cite{brecq},
which learn each weight's rounding direction through a soft relaxation of a
layer- or block-wise reconstruction loss, \method{} learns one hard format
per hardware tile against the end-to-end loss.

\paragraph{Latent parameters.}
Each tile $u$ carries a real latent parameter $\theta_u$ that determines its
format:
\begin{equation}
  m_u = \mathbf{1}[\theta_u > 0] \;\;\text{(forward)}, \qquad
  \frac{\partial m_u}{\partial \theta_u} := 1 \;\;\text{(backward)}.
  \label{eq:ste}
\end{equation}
The forward pass uses the hard map, and the backward pass passes the gradient
straight through the threshold, as in BinaryConnect~\cite{binaryconnect} but
without clipping $\theta_u$. This is a second straight-through estimator. It
acts on the map, whereas the one in Eq.~\eqref{eq:act-ste} acts on the
activations. With it, the gradient that reaches each latent parameter is the
tile gradient at the current map:
\begin{equation}
  \frac{\partial \mathcal{L}}{\partial \theta_u}
  = \frac{\partial \mathcal{L}}{\partial m_u}\,
    \frac{\partial m_u}{\partial \theta_u}
  = g_u .
  \label{eq:latent-grad}
\end{equation}

\paragraph{Initialization and updates.}
We initialize $\theta_u = -1$ for every tile. The initial map is therefore
all E2M1, which is the FourOverSix model. All latent parameters are updated
jointly with Adam~\cite{adam}, and Section~\ref{sec:eval:setup} lists the
hyperparameters. Several $\theta_u$ may cross zero in the same step, so one
update can flip several tiles. A tile can also flip back later, because
$\theta_u$ stays trainable after it crosses zero. After the last update, we
use the hard map directly, without early stopping or checkpoint selection.

\paragraph{Cost and memory.}
A naive way to measure each tile's effect is to flip the tile and run an
extra forward pass, which takes 6.8 million forward passes per step for
Mistral-7B at $16 \times 64$. Instead, one forward and one backward pass
over a batch yield $g_u$ for every tile at once. Beyond backpropagating to
the activations, each layer adds only one GEMM to form $\tilde G$ by
Eq.~\eqref{eq:weight-grad} and a reduction of $\tilde G \odot \Delta$ over
its tiles, so re-evaluating the tile gradients after every update is
affordable. Only the Adam update grows with the number of tiles, and it is
small next to the GEMMs, so finer tiles add almost no cost.
The map adds little memory either. The weights receive no gradients or
optimizer state: only $\theta$ is trained, so the optimizer keeps a few
floats per tile, and each layer's $\tilde G$ is discarded after its
reduction. Memory is instead dominated by the two candidates (stored in packed 4-bit
form, in place of the BF16 weights), the modules that stay in BF16, and the
activations of the forward and backward passes. The teacher's top-1{,}000 log-probabilities
take about 3~GiB for our calibration set and can stay on the GPU or in host memory. Section~\ref{sec:eval:cost} reports the measured cost.

\begin{algorithm}[t]
  \caption{Learning the flip map from tile gradients.}
  \label{alg:train-map}
  \begin{algorithmic}
    \STATE {\bfseries Input:} BF16 model, calibration set $\mathcal{D}$,
           tile shape $r \times 64$, epochs $E$, step size $\eta$
    \STATE {\bfseries Output:} flip map $m$
    \STATE For every layer: $W^{(0)} \gets Q_{\mathrm{E2M1}}(W)$,
           $W^{(1)} \gets Q_{\mathrm{E0M3}}(W)$,
           $\Delta \gets W^{(1)} - W^{(0)}$
    \STATE Cache the teacher's top-1{,}000 token indices and
       log-probabilities on $\mathcal{D}$
    \STATE $\theta_u \gets -1$ for every tile $u$ \COMMENT{all E2M1}
    \FOR{epoch $= 1$ {\bfseries to} $E$}
      \FOR{each batch $\mathcal{B}$ of $\mathcal{D}$, in shuffled order}
        \STATE $m_u \gets \mathbf{1}[\theta_u > 0]$ for every tile $u$
        \STATE Forward $\mathcal{B}$ with weights $\hat W(m)$
               (Eq.~\eqref{eq:map}) and STE activations
               (Eq.~\eqref{eq:act-ste})
        \STATE $\mathcal{L}_{\mathcal{B}} \gets$ mean of
               $\mathrm{KL}(\bar p \,\|\, \bar q_m)$ over $\mathcal{B}$
               (Eq.~\eqref{eq:objective})
        \STATE Backpropagate $\mathcal{L}_{\mathcal{B}}$;
               $g_u \gets \sum_{(i,j)\in u} \tilde G_{ij}\,\Delta_{ij}$
               for every tile $u$ (Eq.~\eqref{eq:tile-grad})
        \STATE Update $\theta$ with Adam on $g$, step size $\eta$
               (Eq.~\eqref{eq:latent-grad})
      \ENDFOR
    \ENDFOR
    \STATE {\bfseries return} $m_u = \mathbf{1}[\theta_u > 0]$
  \end{algorithmic}
\end{algorithm}

\newcommand{\figpro}{\faThumbsUp[regular]}
\newcommand{\figcon}{\faThumbsDown[regular]}

\section{Hardware-Native Mixed-FP4 Execution}
\label{sec:execution}
\label{sec:execution:mapping}\label{sec:execution:packing}

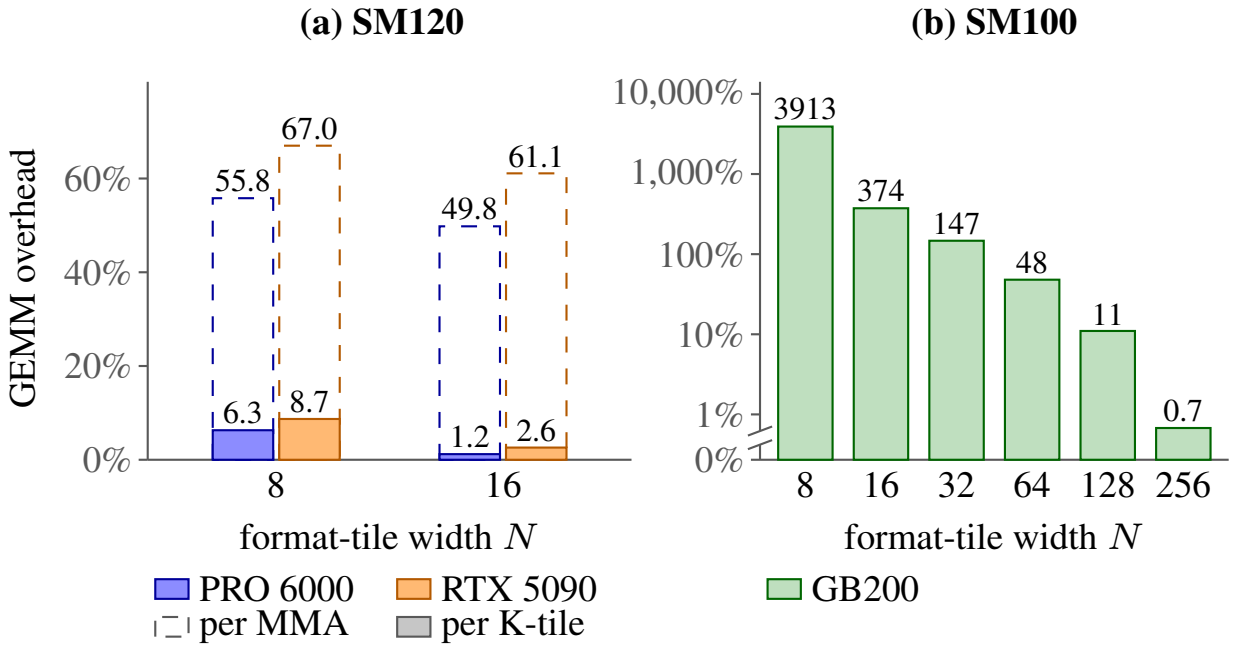
\begin{figure}[h]
\centering
\begin{tikzpicture}[
    font=\scriptsize,
    axis/.style={thin, gray!70!black},
    tick/.style={gray!70!black},
    pro/.style={fill=blue!45, draw=blue!60!black, thin},
    rtx/.style={fill=orange!55, draw=orange!70!black, thin},
    probefore/.style={draw=blue!60!black, thin, dashed},
    rtxbefore/.style={draw=orange!70!black, thin, dashed},
    meas/.style={fill=green!50!black!25, draw=green!40!black, thin}, %
    val/.style={font=\tiny, anchor=south, inner sep=1pt},
  ]

  \node[font=\scriptsize\bfseries, anchor=south] at (1.55, 2.65) {(a) SM120};
  \draw[axis] (0, 0) -- (3.2, 0);
  \draw[axis] (0, 0) -- (0, 2.5);
  \foreach \t in {0, 20, 40, 60} {
    \draw[tick] (0, {0.031*\t}) -- (-0.06, {0.031*\t}) node[left, inner sep=1pt] {\t\%};
  }
  \node[rotate=90, anchor=south] at (-0.62, 1.2) {GEMM overhead};
  \foreach \x/\pb/\pa/\rb/\ra in {0.85/55.8/6.3/67.0/8.7, 2.35/49.8/1.2/61.1/2.6} {
    \draw[probefore] ({\x-0.42}, 0) rectangle ({\x-0.02}, {0.031*\pb});
    \draw[pro]       ({\x-0.42}, 0) rectangle ({\x-0.02}, {0.031*\pa});
    \node[val] at ({\x-0.22}, {0.031*\pb}) {\pb};
    \node[val] at ({\x-0.22}, {0.031*\pa}) {\pa};
    \draw[rtxbefore] ({\x+0.02}, 0) rectangle ({\x+0.42}, {0.031*\rb});
    \draw[rtx]       ({\x+0.02}, 0) rectangle ({\x+0.42}, {0.031*\ra});
    \node[val] at ({\x+0.22}, {0.031*\rb}) {\rb};
    \node[val] at ({\x+0.22}, {0.031*\ra}) {\ra};
  }
  \node[anchor=north, inner sep=2pt] at (0.85, 0) {8};
  \node[anchor=north, inner sep=2pt] at (2.35, 0) {16};
  \node[anchor=north] at (1.6, -0.3) {format-tile width $N$};
  \draw[pro] (0.05, -0.92) rectangle ++(0.22, 0.13);
  \node[anchor=west, inner sep=1pt] at (0.3, -0.855) {PRO 6000};
  \draw[rtx] (1.65, -0.92) rectangle ++(0.22, 0.13);
  \node[anchor=west, inner sep=1pt] at (1.9, -0.855) {RTX 5090};
  \draw[gray!70!black, thin, densely dashed] (0.05, -1.17) rectangle ++(0.22, 0.13);
  \node[anchor=west, inner sep=1pt] at (0.3, -1.105) {per MMA};
  \draw[fill=gray!45, draw=gray!70!black, thin] (1.65, -1.17) rectangle ++(0.22, 0.13);
  \node[anchor=west, inner sep=1pt] at (1.9, -1.105) {per K-tile};

  \begin{scope}[xshift=4.05cm]
    \node[font=\scriptsize\bfseries, anchor=south] at (1.55, 2.65) {(b) SM100};
    \draw[axis] (0, 0) -- (3.1, 0);
    \draw[axis] (0, 0) -- (0, 2.5);
    \draw[tick] (0, 0) -- (-0.06, 0) node[left, inner sep=1pt] {0\%};
    \foreach \t/\lab in {0/1\%, 1/10\%, 2/100\%, 3/1{,}000\%, 4/10{,}000\%} {
      \draw[tick] (0, {0.3+0.53*\t}) -- (-0.06, {0.3+0.53*\t}) node[left, inner sep=1pt] {\lab};
    }
    \fill[white] (-0.08, 0.1) rectangle (0.08, 0.19);
    \draw[axis] (-0.08, 0.07) -- (0.08, 0.13);
    \draw[axis] (-0.08, 0.16) -- (0.08, 0.22);
    \foreach \i/\n/\v/\lab in {0/8/3913/3913, 1/16/374/374, 2/32/147/147,
                               3/64/48/48, 4/128/11/11} {
      \draw[meas] ({0.3+0.5*\i-0.18}, 0) rectangle ({0.3+0.5*\i+0.18}, {0.3+0.53*log10(\v)});
      \node[val] at ({0.3+0.5*\i}, {0.3+0.53*log10(\v)}) {\lab};
      \node[anchor=north, inner sep=2pt] at ({0.3+0.5*\i}, 0) {\n};
    }
    \draw[meas] (2.62, 0) rectangle (2.98, 0.21);
    \node[val] at (2.8, 0.21) {0.7};
    \node[anchor=north, inner sep=2pt] at (2.8, 0) {256};
    \node[anchor=north] at (1.55, -0.3) {format-tile width $N$};
    \draw[meas] (0.05, -0.92) rectangle ++(0.22, 0.13);
    \node[anchor=west, inner sep=1pt] at (0.3, -0.855) {GB200};
  \end{scope}
\end{tikzpicture}
\vskip -0.05in
\caption{Extra GEMM time over the stock NVFP4 kernel for $N \times 64$
         format tiles. (a)~SM120, $4096^3$: a branch per MMA against our branch
         per K-tile (Appendix~\ref{app:exec:dispatch}). (b)~SM100, $8192^3$ on
         a GB200: one MMA of width $N$ per tile (Appendix~\ref{app:exec:sm100};
         $N = 8$ runs on one SM); log scale above 1\%.}
\label{fig:overhead}
\vskip -0.1in
\end{figure}

A flip map is useful only if the GEMM executes it natively at close to the
cost of NVFP4. An E0M3 tile changes only the values written in the NVFP4
storage format, so every weight still takes 4.5 bits. The tag is free on
SM120, where it uses the unused sign bit of the tile's scales as in IF4 and
MixFP4~\cite{if4,mixfp4arxiv}, and costs below 0.0001 bits per weight on
SM100, one bit per $256 \times 64$ tile. 
The kernel reads a tile's tag
 before issuing its MMAs. SM120 fixes the 
format in the opcode, so the cost
 lies in the dispatch 
(\S\ref{sec:execution:sm120}); SM100 reads it from a
 register, so the cost 
lies in the operand width
 (\S\ref{sec:execution:sm100}). 
Figure~\ref{fig:overhead} summarizes the cost. We verify that the tensor core
executes every weight in the format its map assigns, and that native and
simulated perplexity agree within two standard errors (Appendix~\ref{app:native}).
Appendix~\ref{app:exec} gives the details.

\begin{figure}[t]
  \centering
  \definecolor{e2fill}{HTML}{DCE8FB}\definecolor{e2line}{HTML}{2C67D1}
  \definecolor{e0fill}{HTML}{F6C99A}\definecolor{e0line}{HTML}{B85C10}
  \definecolor{brfill}{HTML}{F2B8B2}\definecolor{brline}{HTML}{B8433B}
  \begin{tikzpicture}[
      font=\scriptsize,
      mma/.style={draw, inner sep=0pt, anchor=south west, font=\tiny,
                  minimum width=0.235cm, minimum height=0.40cm},
      e2/.style={mma, draw=e2line, fill=e2fill},
      e0/.style={mma, draw=e0line, fill=e0fill},
      off/.style={mma, draw=gray!70, dashed, text=gray!80,
                  pattern=north east lines, pattern color=gray!45},
      br/.style={draw=brline, fill=brfill},
      rule/.style={gray!25},
      lab/.style={anchor=west, align=left, inner sep=0pt, text width=2.4cm},
      key/.style={anchor=west, align=left, inner sep=0pt, text width=2.5cm},
    ]
    \node[e2, minimum width=0.28cm, minimum height=0.18cm] at (0.00,0.70) {};
    \node[anchor=west, inner sep=1pt] at (0.30,0.79) {E2M1 MMA};
    \node[e0, minimum width=0.28cm, minimum height=0.18cm] at (1.75,0.70) {};
    \node[anchor=west, inner sep=1pt] at (2.05,0.79) {E0M3 MMA};
    \node[off, minimum width=0.28cm, minimum height=0.18cm] at (3.50,0.70) {};
    \node[anchor=west, inner sep=1pt] at (3.80,0.79) {predicated off};
    \draw[br] (5.55,0.70) rectangle (5.83,0.88);
    \node[anchor=west, inner sep=1pt] at (5.85,0.79) {branch};

    \draw[-{Stealth[length=4pt]}] (2.5,0) -- (5.5,0)
      node[midway, above=1pt, inner sep=0pt] {time};
    \node[key, font=\scriptsize\itshape] at (5.7,0.12) {pros / cons};

    \draw[rule] (0,-0.88) -- (8.2,-0.88);
    \draw[rule] (0,-1.47) -- (8.2,-1.47);
    \draw[rule] (0,-2.45) -- (8.2,-2.45);

    \node[lab] at (0,-0.55) {(a) Stock NVFP4};
    \foreach \x in {2.50,2.75,3.00,3.25} \node[e2] at (\x,-0.75) {E2};
    \node[key] at (5.7,-0.55) {---};

    \node[lab] at (0,-1.18) {(b) Branch per MMA\\w/ predication};
    \node[e2]  at (2.50,-1.38) {E2};
    \node[off] at (2.75,-1.38) {E0};
    \node[e2]  at (3.00,-1.38) {E2};
    \node[off] at (3.25,-1.38) {E0};
    \node[off] at (3.50,-1.38) {E2};
    \node[e0]  at (3.75,-1.38) {E0};
    \node[e2]  at (4.00,-1.38) {E2};
    \node[off] at (4.25,-1.38) {E0};
    \node[key] at (5.7,-1.18) {\figcon\ wasted MMAs};

    \node[lab] at (0,-1.96) {(c) Branch per MMA\\w/o predication};
    \draw[br] (2.50,-2.16) rectangle (2.985,-1.76); \node[e2] at (3.00,-2.16) {E2};
    \draw[br] (3.25,-2.16) rectangle (3.735,-1.76); \node[e2] at (3.75,-2.16) {E2};
    \draw[br] (4.00,-2.16) rectangle (4.485,-1.76); \node[e0] at (4.50,-2.16) {E0};
    \draw[br] (4.75,-2.16) rectangle (5.235,-1.76); \node[e2] at (5.25,-2.16) {E2};
    \node[key] at (5.7,-1.96) {\figcon\ a jump per MMA\\\figcon\ idle tensor pipe};

    \node[lab] at (0,-2.93) {(d) Branch per K-tile\\(ours)};
    \draw[br] (2.50,-3.13) rectangle (2.55,-2.73);
    \node[e2] at (2.5625,-3.13) {E2};
    \node[e2] at (2.8125,-3.13) {E2};
    \node[e0] at (3.0625,-3.13) {E0};
    \node[e2] at (3.3125,-3.13) {E2};
    \node[key] at (5.7,-2.93) {\figpro\ no wasted MMAs\\\figpro\ few branches\\\figpro\ busy tensor pipe};
  \end{tikzpicture}
  \caption{One warp's K-tile on SM120, simplified to four of its 32 MMAs;
  widths are illustrative (Appendix~\ref{app:exec:dispatch} measures (b)--(d)).}
  \label{fig:sm120-dispatch}
\end{figure}
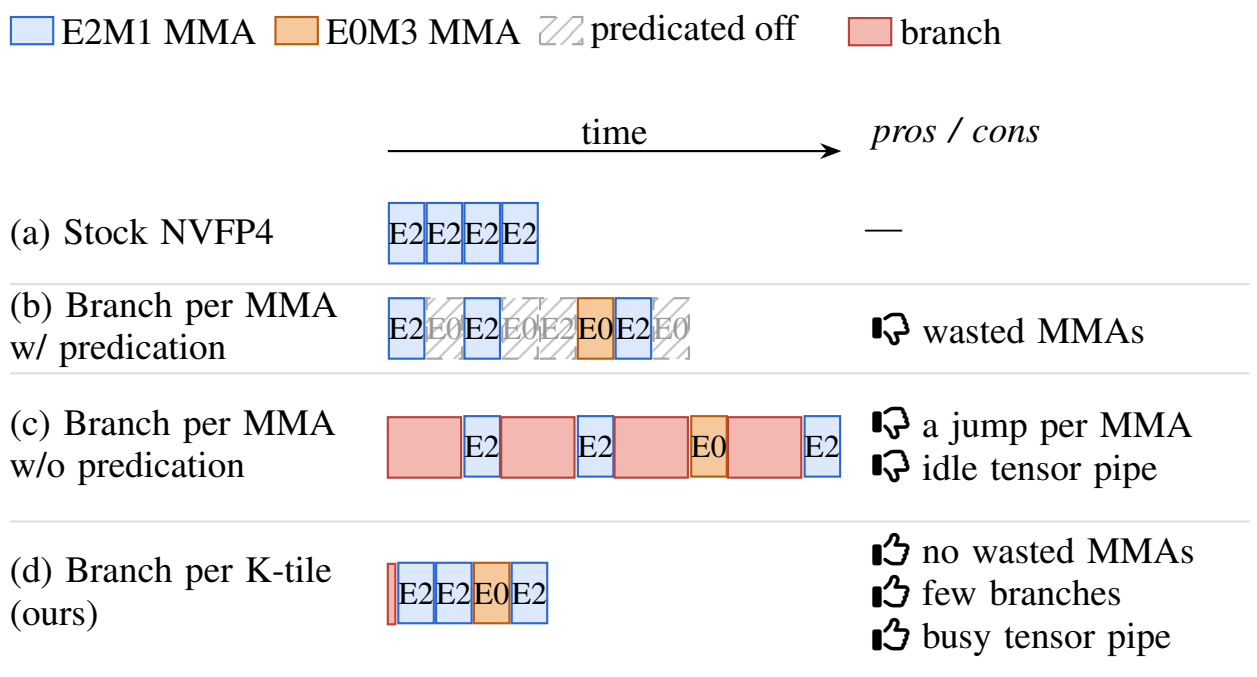

\subsection{SM120: Format Fixed in the Opcode}
\label{sec:execution:sm120}

\paragraph{An opcode per format.}
On SM120, \texttt{mma.sync} with \texttt{kind::mxf4nvf4} encodes each
operand's element format in its opcode, and PTX exposes only E2M1. We
therefore compile E2M1 instructions and patch the format bits of selected
MMA sites after compilation~\cite{revollllt2026e0m3}, so E2M1 and E0M3 MMAs
are different instructions. Since the format cannot be supplied at run time,
selecting E0M3 means choosing which instruction runs, that is, branching.
Because \texttt{mma.sync} executes warp-synchronously, the branch must also
be uniform across the warp, which tiles at least as large as one MMA operand
guarantee. Selecting a format thus costs a branch, and where the branch sits
decides how much (Figure~\ref{fig:sm120-dispatch}).

\paragraph{A branch per K-tile.}
The branch can sit at each MMA or once per K-tile, the 128 elements along
$K$ that one pipeline stage loads and over which each warp issues 32 MMAs.
A branch per MMA is costly whether or not the compiler keeps it. The
compiler turns it into a pair of predicated MMAs, one per format
(Figure~\ref{fig:sm120-dispatch}b), and the predicated-off MMA still
occupies a tensor-pipe issue slot, which doubles the issued MMAs. Of the
forms we tried, only an indirect jump stays a real branch
(Figure~\ref{fig:sm120-dispatch}c). It wastes no MMAs, but before every MMA
the warp must load the jump target from constant memory, branch, and
resynchronize. The shared-memory loads are still prefetched a K-block ahead,
but the MMAs no longer issue back to back: the tensor pipe sits idle most of
the time, and the GEMM takes 3.5--3.8$\times$ as long as the stock kernel
(Appendix~\ref{app:exec:dispatch}).
Our kernels instead branch once per K-tile. For every combination of
formats that the tiles a warp consumes in one K-tile can take, the kernel
contains a copy of the K-tile's mainloop body in which the format of every
MMA is fixed at compile time; we call these copies \emph{arms}. With
$16 \times 64$ tiles, for example, a warp consumes four tiles per K-tile, so
the kernel has $2^4 = 16$ arms (Appendix~\ref{app:exec:dispatch}). At each
K-tile, a warp reads the format tags of its tiles, combines them into an arm
index, and branches to that arm; all threads of the warp read the same tags,
so the branch is warp-uniform (Figure~\ref{fig:sm120-dispatch}d). Each arm
encloses the stage's barriers, which the compiler does not if-convert, so
the branch stays a direct branch, and the chosen arm issues each MMA once,
in its tile's format: as many MMAs as the stock kernel, back to back as in
the stock kernel. The arms enlarge the kernel's code, not the instructions
it executes, and the 32 MMAs of a K-tile share one branch. At $4096^3$, this
makes the GEMM 1.47--1.57$\times$ faster than the predicated branch per MMA,
and against a stock NVFP4 kernel with the same tile tuning, the GEMMs of a
forward pass take 1.9\%, 1.4\% and 7.3\% longer with $16 \times 64$,
$256 \times 64$ and $8 \times 64$ tiles.
Appendices~\ref{app:exec:dispatch}--\ref{app:exec:gemm} give the kernels,
their optimizations, and these measurements.

\subsection{SM100: Format Set in the MMA Descriptor}
\label{sec:execution:sm100}

\paragraph{A register write per MMA.}
On SM100, \texttt{tcgen05.mma} with \texttt{kind::mxf4nvf4} reads the operand
formats from the instruction descriptor, which is passed in a general-purpose
register. Before issuing the MMA for K-block $k$ of output tile $n$, the
issuing thread reads the map entry of tile $(n, k)$ and writes the $B$-format
field of the descriptor. The same instruction executes either
format, with no branch or extra MMA, so selecting E0M3 costs one
register write.

\paragraph{Why $256 \times 64$ tiles.}
The weights are the $B$ operand, and \texttt{tcgen05.mma} accepts weight
operands from 8 to 256 output channels wide in steps of 8, so tiles from
$8 \times 64$ to $256 \times 64$ are all executable. Narrow operands, however,
lower the arithmetic intensity of every MMA: each instruction reuses the
same $128 \times 64$ activation tile across only $N$ weight columns, so the
kernel issues more instructions and moves more operand and scale data per
FLOP. On a GB200, with each 256-wide MMA of our kernel issued as $256/N$
narrower ones, the GEMM at $8192^3$ takes 11\% longer than the stock kernel at
$N = 128$, 48\% longer at $N = 64$, $4.7\times$ as long at $N = 16$, and
$40\times$ as long at $N = 8$, where the instruction also restricts the kernel
to one SM (Figure~\ref{fig:overhead}b). We therefore use $256 \times 64$
tiles on SM100, the widest operand, and
Section~\ref{sec:eval} measures what this coarser granularity costs in
quality. Appendix~\ref{app:exec:sm100} gives the measurements.

\providecommand{\pend}[1]{\textcolor{red}{#1}} %
\section{Evaluation and Analysis}
\label{sec:eval}

We first describe the setup (\S\ref{sec:eval:setup}). We then measure the
accuracy of \method{} (\S\ref{sec:eval:quality}), its cost at inference on both
Blackwell variants (\S\ref{sec:eval:latency}), and the cost and robustness
of its calibration (\S\ref{sec:eval:cost}). Finally, we analyze how the map is
chosen and how \method{} combines with existing post-training quantization
methods (\S\ref{sec:eval:analysis}).

\setcounter{dbltopnumber}{1}
\begin{table*}[t]
  \caption{W4A4 perplexity on WikiText-2 (Wiki) and C4 (lower is better).
           Bold and underlined: best and second-best natively executable
           quantized result per column (BF16 excluded; ties at the displayed
           precision share the mark). $^{\ddagger}$: GPTQ-quantized
           NVFP4 weights with FourOverSix activations. Last column: the share
           of NVFP4's perplexity gap to BF16 that a method closes,
           $100 \cdot (\mathrm{PPL}_{\mathrm{NVFP4}} - \mathrm{PPL}) / (\mathrm{PPL}_{\mathrm{NVFP4}} - \mathrm{PPL}_{\mathrm{BF16}})$,
           averaged over the twelve model--corpus pairs.}
  \label{tab:main-ppl}
  \vskip 0.1in
  \begin{center}
  \begin{footnotesize}
  \setlength{\tabcolsep}{3.4pt}
  \begin{tabular}{@{}llccccccccccccc@{}}
    \toprule
     & & \multicolumn{2}{c}{Llama-3.2} & \multicolumn{2}{c}{Llama-3.1} & \multicolumn{2}{c}{Mistral} & \multicolumn{2}{c}{Nemotron} & \multicolumn{2}{c}{Phi-4} & \multicolumn{2}{c}{Qwen3.8} & PPL gap \\
    \cmidrule(lr){3-4} \cmidrule(lr){5-6} \cmidrule(lr){7-8} \cmidrule(lr){9-10} \cmidrule(lr){11-12} \cmidrule(lr){13-14}
     & & \multicolumn{2}{c}{1B} & \multicolumn{2}{c}{8B} & \multicolumn{2}{c}{7B} & \multicolumn{2}{c}{9B} & \multicolumn{2}{c}{14B} & \multicolumn{2}{c}{27B} & closed \\
    \cmidrule(lr){3-4} \cmidrule(lr){5-6} \cmidrule(lr){7-8} \cmidrule(lr){9-10} \cmidrule(lr){11-12} \cmidrule(lr){13-14}
    Method & Tile & Wiki & C4 & Wiki & C4 & Wiki & C4 & Wiki & C4 & Wiki & C4 & Wiki & C4 & (\%)~$\uparrow$ \\
    \midrule
    BF16 & -- & 13.16 & 18.52 & 7.22 & 10.39 & 5.50 & 8.14 & 8.09 & 11.15 & 6.46 & 10.31 & 7.05 & 9.89 & 100.0 \\
    \midrule
    \multicolumn{15}{@{}l}{\emph{Uniform format, native on existing GPUs}} \\
    NVFP4 & -- & 15.54 & 21.77 & 7.89 & 11.34 & 5.73 & 8.42 & 8.45 & 11.54 & 6.69 & 10.58 & 7.57 & 10.22 & 0.0 \\
    FourOverSix & -- & 15.32 & 21.45 & 7.82 & 11.27 & 5.71 & 8.40 & 8.43 & 11.50 & 6.66 & 10.54 & 7.30 & 10.19 & 13.3 \\
    GPTQ$^{\ddagger}$ & -- & 15.47 & 21.10 & 7.93 & 11.43 & 5.73 & 8.40 & 8.40 & 11.49 & 6.67 & 10.54 & 7.44 & 10.14 & 9.5 \\
    \midrule
    \multicolumn{15}{@{}l}{\emph{Per-block format, simulated (needs new hardware)}} \\
    IF4 & $1{\times}16$ & 15.34 & 21.07 & 7.74 & 11.18 & 5.67 & 8.36 & 8.39 & 11.45 & 6.63 & 10.51 & 7.33 & 10.17 & 22.3 \\
    MixFP4 & $1{\times}16$ & 15.25 & 21.05 & 7.73 & 11.16 & 5.66 & 8.37 & 8.41 & 11.46 & 6.64 & 10.51 & 7.35 & 10.16 & 22.3 \\
    \midrule
    \multicolumn{15}{@{}l}{\emph{Per-tile format, native (ours)}} \\
    \textbf{\method{}} & $8{\times}64$ & \underline{14.40} & \textbf{19.57} & \textbf{7.65} & \underline{11.05} & \textbf{5.63} & \textbf{8.30} & \textbf{8.32} & \textbf{11.44} & \textbf{6.61} & \textbf{10.49} & \textbf{7.05} & \textbf{10.11} & \textbf{44.2} \\
    \textbf{\method{}} & $16{\times}64$ & \textbf{14.39} & \underline{19.61} & \textbf{7.65} & \textbf{11.04} & \underline{5.64} & \textbf{8.30} & \underline{8.33} & \textbf{11.44} & \underline{6.62} & \underline{10.50} & \underline{7.08} & \underline{10.12} & \underline{42.5} \\
    \textbf{\method{}} & $256{\times}64$ & 14.70 & 20.04 & \underline{7.71} & 11.12 & \underline{5.64} & \underline{8.31} & 8.35 & \underline{11.46} & 6.64 & 10.52 & 7.17 & 10.15 & 34.5 \\
    \bottomrule
  \end{tabular}
  \end{footnotesize}
  \end{center}
  \vskip -0.1in
\end{table*}

\begin{table*}[t]
  \caption{Downstream accuracy (\%, higher is better). 7-task: mean over the
           seven log-likelihood tasks of Section~\ref{sec:eval:setup}; GSM8K:
           8-shot chain of thought, strict match; Mean: over the six models.
           $^{\ddagger}$ and marks as in Table~\ref{tab:main-ppl}. Per-task
           results are in Appendix~\ref{app:downstream}.}
  \label{tab:main-acc}
  \vskip 0.1in
  \begin{center}
  \begin{footnotesize}
  \setlength{\tabcolsep}{1.4pt}
  \begin{tabular}{@{}llcccccccccccccc@{}}
    \toprule
     & & \multicolumn{2}{c}{Llama-3.2} & \multicolumn{2}{c}{Llama-3.1} & \multicolumn{2}{c}{Mistral} & \multicolumn{2}{c}{Nemotron} & \multicolumn{2}{c}{Phi-4} & \multicolumn{2}{c}{Qwen3.8} & \multicolumn{2}{c}{Mean} \\
    \cmidrule(lr){3-4} \cmidrule(lr){5-6} \cmidrule(lr){7-8} \cmidrule(lr){9-10} \cmidrule(lr){11-12} \cmidrule(lr){13-14}
     & & \multicolumn{2}{c}{1B} & \multicolumn{2}{c}{8B} & \multicolumn{2}{c}{7B} & \multicolumn{2}{c}{9B} & \multicolumn{2}{c}{14B} & \multicolumn{2}{c}{27B} & \multicolumn{2}{c}{} \\
    \cmidrule(lr){3-4} \cmidrule(lr){5-6} \cmidrule(lr){7-8} \cmidrule(lr){9-10} \cmidrule(lr){11-12} \cmidrule(lr){13-14} \cmidrule(lr){15-16}
    Method & Tile & 7-task & GSM8K & 7-task & GSM8K & 7-task & GSM8K & 7-task & GSM8K & 7-task & GSM8K & 7-task & GSM8K & 7-task & GSM8K \\
    \midrule
    BF16 & -- & 57.95 & 45.4 & 72.94 & 85.9 & 74.06 & 47.0 & 72.67 & 87.4 & 74.50 & 93.3 & 75.20 & 95.7 & 71.22 & 75.8 \\
    \midrule
    \multicolumn{16}{@{}l}{\emph{Uniform format, native on existing GPUs}} \\
    NVFP4 & -- & 54.30 & 29.2 & 70.84 & 80.0 & 73.39 & 44.5 & \textbf{71.72} & 87.3 & 73.38 & 93.0 & 74.94 & 95.4 & 69.76 & 71.6 \\
    FourOverSix & -- & 54.11 & 29.7 & 70.88 & 79.5 & 73.20 & 44.5 & 71.20 & \textbf{88.9} & 73.37 & 91.7 & 74.86 & 94.6 & 69.60 & 71.5 \\
    GPTQ$^{\ddagger}$ & -- & 54.52 & \underline{33.6} & 70.76 & \textbf{81.4} & \underline{73.42} & 44.0 & 71.39 & 87.0 & \textbf{73.76} & \textbf{93.3} & 74.57 & \underline{96.2} & 69.74 & 72.6 \\
    \midrule
    \multicolumn{16}{@{}l}{\emph{Per-block format, simulated (needs new hardware)}} \\
    IF4 & $1{\times}16$ & 54.83 & 33.5 & 71.24 & 80.4 & 73.37 & 45.2 & 71.30 & 87.1 & 73.69 & 92.6 & 75.66 & 95.6 & 70.01 & 72.4 \\
    MixFP4 & $1{\times}16$ & 55.05 & 33.6 & 71.12 & 80.7 & 73.57 & 45.6 & 71.38 & 85.5 & 73.68 & 93.4 & 75.11 & 95.4 & 69.98 & 72.4 \\
    \midrule
    \multicolumn{16}{@{}l}{\emph{Per-tile format, native (ours)}} \\
    \textbf{\method{}} & $8{\times}64$ & 55.55 & 32.0 & \textbf{71.38} & \underline{81.2} & 73.31 & \textbf{46.6} & \underline{71.58} & 88.0 & 73.40 & 92.3 & 75.18 & 95.8 & \underline{70.07} & \underline{72.7} \\
    \textbf{\method{}} & $16{\times}64$ & \underline{55.65} & \textbf{35.2} & \underline{71.34} & 80.4 & \textbf{73.47} & 45.8 & 71.44 & \underline{88.6} & \underline{73.69} & \underline{93.2} & \textbf{75.59} & 95.9 & \textbf{70.20} & \textbf{73.2} \\
    \textbf{\method{}} & $256{\times}64$ & \textbf{55.77} & 32.9 & 70.83 & 80.4 & 73.04 & \underline{45.9} & 71.56 & 86.5 & 73.57 & 92.8 & \underline{75.21} & \textbf{96.5} & 70.00 & 72.5 \\
    \bottomrule
  \end{tabular}
  \end{footnotesize}
  \end{center}
  \vskip -0.1in
\end{table*}

\subsection{Experimental Setup}
\label{sec:eval:setup}

\paragraph{Models.}
We evaluate six models from 1B to 27B parameters: Llama-3.2-1B-Instruct and
Llama-3.1-8B-Instruct~\cite{llama3} (Llama-3.2-1B and Llama-3.1-8B),
Mistral-7B-Instruct-v0.3~\cite{mistral} (Mistral-7B),
Nemotron-Nano-9B-v2~\cite{nemotronnano2} (Nemotron-9B, hybrid
Mamba-Transformer), Phi-4~\cite{phi4}, and Qwen3.8-27B~\cite{qwen38} (hybrid
linear attention). We use the instruction-tuned release of each, as deployed
and as our generative benchmarks require, so perplexity serves only to compare
methods on the same model. Appendix~\ref{app:setup:models} lists the pinned
releases and the quantized layers.

\paragraph{Quantization and baselines.}
We quantize the weights and inputs of every linear layer except the output
head (W4A4), with one FP32 activation scale per token for every quantization method; all
other modules stay in BF16. We report \method{} with FourOverSix activations at three tile sizes, 
$8 \times 64$, $16 \times 64$, and $256 \times 64$. We compare with BF16; with
uniform-format methods that run natively on existing GPUs: NVFP4,
FourOverSix~\cite{fouroversix},
and GPTQ~\cite{gptq} on NVFP4 weights with FourOverSix
activations; and with per-block formats that need new hardware and are
simulated: IF4~\cite{if4} and MixFP4~\cite{mixfp4arxiv}.
Appendix~\ref{app:setup:baselines} gives the detailed settings of every baseline.

\paragraph{Calibration.}
Each map is trained on 1{,}024 sequences of 512 tokens from
OpenWebMath~\cite{openwebmath} and CodeParrot~\cite{codeparrot}, disjoint from
all evaluation data. The loss is the bucketed KL divergence to the BF16 teacher
(Eq.~\eqref{eq:objective}). We use Adam ($\eta = 0.02$,
$\epsilon = 10^{-12}$) with 8 sequences per step for 5 epochs (640 steps),
starting from $\theta_u = -1$, with seed 0 and deterministic kernels. No
held-out data is consulted, and the map after the last step is deployed
(Appendix~\ref{app:calib}).

\paragraph{Evaluation.}
We report perplexity on WikiText-2~\cite{wikitext} (2048-token windows) and
C4~\cite{t5} (256 excerpts of 2048 tokens), with paired $\Delta$NLL and two
standard errors. Downstream accuracy is measured on all six models with
lm-eval 0.4.11~\cite{lmeval}. Following RaZeR~\cite{razer}, we evaluate
PIQA~\cite{piqa}, HellaSwag~\cite{hellaswag}, WinoGrande~\cite{winogrande},
ARC-Easy and ARC-Challenge~\cite{arc}, and LAMBADA~\cite{lambada} zero-shot,
and add MMLU~\cite{mmlu} (5-shot). These
tasks are scored by log-likelihood without a chat template, with normalized
accuracy for PIQA, HellaSwag, and ARC. GSM8K~\cite{gsm8k} also follows RaZeR:
8-shot chain of thought in the chat template, greedy decoding of at most 1024
new tokens with thinking off, and strict matching of the final answer.
Appendix~\ref{app:setup:eval} lists the prompts, dataset revisions,
and metrics.

\paragraph{Execution and measurement.}
Perplexity and downstream accuracy are measured natively on RTX PRO 6000
GPUs, and latency on the RTX PRO 6000 and a GB200. IF4 and MixFP4 need new hardware and are simulated; simulating
\method{} maps, which also run natively, reproduces their native perplexity
(Appendix~\ref{app:native}). Latencies are medians of five rounds, one process
per (model, policy, round), with the policy order shuffled in every round, and
relative changes are paired within rounds; end-to-end latencies use CUDA
graphs (Appendix~\ref{app:setup:eval}).

\subsection{Accuracy}
\label{sec:eval:quality}

Table~\ref{tab:main-ppl} compares perplexity on all six models, and
Table~\ref{tab:main-acc} asks whether the gains carry over to downstream
tasks.

\paragraph{Perplexity.}
\method{} matches or beats every natively executable baseline in every column
of Table~\ref{tab:main-ppl}. Because the maps start from FourOverSix, the
comparison with it isolates their effect: \method{} is significantly better
than FourOverSix and NVFP4 on every model and corpus at all three tiles, and at
$16 \times 64$ closes 42.5\% of NVFP4's PPL gap to BF16, 3.2 times the 13.3\%
of FourOverSix. It also exceeds the per-block formats IF4 and MixFP4 (22.3\%
each), which need new hardware. $8 \times 64$ and $16 \times 64$ are about equally accurate, differing in
$\Delta$NLL by at most 0.005 nats per token; $256 \times 64$, with 16 times as
many weights per decision, gives up about a quarter of the $16 \times 64$ gain
over FourOverSix but still closes 34.5\% of the gap, 2.6 times as much as
FourOverSix.

\begin{table}[!b]
      \caption{End-to-end prefill latency.}
  \label{tab:latency}
  \vskip 0.1in
  \begin{center}
  \begin{scriptsize}
  \setlength{\tabcolsep}{2.6pt}
  \begin{tabular}{@{}lcccccc@{}}
    \toprule
     & Llama-3.2 & Llama-3.1 & Mistral & Nemotron & Phi-4 & Qwen3.8 \\
    Method & 1B & 8B & 7B & 9B & 14B & 27B \\
    \midrule
    \multicolumn{7}{c}{\emph{RTX PRO 6000 (SM120): prefill, $1 \times 2048$ tokens, ms}} \\
    \midrule
    BF16 & 22.9 & 131.2 & 125.1 & 151.1 & 244.7 & 453.7 \\
    NVFP4 & 12.0 & 61.3 & 55.6 & 79.8 & 109.9 & 201.7 \\
    FourOverSix & 12.0 & 61.6 & 55.9 & 80.7 & 110.5 & 202.7 \\
    \cmidrule(lr){1-7}
    \textbf{\method{}} $16{\times}64$ & 12.2 & 61.7 & 56.0 & 81.4 & 111.8 & 204.5 \\
    \textbf{\method{}} $8{\times}64$ & 12.6 & 64.8 & 58.7 & 84.1 & 116.7 & 212.7 \\
    \midrule
    \multicolumn{7}{c}{\emph{GB200 (SM100): prefill, $1 \times 2048$ tokens, ms}} \\
    \midrule
    BF16 & 7.8 & 32.4 & 31.6 & 40.7 & 67.9 & 122.0 \\
    NVFP4 & 6.7 & 22.5 & 21.6 & 30.9 & 44.5 & 81.5 \\
    FourOverSix & 6.8 & 22.9 & 22.1 & 31.3 & 45.3 & 82.6 \\
    \cmidrule(lr){1-7}
    \textbf{\method{}} $256{\times}64$ & 6.8 & 23.0 & 22.2 & 31.4 & 45.4 & 82.8 \\
    \bottomrule
  \end{tabular}
  \end{scriptsize}
  \end{center}
  \vskip -0.1in
\end{table}

\paragraph{Downstream accuracy.}
Table~\ref{tab:main-acc} tests whether the perplexity gains carry over to
downstream tasks. Averaged over the seven log-likelihood tasks, \method{} at
$16 \times 64$ changes the accuracy of FourOverSix by $+0.24$ to $+1.53$
points across models. Its six-model mean is 70.20\%, the highest of all quantized
methods, against 69.74\% for GPTQ and 70.0\% for IF4 and MixFP4. On GSM8K, where errors can
compound over a chain of thought, $16 \times 64$ changes the accuracy of
FourOverSix by $-0.3$ to $+5.5$ points across the six models, averaging 73.17\%
against 71.49\%. It has the highest six-model GSM8K mean of all quantized
methods, natively executable or simulated, followed by $8 \times 64$ at
72.67\%, GPTQ at 72.58\%, and $256 \times 64$ at 72.52\%; the simulated IF4 and
MixFP4 means are 72.42\% and 72.35\%. Per model, however, the rankings
depend on the task; the perplexity advantage does not translate
into a uniform lead in downstream accuracy. Appendix~\ref{app:downstream}
gives the individual accuracies for the seven log-likelihood tasks.

\subsection{Inference Efficiency}
\label{sec:eval:latency}
Section~\ref{sec:execution} explains where the GEMM overhead of the mixed
format comes from on each variant; here we measure its end-to-end effect
(Table~\ref{tab:latency}).

\paragraph{SM120.}
At $1 \times 2048$ tokens, the $16 \times 64$ maps are within $+0.1$ to
$+1.1\%$ of FourOverSix on every model, and \method{} stays
1.9--2.2$\times$ faster than BF16. The $8 \times 64$ maps cost $+4.2$ to
$+5.7\%$ (Section~\ref{sec:execution:sm120}).

\paragraph{SM100.}
On the GB200, the $256 \times 64$ maps change the prefill latency of
FourOverSix by $+0.2$ to $+0.4\%$, while keeping about three
quarters of the $16 \times 64$ perplexity gain over FourOverSix
(\S\ref{sec:eval:quality}). FP4 gains less over BF16 here than on SM120
(FourOverSix takes 0.67--0.88$\times$ the BF16 time).

\paragraph{Memory.}
The map lives in bits that the NVFP4 scales already store (SM120) or in one
bit per $256 \times 64$ tile (SM100; at most 0.21\,MB per model), and the
kernels need no workspace.
On SM120, \method{}, NVFP4, and FourOverSix store the same weight bytes,
and their memory after loading and at the prefill peak differs by at most
0.013\,GiB.

\subsection{Calibration Cost and Robustness}
\label{sec:eval:cost}

\paragraph{Cost.}
We measure one calibration end to end, from model loading to the written
map, with no development set and deterministic kernels. In total, the models calibrate in 3.7 to 170 minutes
on one GPU, with at most 91\,GiB of GPU memory. Training accounts for 73--93\%
of the trainer's own total (setup plus training), or approximately 70--93\% of
the end-to-end time (Appendix~\ref{app:cost:calib}). The tile shape barely
matters: going from
$256 \times 64$ to $8 \times 64$ multiplies the number of tiles by 32 but
changes the calibration time by at most 1.9\% and the peak GPU memory by at most
0.9\,GiB, as Section~\ref{sec:method:training} predicts
(Appendix~\ref{app:cost:calib}).

\paragraph{Robustness.}
On Mistral-7B at $16 \times 64$ (Appendix~\ref{app:cost:robust}), other
seeds and non-deterministic kernels give maps whose E0M3 tiles overlap the
paper's by a Jaccard index of only 0.34--0.38, yet match its $\Delta$NLL
within two standard errors: the map is not unique, but its quality is.
Calibrating on web text or instruction data instead of math and code keeps
82--103\% of the gain over FourOverSix. More calibration helps only a
little: four times the paper's 640 steps improve WikiText-2 by at most a
further $2.6 \times 10^{-3}$ nats per token, and the best of these maps flip
about 20\% of the tiles, which raises the inference cost
(Appendix~\ref{app:exec:opt}).

\subsection{Analysis}
\label{sec:eval:analysis}

\begin{table}[!t]
  \caption{How the map is chosen: paired $\Delta$NLL against FourOverSix
           on WikiText-2 ($\times 10^{-3}$ nats per token, negative is better;
           2\,SE 0.9--8.6). Bold: best per column and maps within 2\,SE of it.}
  \label{tab:ablation}
  \vskip 0.1in
  \begin{center}
  \begin{footnotesize}
  \setlength{\tabcolsep}{3pt}
  \begin{tabular*}{\columnwidth}{@{\extracolsep{\fill}}lccc@{}}
    \toprule
    Map & $8 \times 64$ & $16 \times 64$ & $256 \times 64$ \\
    \midrule
    \multicolumn{4}{c}{\emph{Nemotron-9B}} \\
    \midrule
    Random & $-0.7$ & $-0.3$ & $+0.7$ \\
    Activation-weighted & $-1.3$ & $-0.8$ & $+2.8$ \\
    One-shot gradient & $+89.9$ & $+56.8$ & $-2.1$ \\
    \cmidrule(lr){1-4}
    \textbf{\method{}} & $\mathbf{-12.9}$ & $\mathbf{-11.9}$ & $\mathbf{-9.6}$ \\
    \midrule
    \multicolumn{4}{c}{\emph{Qwen3.8-27B}} \\
    \midrule
    Random & $+2.1$ & $-2.8$ & $+4.2$ \\
    Activation-weighted & $-8.3$ & $-8.3$ & $-3.9$ \\
    One-shot gradient & $+3.9$ & $-14.5$ & $\mathbf{-29.9}$ \\
    \cmidrule(lr){1-4}
    \textbf{\method{}} & $\mathbf{-35.4}$ & $\mathbf{-31.1}$ & $-17.7$ \\
    \bottomrule
  \end{tabular*}
  \end{footnotesize}
  \end{center}
  \vskip -0.1in
\end{table}

\paragraph{Which tiles, and how they are chosen.}
The learned maps keep most tiles in E2M1: they flip 2.3--14.8\% of the tiles to E0M3 at
$8 \times 64$ and $16 \times 64$, and 7.1--39.7\% at $256 \times 64$
(Appendix~\ref{app:eval:share}). The maps of Table~\ref{tab:ablation} flip as
many tiles as \method{} in every layer and differ only in which. Random maps
(mean of three seeds) are no better than FourOverSix ($-2.8$ to $+4.2 \times 10^{-3}$), so the gain
comes from which tiles are flipped. Activation-weighted selection, which uses
each layer's output error alone and so ignores the sensitivity of later layers
(\S\ref{sec:method:gradient}), recovers little of it ($-8.3$ to $+2.8$, against $-9.6$ to $-35.4 \times 10^{-3}$ for
\method{}). Selecting once from the tile gradient at the all-E2M1 map
(one-shot gradient), without the iterative optimization of \S\ref{sec:method:training}, fails at the fine tiles: at
$8 \times 64$ and $16 \times 64$ it is far worse than FourOverSix on Nemotron-9B,
and on Qwen3.8-27B it is worse than FourOverSix at $8 \times 64$ and reaches
about half of \method{}'s gain at $16 \times 64$. At $256 \times 64$, it gains
less than \method{} on Nemotron-9B but more on Qwen3.8-27B.

\paragraph{Does the tile gradient rank flips correctly?}
On Mistral-7B at $16 \times 64$ (Appendix~\ref{app:eval:gradcheck}), $g_u$
ranks groups of 1{,}000 flipped tiles by their measured decrease in bucketed
KL divergence (Spearman $+0.72$ on held-out and $+0.88$ on calibration
sequences), while scores that look at one layer at a time do not. For up to
100 tiles flipped together, the measured change is 0.73--0.90 times its
first-order prediction on the calibration sequences and 0.74--0.84 times on
held-out ones. The prediction fails for more tiles: the
359{,}840 tiles with the most negative $g_u$ raise the held-out bucketed KL
divergence by $69.4 \times 10^{-3}$, whereas the paper's map, as many tiles chosen by
re-evaluating the gradient, lowers it by
$22.5 \times 10^{-3}$. This is why \method{} re-evaluates the gradient during
calibration.

\paragraph{Combining with post-training quantization.}
\method{} changes only how each tile's codes are interpreted, so it composes
with methods that change the values being quantized. Applied after GPTQ or a
block-16 Hadamard rotation, with the map trained on the method's result,
\method{} improves on the method's FourOverSix variant on Nemotron-9B and
Qwen3.8-27B and on both corpora. Neither method adds to \method{} alone on
WikiText-2; GPTQ adds $1.1$ and $2.0 \times 10^{-3}$ nats per token on C4 but
loses GSM8K accuracy on Nemotron-9B (Appendix~\ref{app:eval:ptq}).
\setcounter{dbltopnumber}{2}

\section{Conclusion}
\method{} makes mixed FP4 format selection executable at MMA operand
granularity. It learns a hard E2M1/E0M3 map from end-to-end gradients and
runs it with low GEMM overhead on SM120 and SM100. Across six models and two
corpora, the $16 \times 64$ map closes 42.5\% of NVFP4's PPL gap to BF16,
compared with 13.3\% for FourOverSix. Downstream accuracy remains dependent
on the task and model.

\bibliography{refs}
\bibliographystyle{mlsys2025}

\section*{AI Usage Disclosure}

The authors used ChatGPT (OpenAI) and Claude (Anthropic) as
assistive tools during the research and preparation of this manuscript.

During research development, these tools were used to support
brainstorming, discussion of methodological alternatives and
mathematical formulations, and the discovery, summarization, and
comparison of related literature. During implementation and
evaluation, they assisted with code generation, modification, and
debugging, as well as experimental planning, baseline selection,
and analysis of experimental results.

During manuscript preparation, the tools were used for drafting and
revising text, improving grammatical correctness and clarity,
preparing or modifying figures and tables, and checking submission
requirements, formatting, and bibliographic information.

The authors retain full responsibility for the correctness,
originality, and integrity of the research and all submitted materials.

\clearpage
\twocolumn[
\mlsystitle{Gradient-Guided Hardware-Native MixFP4 (Appendix)}
\vskip 0.3in
]
\setcounter{table}{0}\setcounter{figure}{0}

\appendix
\counterwithin{table}{section}
\counterwithin{figure}{section}
\renewcommand{\theHtable}{app.\thesection.\arabic{table}}
\renewcommand{\theHfigure}{app.\thesection.\arabic{figure}}

\section{Experimental Setup}
\providecommand{\pend}[1]{\textcolor{red}{#1}} %
\label{app:setup}
Section numbers refer to the main paper, and appendix letters to this
document.

\subsection{Models}
\label{app:setup:models}

Table~\ref{tab:app-models} lists the six models. We use the
instruction-tuned release of each at a pinned revision, in BF16, with
Transformers 5.16.1 and PyTorch 2.9.0 (CUDA 12.8).

\begin{table*}[!b]
  \caption{Models. Type: DeltaNet hybrid = Gated DeltaNet and attention. Layers: the quantized linear layers (W4A4). Weights:
           the quantized weights in billions. Tiles: format tiles at
           $16 \times 64$ in millions; $8 \times 64$ has twice as many, and
           $256 \times 64$ one sixteenth, rounded up per matrix (Appendix~\ref{app:exec:mapping}).}
  \label{tab:app-models}
  \vskip 0.1in
  \begin{center}
  \begin{footnotesize}
  \setlength{\tabcolsep}{4pt}
  \begin{tabular}{@{}lllrrr@{}}
    \toprule
    Model & Hugging Face ID & Type & Layers & Weights (B) & Tiles (M) \\
    \midrule
    Llama-1B    & \texttt{meta-llama/Llama-3.2-1B-Instruct}  & Transformer                    & 112 & 0.97  & 0.95  \\
    Llama-8B    & \texttt{meta-llama/Llama-3.1-8B-Instruct}  & Transformer                    & 224 & 6.98  & 6.82  \\
    Mistral-7B  & \texttt{mistralai/Mistral-7B-Instruct-v0.3} & Transformer                    & 224 & 6.98  & 6.82  \\
    Nemotron-9B & \texttt{nvidia/NVIDIA-Nemotron-Nano-9B-v2}  & Mamba-2 hybrid & 120 & 7.71  & 7.53  \\
    Phi-4       & \texttt{microsoft/phi-4}                    & Transformer                    & 160 & 13.63 & 13.31 \\
    Qwen3.8-27B & \texttt{Qwen/Qwen3.8-27B}                   & DeltaNet hybrid & 496 & 24.35 & 23.78 \\
    \bottomrule
  \end{tabular}
  \end{footnotesize}
  \end{center}
  \vskip -0.1in
\end{table*}

\paragraph{Quantized layers.}
Every linear layer except the output head is quantized, weights and inputs
alike; embeddings, normalization layers, the output head, and every operator
that is not a linear layer stay in BF16. In the Transformer layers these are
the query, key, value, and output projections and the MLP projections; Phi-4
stores the query, key, and value projections, and the gate and up
projections, as fused matrices, which we quantize as they are stored. Every
input dimension is a multiple of 64, so every K-block is a full MMA operand
and no layer is left unquantized.

\paragraph{Hybrid layers.}
Nemotron-9B has 27 Mamba-2 layers, 25 MLP layers, and 4 attention layers. We
quantize the input and output projections of each Mamba-2 layer; the
convolution, the selective scan, and its parameters ($A$, $D$, $\Delta t$)
stay in BF16. Qwen3.8-27B has 48 Gated DeltaNet layers, 16 attention layers,
and 64 MLP layers. We quantize all five projections of each Gated DeltaNet
layer, including the two 48-row projections that produce the gates; the
convolution and the recurrence stay in BF16. For Qwen3.8-27B, only the linear layers of the language model are quantized;
its vision encoder is loaded but stays in BF16 and never runs on our text-only
inputs, and its multi-token-prediction head is not loaded.

\subsection{Baselines}
\label{app:setup:baselines}

All FP4 methods share the NVFP4 container: an FP32 tensor scale
$\alpha = \mathrm{amax}/(6 \cdot 448)$, computed per token for activations,
one UE4M3 scale per 16 elements along $K$, and 4-bit sign-magnitude codes.
They differ in how they choose scales, codes, and formats.

\paragraph{NVFP4~\cite{nvfp4blog}.}
Each 16-element block takes the scale $\mathrm{e4m3}(\max|x|/6)$, relative
to $\alpha$, and each element the nearest E2M1 value, saturating at 6. The
NVFP4 rows quantize the activations with NVFP4 as well, with one tensor scale
per token.

\paragraph{FourOverSix~\cite{fouroversix}.}
Each block tries the scales $\mathrm{e4m3}(\max|x|/6)$ and
$\mathrm{e4m3}(\max|x|/4)$ and keeps the one with the lower squared error of
the block, taking $\max/4$ only when it is strictly lower. Activations are
quantized the same way, with one FP32 tensor scale per token, so a token's
quantization never depends on other tokens and every forward pass is causal.
This is also \method{}'s E2M1 candidate and activation quantizer:
FourOverSix is \method{}'s all-E2M1 map.

\paragraph{GPTQ~\cite{gptq}.}
GPTQ rounds the weights of a layer column by column and compensates the error
of each rounded column on the columns not yet rounded. Our implementation
follows the GPTQ of NVIDIA's ModelOpt~0.47.0: damping
$0.01 \cdot \mathrm{mean}(\mathrm{diag}\,H)$, lazy updates in blocks of 128
columns, ModelOpt's handling of dead columns, and the identity in place of a
Hessian that is not positive definite. Columns are processed in their natural
order, without act-order. The quantization grid is fixed beforehand by
round-to-nearest: the block and tensor scales are computed from the original
weights, and GPTQ chooses only the code of each element. GPTQ quantizes the
weights only: during calibration the activations stay in BF16, and the layers
are processed in order, each seeing the outputs of the already quantized
layers, so the codes do not depend on the activation quantizer used at
evaluation. Calibration uses our calibration set (Appendix~\ref{app:calib}),
1{,}024 sequences of 512 tokens, in batches of 8 sequences (1 for Qwen3.8-27B).
The GPTQ baseline rows of Tables~\ref{tab:main-ppl} and~\ref{tab:main-acc}
apply GPTQ to standard NVFP4 weights and run them natively with per-token
FourOverSix activations.
Table~\ref{tab:ptq} (Appendix~\ref{app:eval:ptq}) applies GPTQ with the
activations quantized during calibration to the FourOverSix grid and to the
candidates of a \method{} map.

\paragraph{IF4~\cite{if4}.}
IF4 chooses, for each 16-element block, between E2M1 and a uniform INT4 grid
under one shared block scale $s = \mathrm{e4m3}(\max|x|/(6\alpha))$, without
a clamp, so a block whose scale underflows becomes zero. With
$x_b = x/(\alpha s)$, the E2M1 candidate rounds $x_b$ to the nearest E2M1
value, ties to even, and the INT4 candidate is
$\mathrm{round}(\mathrm{clamp}(1.16666666\,x_b, -7, 7))$, dequantized with the
factor $0.8571428571$; these are the truncated constants $7/6$ and $6/7$ of
the official code. The block takes the candidate with the lower squared error
in the original units, and keeps E2M1 on a tie. Activations follow the same
rule per 16-element block on every forward pass. We use our own
implementation of IF4, written operation by operation after the authors'
reference code and bit-exact with it: on synthetic tensors and on 11 weight
matrices of Llama-3.1-8B, the choices, codes, scale bytes, and dequantized
values are identical bit for bit.
The one difference is the activation tensor scale, one per token in our runs
and one per tensor in the reference code; we use the per-token scale of every
other method, so that all forward passes are causal.

\paragraph{MixFP4~\cite{mixfp4arxiv}.}
No code has been released for MixFP4, so we implement Algorithm~1 of the
paper. The tensor scale is $\mathrm{amax}/2688$. Each 16-element block has
two candidates with their own scales: E2M1 with scale
$\mathrm{e4m3}(\max|x|/6)$, rounded to nearest, ties to even, and E1M2 with
scale $\mathrm{e4m3}(\max|x|/7)$, whose values, after the scale, are the
integers $-7, \dots, 7$. The candidate with the lower squared error wins, and
a tie goes to E1M2. Activations follow the same rule per block on every
forward pass, with a per-token tensor scale. The E1M2 candidate equals the
E0M3 candidate of \method{} element for element, so MixFP4 and \method{} choose
between the same uniform grid and an E2M1 grid; they differ in the E2M1
candidate (plain NVFP4 versus FourOverSix), in the granularity, and in the
rule. Without official code, we check our implementation in three ways: it
reproduces bit for bit an earlier implementation of ours, written separately; its E1M2
candidate equals our E0M3 quantizer on every element; and unit tests cover
the tie rule and hand-computed activation blocks. The paper does not specify
how E1M2 rounds; we assume round-to-nearest-even.

\paragraph{Simulation.}
IF4 and MixFP4 change the format every 16 elements, which no current MMA
instruction supports, so they are simulated. The weights are quantized and
dequantized in FP32 and stored in BF16; the activations are quantized and
dequantized in FP32 on every forward pass and cast to BF16; the matrix
product is a BF16 GEMM. All simulated rows, including any simulated
\method{} row, use this procedure. Appendix~\ref{app:native} compares it with
native execution.

\paragraph{Calibration data.}
Only GPTQ uses calibration data, the same set as \method{}
(Appendix~\ref{app:calib}). NVFP4, FourOverSix, IF4, and MixFP4 are
data-free: their weights are rounded to nearest, and their activations are
quantized on the fly.

\subsection{Evaluation Protocols}
\label{app:setup:eval}

\paragraph{Perplexity.}
WikiText-2~\cite{wikitext} is the raw test split, joined and tokenized once
and cut into non-overlapping 2{,}048-token windows; the last partial window
is dropped, and only the first window starts with the BOS token. C4~\cite{t5}
consists of 256 documents drawn with a fixed seed from the first English
validation shard, each longer than 2{,}049 tokens, with one random
2{,}048-token crop each.

\paragraph{Paired comparisons.}
$\Delta$NLL is the mean over windows of the per-window difference in negative
log-likelihood between two methods evaluated on identical windows (nats per
token), reported with two standard errors over windows; a difference is
significant when its magnitude exceeds two standard errors ($^{*}$ in the
tables).

\paragraph{Multiple-choice tasks.}
MMLU~\cite{mmlu} (5-shot),
PIQA~\cite{piqa}, HellaSwag~\cite{hellaswag}, WinoGrande~\cite{winogrande},
ARC-Easy and ARC-Challenge~\cite{arc}, and LAMBADA~\cite{lambada} (0-shot)
are evaluated with lm-eval~0.4.11~\cite{lmeval} and its Hugging Face
backend, by log-likelihood: a question is answered by the choice with the
highest log-likelihood, normalized by length for PIQA, HellaSwag, and ARC
and unnormalized for the other tasks, and LAMBADA counts a word as correct
when it is the greedy continuation. MMLU is the micro-average over the
14{,}042 questions of its 57 subjects.
Every task uses
lm-eval's default prompts, without a chat template, and its default handling
of the BOS token. Batches hold 16 questions (8 for Qwen3.8-27B), TF32 is off, lm-eval's
seeds are fixed, and every dataset is pinned to a revision. All methods run
natively except IF4 and MixFP4, which are simulated.

\paragraph{Generative tasks.}
GSM8K~\cite{gsm8k} (1{,}319 problems) uses lm-eval's \texttt{gsm8k\_llama}
task through the same harness: 8-shot chain of thought with its fixed
examples as earlier turns of the model's chat template, an instruction to end
with ``The final answer is'' and the number, greedy decoding with thinking
switched off (Qwen3.8-27B's \texttt{enable\_thinking} flag, Nemotron-9B's
\texttt{/no\_think} system message), at most 1{,}024 new tokens, and batches of
64 prompts (48 for Qwen3.8-27B in BF16). We report strict-match accuracy, the
number after ``The final answer is''. It runs natively, except IF4 and
MixFP4, which are simulated; Table~\ref{tab:ptq} uses the same protocol.
For Llama-1B, the date in the chat template is fixed to 15 October 2026 for
all methods. Qwen3.8-27B's simulated rows use expandable GPU memory allocation
to keep batch 64. The batch size and length-sorted batch composition are
otherwise held fixed within each model; Qwen3.8-27B BF16 is the exception at
batch 48.

\paragraph{Latency.}
Prefill is measured on prompts of $1 \times 128$ to $1 \times 8{,}192$ tokens
and $4 \times 2{,}048$, with random tokens from a fixed seed, after two
warm-up runs, over seven repetitions timed with CUDA events. Each prompt is
also captured in a CUDA graph, whose logits must equal those of eager
execution bit for bit. Decoding runs from a static KV cache in a CUDA graph at
batch sizes 1, 4, and 16 and contexts of 512 and 2{,}048 tokens, timing 64
tokens. Isolated GEMMs are timed with CUPTI over three rotated rounds of 30
calls after three warm-up calls, flushing the L2 cache with 512\,MiB between
calls. GPU clocks are not locked; the board's power limit, clocks, and driver
are recorded with every measurement.
End-to-end prefill and decoding are measured in five rounds, one process
per model, method, and round, with the order of the methods shuffled in every
round; the reported time is the median over the rounds of each process's
median.

\paragraph{Hardware and software.}
All evaluations, native and simulated, and the SM120 latency measurements run
on RTX PRO 6000 Blackwell GPUs (power limit 500\,W); the SM100 kernels and
latency measurements and the native-versus-simulated perplexity of the
$256 \times 64$ maps run on a GB200 (driver 580.105.08). The software is
Python 3.11, PyTorch 2.9.0 with CUDA 12.8 (SM120) or 13.0 (SM100),
Transformers 5.16.1, and lm-eval 0.4.11; the SM100 kernels build on
CUTLASS~4.6. The recurrent layers of the hybrid models use the kernels of
mamba-ssm 2.3.2, causal-conv1d 1.7.0, and flash-linear-attention
0.5.2. The code and the calibrated maps, with their sha256, will be
released.

\subsection{Native versus Simulated Evaluation}
\label{app:native}

\paragraph{Purpose.}
Several numbers in this paper can only be simulated: IF4 and MixFP4 choose the
element format of every 16-element block of their activations, and at
$1 \times 16$ also of their weights, which no current MMA instruction
supports, so every IF4 and MixFP4 number in this paper is simulated.
These numbers are set beside
natively executed ones. This appendix measures how far the simulation of
Appendix~\ref{app:setup:baselines} is from native execution, using the
\method{} maps, which run both ways.

\paragraph{Setup.}
For five models and the three released maps of each, we evaluate perplexity
twice on the same windows of WikiText-2 (141 to 163 windows, depending on the
tokenizer) and C4 (256 windows) (Appendix~\ref{app:setup:eval}), both times on
one RTX PRO 6000: natively, with the SM120 kernels of Section~5, which execute
a $256 \times 64$ map as 32-row granules
(Appendix~\ref{app:exec:mapping}); and simulated, with the procedure of
Appendix~\ref{app:setup:baselines}, the same weights and maps, and the same
per-token FourOverSix activations. Qwen3.8-27B is not included in this
SM120 comparison, because its simulation does not fit in the 96\,GB of one
GPU. Both runs of a model use
the same software environment; for Nemotron-9B, both use the fast kernels for
its Mamba-2 layers that its native results in Section~5 use. For every model,
tile, and corpus we compute the paired $\Delta$NLL, native minus simulated,
with two standard errors; a cell agrees when $|\Delta\mathrm{NLL}|$ is at most
two standard errors.

\begin{table}[!htbp]
  \caption{Native minus simulated evaluation of the \method{} maps of
           Table~\ref{tab:main-ppl} on the RTX PRO 6000: paired $\Delta$NLL
           ($\times 10^{-3}$ nats per token, $\pm$2\,SE). $^{\S}$:
           $|\Delta\mathrm{NLL}|$ exceeds 2\,SE (1 of 30 cells).}
  \label{tab:app-native}
  \vskip 0.1in
  \begin{center}
  \begin{footnotesize}
  \setlength{\tabcolsep}{3pt}
  \begin{tabular}{@{}llcc@{}}
    \toprule
    Model & Tile & WikiText-2 & C4 \\
    \midrule
    Llama-1B & $8{\times}64$ & $+0.06 \pm 1.41$ & $-0.94 \pm 1.22$ \\
     & $16{\times}64$ & $-0.71 \pm 1.85$ & $-0.29 \pm 1.21$ \\
     & $256{\times}64$ & $+0.15 \pm 1.76$ & $+0.17 \pm 1.19$ \\
    \midrule
    Llama-8B & $8{\times}64$ & $-0.81 \pm 1.29$ & $-0.87 \pm 1.24$ \\
     & $16{\times}64$ & $+0.17 \pm 1.28$ & $-0.75 \pm 1.08$ \\
     & $256{\times}64$ & $+0.44 \pm 1.40$ & $+0.21 \pm 1.33$ \\
    \midrule
    Mistral-7B & $8{\times}64$ & $-0.20 \pm 0.95$ & $-0.03 \pm 0.96$ \\
     & $16{\times}64$ & $-0.16 \pm 0.98$ & $-0.17 \pm 0.71$ \\
     & $256{\times}64$ & $+0.39 \pm 0.92$ & $-0.28 \pm 0.70$ \\
    \midrule
    Nemotron-9B & $8{\times}64$ & $-1.04 \pm 1.03$$^{\S}$ & $+0.36 \pm 0.67$ \\
     & $16{\times}64$ & $+0.29 \pm 0.90$ & $+0.36 \pm 0.66$ \\
     & $256{\times}64$ & $-0.45 \pm 1.08$ & $+0.63 \pm 0.66$ \\
    \midrule
    Phi-4 & $8{\times}64$ & $+0.10 \pm 1.16$ & $+0.12 \pm 0.65$ \\
     & $16{\times}64$ & $+0.67 \pm 1.13$ & $+0.11 \pm 0.64$ \\
     & $256{\times}64$ & $-0.37 \pm 1.13$ & $+0.08 \pm 0.75$ \\
    \bottomrule
  \end{tabular}
  \end{footnotesize}
  \end{center}
  \vskip -0.1in
\end{table}

\paragraph{Result.}
Native and simulated evaluation agree in 29 of the 30 cells
(Table~\ref{tab:app-native}). The largest difference is
$1.0 \times 10^{-3}$ nats per token (Nemotron-9B, $8 \times 64$, WikiText-2);
since perplexity is the exponential of the mean NLL, native and simulated
perplexities differ by at most 0.10\% in every cell. The one cell outside
two standard errors is at the threshold ($|z| = 2.0$); with 30 cells, about
1.4 cells are expected beyond two standard errors by chance. The two paths compute the same
quantized values and differ only in rounding: the native kernel decodes the
4-bit codes in the tensor core and accumulates in FP32, whereas the simulation
stores the dequantized weights in BF16 and multiplies in BF16. These
differences are small per layer and compound through the network.

\paragraph{SM100.}
On a GB200, the released $256 \times 64$ maps of all six models are evaluated
natively and in simulation on the same WikiText-2 and C4 windows. Native and
simulated evaluation agree within two standard errors in 11 of the 12 cells.
The exception, Phi-4 on WikiText-2, differs by $1.5 \times 10^{-3}$ nats per
token against two standard errors of $1.2 \times 10^{-3}$, exceeding the bound
by $0.3 \times 10^{-3}$; no cell differs by more than $1.6 \times 10^{-3}$.

\clearpage
\section{Fine-to-Coarse Format Selection}
\label{app:coarse}
This appendix gives the protocol of the experiment of
Section~\ref{sec:background:gap} and its results on all six models.

\subsection{Protocol}
\label{app:coarse:protocol}
The rules are IF4 and MixFP4 as defined in Appendix~\ref{app:setup:baselines},
each with its own weight and activation formats. At $1 \times 16$, a rule
chooses the format of every 16-element block of the weights. To apply it at a
hardware tile of $r \times 64$ weights of the matrix $[\text{out}, \text{in}]$
($8 \times 64$, $16 \times 64$, or $256 \times 64$;
Section~\ref{sec:background:gap}), we sum the
squared reconstruction error of each candidate over all elements of the tile,
give the whole tile the candidate with the smaller sum, and break ties with
the rule's own tie rule; every 16-element block keeps the block scale of the
chosen candidate. Coarsening changes only the weights: the activations are
still quantized by the rule's own per-block choice at $1 \times 16$, with one
tensor scale per token, so the coarsened rules are simulated as well
(Appendix~\ref{app:native}).

For each model, the rules at every granularity, \method{}, and the references
are evaluated on identical windows of WikiText-2 and C4
(Appendix~\ref{app:setup:eval}), and paired differences are taken over these
windows.

\subsection{Results on All Models}
\label{app:coarse:results}

\begin{table*}[p]
  \caption{Perplexity of the per-block rules at $1 \times 16$ and at hardware
           tiles, and of \method{} at the same tiles (lower is better).}
  \label{tab:app-coarse-ppl}
  \vskip 0.1in
  \begin{center}
  \begin{small}
  \setlength{\tabcolsep}{5pt}
  \newcommand{\hc}[1]{\makebox[36pt]{#1}}
  \begin{tabular}{@{}llcccccccc@{}}
    \toprule
     & & \multicolumn{4}{c}{WikiText-2} & \multicolumn{4}{c}{C4} \\
    \cmidrule(lr){3-6} \cmidrule(l){7-10}
    Model & Method & \hc{$1{\times}16$} & \hc{$8{\times}64$} & \hc{$16{\times}64$} & \hc{$256{\times}64$} & \hc{$1{\times}16$} & \hc{$8{\times}64$} & \hc{$16{\times}64$} & \hc{$256{\times}64$} \\
    \midrule
    Llama-1B & IF4 & 15.34 & 15.97 & 16.03 & 16.01 & 21.07 & 22.35 & 22.24 & 22.39 \\
     & MixFP4 & 15.25 & 15.93 & 16.07 & 16.05 & 21.05 & 22.20 & 22.21 & 22.26 \\
     & \method{} & -- & 14.40 & 14.39 & 14.70 & -- & 19.57 & 19.61 & 20.04 \\
    \cmidrule(l){2-10}
     & Reference & \multicolumn{4}{c}{\footnotesize BF16 13.16, FourOverSix 15.32, NVFP4 15.54}
                 & \multicolumn{4}{c}{\footnotesize BF16 18.52, FourOverSix 21.45, NVFP4 21.77} \\
    \midrule
    Llama-8B & IF4 & 7.74 & 7.82 & 7.82 & 7.84 & 11.18 & 11.28 & 11.30 & 11.31 \\
     & MixFP4 & 7.73 & 7.82 & 7.80 & 7.81 & 11.16 & 11.28 & 11.31 & 11.29 \\
     & \method{} & -- & 7.65 & 7.65 & 7.71 & -- & 11.05 & 11.04 & 11.12 \\
    \cmidrule(l){2-10}
     & Reference & \multicolumn{4}{c}{\footnotesize BF16 7.22, FourOverSix 7.82, NVFP4 7.89}
                 & \multicolumn{4}{c}{\footnotesize BF16 10.39, FourOverSix 11.27, NVFP4 11.34} \\
    \midrule
    Mistral-7B & IF4 & 5.67 & 5.70 & 5.70 & 5.69 & 8.36 & 8.38 & 8.38 & 8.38 \\
     & MixFP4 & 5.66 & 5.69 & 5.69 & 5.69 & 8.37 & 8.38 & 8.39 & 8.39 \\
     & \method{} & -- & 5.63 & 5.64 & 5.64 & -- & 8.30 & 8.30 & 8.31 \\
    \cmidrule(l){2-10}
     & Reference & \multicolumn{4}{c}{\footnotesize BF16 5.50, FourOverSix 5.71, NVFP4 5.73}
                 & \multicolumn{4}{c}{\footnotesize BF16 8.14, FourOverSix 8.40, NVFP4 8.42} \\
    \midrule
    Nemotron-9B & IF4 & 8.39 & 8.41 & 8.40 & 8.43 & 11.45 & 11.50 & 11.49 & 11.51 \\
     & MixFP4 & 8.41 & 8.44 & 8.41 & 8.44 & 11.46 & 11.50 & 11.50 & 11.52 \\
     & \method{} & -- & 8.32 & 8.33 & 8.35 & -- & 11.44 & 11.44 & 11.46 \\
    \cmidrule(l){2-10}
     & Reference & \multicolumn{4}{c}{\footnotesize BF16 8.09, FourOverSix 8.43, NVFP4 8.45}
                 & \multicolumn{4}{c}{\footnotesize BF16 11.15, FourOverSix 11.50, NVFP4 11.54} \\
    \midrule
    Phi-4 & IF4 & 6.63 & 6.66 & 6.66 & 6.67 & 10.51 & 10.55 & 10.55 & 10.55 \\
     & MixFP4 & 6.64 & 6.67 & 6.68 & 6.68 & 10.51 & 10.55 & 10.55 & 10.55 \\
     & \method{} & -- & 6.61 & 6.62 & 6.64 & -- & 10.49 & 10.50 & 10.52 \\
    \cmidrule(l){2-10}
     & Reference & \multicolumn{4}{c}{\footnotesize BF16 6.46, FourOverSix 6.66, NVFP4 6.69}
                 & \multicolumn{4}{c}{\footnotesize BF16 10.31, FourOverSix 10.54, NVFP4 10.58} \\
    \midrule
    Qwen3.8-27B & IF4 & 7.33 & 7.29 & 7.32 & 7.37 & 10.17 & 10.20 & 10.21 & 10.21 \\
     & MixFP4 & 7.35 & 7.15 & 7.16 & 7.20 & 10.16 & 10.20 & 10.20 & 10.21 \\
     & \method{} & -- & 7.05 & 7.08 & 7.17 & -- & 10.11 & 10.12 & 10.15 \\
    \cmidrule(l){2-10}
     & Reference & \multicolumn{4}{c}{\footnotesize BF16 7.05, FourOverSix 7.30, NVFP4 7.57}
                 & \multicolumn{4}{c}{\footnotesize BF16 9.89, FourOverSix 10.19, NVFP4 10.22} \\
    \bottomrule
  \end{tabular}
  \end{small}
  \end{center}
  \vskip -0.1in
\end{table*}

\paragraph{\method{} against the rules.}
We compare perplexities computed before rounding. \method{} has a lower
perplexity than both rules at the same tile in all 72
model--rule--tile--corpus cells (Table~\ref{tab:app-coarse-ppl}). At $8 \times 64$
and $16 \times 64$, \method{} also has a lower perplexity than both rules at
$1 \times 16$, the per-block form that needs new hardware, in all 48
cells. These
comparisons set native \method{} results beside simulated rules and are not
paired tests.

\paragraph{The rules lose their gain at the tiles.}
We measure a rule's gain at a granularity as its paired
NLL reduction over FourOverSix, and the retained gain at a tile as the ratio
of its gain there to its gain at $1 \times 16$. On Mistral-7B, both rules are
significantly better than FourOverSix at $1 \times 16$ (paired $\Delta$NLL of
$-6.5$ and $-7.7 \times 10^{-3}$ nats per token on WikiText-2, $-4.1 \times
10^{-3}$ on C4), and at every tile both are significantly worse than at
$1 \times 16$ (by $4.1$ to $5.1 \times 10^{-3}$ on WikiText-2 and $1.5$ to
$2.3 \times 10^{-3}$ on C4). On WikiText-2 they retain 23--46\% of their
$1 \times 16$ gain. Llama-3.1-8B, Nemotron-9B, and Phi-4 behave the
same way: both rules are significantly better than FourOverSix at
$1 \times 16$ on both corpora, and worse at every tile, significantly in 34 of
the 36 cells (the exceptions are Nemotron-9B at $16 \times 64$ on
WikiText-2), and so does Qwen3.8-27B on C4, significantly in all six cells.
On Llama-3.1-8B, the rules keep at most 12\% of their $1 \times 16$ gain at
any tile. On Llama-3.2-1B, the rules at $1 \times 16$ do not differ
significantly from FourOverSix on WikiText-2 and are significantly better on
C4; at every tile, both are significantly worse than at $1 \times 16$ (by
$41$ to $61 \times 10^{-3}$) and than FourOverSix. On Qwen3.8-27B's
WikiText-2, both rules at $1 \times 16$ are significantly
worse than FourOverSix, so they have no gain to lose; coarsening lowers their
NLL in five of the six cells. \method{} has a lower perplexity than the
rules at every tile on this model as well.

\subsection{Mixed Preferences Inside Tiles}
\label{app:coarse:why}
Section~\ref{sec:background:gap} states that the per-block preferences of the rules are mixed
inside the hardware tiles. To measure this, we take each rule's
$1 \times 16$ choices and group them by tile. Table~\ref{tab:app-coarse-mixing}
gives the share of tiles whose 16-element blocks choose both formats, and the
mean share of the less frequent format among a tile's blocks.

\begin{table*}[p]
  \caption{Mixing of each rule's $1 \times 16$ choices inside the hardware
           tiles. Mixed tiles: the share (\%) of tiles whose blocks choose both
           formats, truncated to four decimals. Minority share: the mean share (\%)
           of the less frequent format among the blocks of a tile.}
  \label{tab:app-coarse-mixing}
  \vskip 0.1in
  \begin{center}
  \begin{small}
  \begin{tabular}{@{}llrrrrrr@{}}
    \toprule
     & & \multicolumn{3}{c}{Mixed tiles (\%)} & \multicolumn{3}{c}{Minority share (\%)} \\
    \cmidrule(lr){3-5} \cmidrule(l){6-8}
    Model & Rule & $8{\times}64$ & $16{\times}64$ & $256{\times}64$ & $8{\times}64$ & $16{\times}64$ & $256{\times}64$ \\
    \midrule
    Llama-1B & IF4 & 100.0000 & 100.0000 & 100.0000 & 37.9 & 38.5 & 38.7 \\
     & MixFP4 & 100.0000 & 100.0000 & 100.0000 & 37.8 & 38.4 & 38.6 \\
    \midrule
    Llama-8B & IF4 & 99.9999 & 100.0000 & 100.0000 & 37.3 & 37.8 & 38.0 \\
     & MixFP4 & 99.9999 & 100.0000 & 100.0000 & 37.2 & 37.7 & 37.8 \\
    \midrule
    Mistral-7B & IF4 & 99.9996 & 99.9999 & 100.0000 & 37.4 & 38.0 & 38.1 \\
     & MixFP4 & 99.9996 & 99.9999 & 100.0000 & 37.3 & 37.8 & 38.0 \\
    \midrule
    Nemotron-9B & IF4 & 99.9999 & 100.0000 & 100.0000 & 37.9 & 38.5 & 38.8 \\
     & MixFP4 & 99.9985 & 99.9999 & 100.0000 & 36.9 & 37.5 & 37.7 \\
    \midrule
    Phi-4 & IF4 & 99.9999 & 100.0000 & 100.0000 & 36.6 & 37.1 & 37.2 \\
     & MixFP4 & 99.9999 & 100.0000 & 100.0000 & 36.5 & 36.9 & 37.0 \\
    \midrule
    Qwen3.8-27B & IF4 & 99.9999 & 100.0000 & 100.0000 & 37.4 & 38.0 & 38.1 \\
     & MixFP4 & 99.9999 & 100.0000 & 100.0000 & 37.3 & 37.8 & 38.0 \\
    \bottomrule
  \end{tabular}
  \end{small}
  \end{center}
  \vskip -0.1in
\end{table*}

On every model, more than 99.998\% of the $8 \times 64$ tiles, more than
99.9999\% of the $16 \times 64$ tiles, and all $256 \times 64$ tiles contain
both formats (more than 99.999\% at every tile on Mistral-7B). The less
frequent format holds 36.5--38.8\% of a tile's blocks (37--38\% on Mistral-7B). Any single format
for a tile therefore overrides, on average, the preference of more than a
third of its blocks.

\clearpage
\section{Calibration}
\label{app:calibration}
This appendix describes how the maps of the paper are calibrated
(Appendix~\ref{app:calib}), what one calibration costs (Appendix~\ref{app:cost:calib}),
how many tiles they flip to E0M3 (Appendix~\ref{app:eval:share}), and how the
result depends on the choices of the calibration
(Appendix~\ref{app:cost:robust}).

\subsection{Calibration Settings}
\label{app:calib}
Table~\ref{tab:app-calib-settings} lists the settings of every \method{} map in
the paper. All six models use the same settings; only the micro-batch differs
where eight sequences do not fit in the memory of one GPU. The calibration
sequences share no 13-gram (Mistral-7B tokenizer) with the WikiText-2 test
split, the C4 evaluation windows, or GSM8K.

\begin{table}[!b]
  \caption{Calibration settings of every \method{} map in the paper.}
  \label{tab:app-calib-settings}
  \vskip 0.1in
  \begin{center}
  \begin{footnotesize}
  \setlength{\tabcolsep}{3pt}
  \begin{tabular}{@{}p{0.27\columnwidth}p{0.69\columnwidth}@{}}
    \toprule
    Setting & Value \\
    \midrule
    Candidates & FourOverSix E2M1 and E0M3, as defined in Section~\ref{sec:method:formulation} \\
    Activations & per-token FourOverSix; straight-through estimator (Eq.~\eqref{eq:act-ste}) \\
    Map parameters & one FP32 latent parameter per tile; forward $\mathbf{1}[\theta > 0]$, backward $\partial m / \partial \theta := 1$ (Eq.~\eqref{eq:ste}) \\
    Initialization & $\theta = -1$ (the all-E2M1 FourOverSix map) \\
    Optimizer & Adam, $\beta = (0.9, 0.999)$, $\epsilon = 10^{-12}$, no weight decay \\
    Step size & $\eta = 0.02$, constant, no warm-up or decay \\
    Batch & 8 sequences per step; micro-batch 2 with 4 accumulation steps for Nemotron-9B and Qwen3.8-27B \\
    Schedule & 5 epochs of 128 steps (640 steps); the order of the sequences is reshuffled every epoch by one random-number generator with seed 0, which sets nothing else \\
    Calibration data & 1{,}024 sequences of 512 tokens, half from OpenWebMath~\cite{openwebmath} and half from CodeParrot~\cite{codeparrot} (pinned revisions): the 128 of the original calibration set, and 896 more drawn by a fixed rule that skips the documents already drawn and those of the C4 evaluation windows \\
    Teacher & the BF16 model; at each position, its 1{,}000 most probable token IDs and log-probabilities and the total probability of the rest of the vocabulary, precomputed once and kept on the GPU (in host memory for Qwen3.8-27B) \\
    Loss & bucketed KL divergence of Eq.~\eqref{eq:objective} over these 1{,}001 outcomes, averaged over the 511 predicted positions of each sequence and then over the 8 sequences of a step \\
    Output & the hard map after the last step; no held-out data, no checkpoint selection \\
    Numerics & deterministic kernels; TF32 off; $\tilde G$ (Eq.~\eqref{eq:weight-grad}) from BF16 operands with FP32 accumulation \\
    Recurrent layers & the Mamba-2 and Gated DeltaNet layers (Appendix~\ref{app:setup:models}) of Nemotron-9B and Qwen3.8-27B run the PyTorch fallback of Transformers during calibration and the fast kernels of mamba-ssm, causal-conv1d, and flash-linear-attention during evaluation; the Hadamard maps of Appendix~\ref{app:eval:ptq} are calibrated on the fast kernels, which are not bit-reproducible for Nemotron-9B \\
    \bottomrule
  \end{tabular}
  \end{footnotesize}
  \end{center}
  \vskip -0.1in
\end{table}

\subsection{Calibration Cost}
\label{app:cost:calib}
We measure one calibration as it would be run in practice, on one RTX PRO 6000
with deterministic kernels: from launching it and loading the model, through tokenizing the
calibration data, computing the teacher's top-1{,}000 log-probabilities and
tail mass, and packing both candidates, to training and writing the map. No development set
is evaluated. The first calibration of a model also prepares its calibration
record once (1.5 to 3.8 minutes); this one-time step is excluded.
Table~\ref{tab:app-calib-models} gives the cost at $16 \times 64$.

\begin{table*}[!t]
  \caption{Calibration cost of \method{} at $16 \times 64$ on one RTX PRO 6000.
           Setup: model load, calibration data, teacher log-probabilities, and
           candidate packing. Training: the five epochs (640 steps) together.
           Total: the whole calibration, from launch to the written map. GPU:
           peak allocated memory. Qwen3.8-27B keeps the teacher's
           log-probabilities in host memory.}
  \label{tab:app-calib-models}
  \vskip 0.1in
  \begin{center}
  \begin{small}
  \begin{tabular}{@{}lrrrr@{}}
    \toprule
    Model & Setup (min) & Training (min) & Total (min) & GPU (GiB) \\
    \midrule
    Llama-1B    &  0.9 &   2.6 &   3.7 & 16.6 \\
    Llama-8B    &  1.8 &  13.3 &  15.4 & 43.9 \\
    Mistral-7B  &  1.7 &  12.7 &  14.6 & 38.8 \\
    Nemotron-9B & 11.2 & 159.0 & 170.4 & 55.9 \\
    Phi-4       &  3.0 &  25.0 &  28.1 & 62.1 \\
    Qwen3.8-27B & 13.7 & 144.0 & 158.0 & 90.5 \\
    \bottomrule
  \end{tabular}
  \end{small}
  \end{center}
  \vskip -0.1in
\end{table*}

A calibration takes 3.7 to 170.4 minutes. Training takes 73--93\% of the
trainer's own time (Setup plus Training); the remaining approximately 0.2 minutes of
Total start the trainer and check and write the map.
The two hybrid models are slower than their sizes suggest because their
recurrent layers run the fallback implementation during calibration
(Table~\ref{tab:app-calib-settings}); Nemotron-9B, for example, needs 603\,s
for the teacher pass against 41--72\,s for Llama-8B, Mistral-7B, and Phi-4,
and 159.0 minutes of training against 12.7--25.0 minutes.
The teacher's top-1{,}000 log-probabilities and tail masses occupy 2.93\,GiB
for every model. The largest model, Qwen3.8-27B, fits on one 96\,GB GPU when
these stay in host memory.

\paragraph{Tile granularity.}
Section~\ref{sec:method:training} argues that finer tiles add only per-tile
state and reductions, so the cost of a calibration should not depend on the
tile. Table~\ref{tab:app-calib-granularity} confirms it: the three tiles differ
by a factor of 32 in the number of tiles, from $256 \times 64$ to
$8 \times 64$, but the calibration time of a model differs among them by at
most 1.9\% (Llama-1B, whose calibration takes about 210\,s) and by at most 1.5\%
for the other models, in no consistent direction. The peak GPU memory grows by at most
0.87\,GiB from $256 \times 64$ to $8 \times 64$ (Qwen3.8-27B), about 21 bytes
per extra tile: the latent parameter, its accumulated gradient, the copy
handed to the optimizer, and Adam's two moments in FP32, and one byte for the
current hard map.

\begin{table}[!htb]
  \caption{Calibration time of \method{} at the three tiles on one RTX PRO
           6000: the trainer's time, Setup plus Training of
           Table~\ref{tab:app-calib-models}, in seconds. The Total of
           Table~\ref{tab:app-calib-models} adds 11--13\,s for starting the
           trainer and checking and writing the map. Max.\ diff.: the largest
           difference among the three tiles, relative to the smallest time.}
  \label{tab:app-calib-granularity}
  \vskip 0.1in
  \begin{center}
  \begin{small}
  \setlength{\tabcolsep}{4pt}
  \begin{tabular}{@{}lrrrr@{}}
    \toprule
     & \multicolumn{3}{c}{Time (s)} & \\
    \cmidrule(lr){2-4}
    Model & $8{\times}64$ & $16{\times}64$ & $256{\times}64$ & Max.\ diff.\ (\%) \\
    \midrule
    Llama-1B    &    211 &    213 &    209 & 1.9 \\
    Llama-8B    &    901 &    910 &    915 & 1.5 \\
    Mistral-7B  &    858 &    864 &    865 & 0.8 \\
    Nemotron-9B & 10{,}163 & 10{,}211 & 10{,}217 & 0.5 \\
    Phi-4       & 1{,}661 & 1{,}677 & 1{,}655 & 1.4 \\
    Qwen3.8-27B & 9{,}476 & 9{,}465 & 9{,}416 & 0.6 \\
    \bottomrule
  \end{tabular}
  \end{small}
  \end{center}
  \vskip -0.1in
\end{table}

\subsection{Tiles Flipped to E0M3}
\label{app:eval:share}
Table~\ref{tab:app-share} gives, for each map, the percentage of its tiles that
calibration flips from E2M1 to E0M3. At $8 \times 64$ and $16 \times 64$ every
tile holds the same number of weights, so this is also the percentage of
weights. At $256 \times 64$, some matrices of Nemotron-9B and Qwen3.8-27B have
a row count that is not a multiple of 256, and their last tiles are smaller,
so there the two percentages differ slightly.

\begin{table}[!htb]
  \caption{Tiles flipped from E2M1 to E0M3 by the learned maps, in percent of
           all tiles of the quantized layers.}
  \label{tab:app-share}
  \vskip 0.1in
  \begin{center}
  \begin{small}
  \begin{tabular}{@{}lrrr@{}}
    \toprule
     & \multicolumn{3}{c}{Tiles flipped to E0M3 (\%)} \\
    \cmidrule(l){2-4}
    Model & $8{\times}64$ & $16{\times}64$ & $256{\times}64$ \\
    \midrule
    Llama-1B    & 11.28 & 14.84 & 39.69 \\
    Llama-8B    & 4.30 & 5.54 & 15.05 \\
    Mistral-7B  & 4.09 & 5.28 & 14.81 \\
    Nemotron-9B & 4.29 & 5.59 & 15.98 \\
    Phi-4       & 2.85 & 3.69 & 10.32 \\
    Qwen3.8-27B & 2.26 & 2.53 & 7.05 \\
    \bottomrule
  \end{tabular}
  \end{small}
  \end{center}
  \vskip -0.1in
\end{table}

The learned maps keep most tiles in E2M1: they flip 2.26--11.28\% of the tiles at
$8 \times 64$, 2.53--14.84\% at $16 \times 64$, and 7.05--39.69\% at
$256 \times 64$, the highest shares on the smallest model, Llama-1B. Coarser tiles flip more of the weights: from $8 \times 64$
to $256 \times 64$, the percentage grows by a factor of 3.1--3.7.
These are also the formats the hardware executes: reading the weights back
through the kernel (Appendix~\ref{app:exec:correct}) finds every weight of every deployed map
decoded in the format the map assigns.

\subsection{Calibration Robustness}
\label{app:cost:robust}
A map should not depend on incidental choices of its calibration, and its
quality should change predictably with the calibration budget. We test both
on Mistral-7B at $16 \times 64$: every run changes the settings of
Table~\ref{tab:app-calib-settings} as stated and keeps the rest. Every run uses
the environment of the paper's maps on one RTX PRO 6000, where a seed-0
recalibration reproduces the paper's map bit for bit, and every map is
evaluated natively on WikiText-2 and C4 (Appendix~\ref{app:setup:eval}). We
compare each map of Table~\ref{tab:app-robust} with the paper's map by the
paired $\Delta$NLL (the budget grid below has its own reference); FourOverSix
is $+12.8$ and $+11.4 \times 10^{-3}$ nats per token from the paper's map on
WikiText-2 and C4.

\paragraph{Incidental choices and calibration data.}
Seeds 1 and 2 change the order of the calibration sequences. Without
deterministic kernels, the order of floating-point summations on the GPU, and
hence the tile gradients, varies from run to run. Instead of math and code,
we also calibrate on 1{,}024 sequences of 512 tokens of generic web text
(FineWeb-Edu), instruction data (UltraChat), or the WikiText-2 training
split, checked against the evaluation data as the paper's calibration set is
(Appendix~\ref{app:calib}). Table~\ref{tab:app-robust} gives the results.

\begin{table}[!htb]
  \caption{Calibration choices on Mistral-7B at $16 \times 64$: paired
           $\Delta$NLL against the paper's map ($\times 10^{-3}$ nats per
           token, $\pm$2\,SE; $^{*}$: beyond 2\,SE; negative is better), and
           the percentage of tiles flipped to E0M3.}
  \label{tab:app-robust}
  \vskip 0.1in
  \begin{center}
  \begin{footnotesize}
  \setlength{\tabcolsep}{3pt}
  \begin{tabular}{@{}lccr@{}}
    \toprule
    Calibration & WikiText-2 & C4 & E0M3 (\%) \\
    \midrule
    Paper's map & 0 & 0 & 5.28 \\
    \midrule
    Seed 1 & $+0.44 \pm 0.95$ & $-0.02 \pm 0.77$ & 5.36 \\
    Seed 2 & $+0.15 \pm 0.98$ & $+0.38 \pm 0.76$ & 5.28 \\
    Non-deterministic & $-0.30 \pm 0.93$ & $-0.26 \pm 0.71$ & 5.30 \\
    \midrule
    Web text & $+0.28 \pm 1.02$ & $-0.39 \pm 0.73$ & 5.47 \\
    Instruction data & $+2.30 \pm 1.02$$^{*}$ & $+0.80 \pm 0.89$ & 5.66 \\
    WikiText-2 train & $-6.08 \pm 1.09$$^{*}$ & $+0.48 \pm 1.08$ & 5.87 \\
    \bottomrule
  \end{tabular}
  \end{footnotesize}
  \end{center}
  \vskip -0.1in
\end{table}

\begin{table*}[t]
  \caption{Calibration budget on Mistral-7B at $16 \times 64$: paired
           $\Delta$NLL against a reference map of 256 sequences and 5 epochs
           (0 / 0) on WikiText-2 / C4 ($\times 10^{-3}$ nats per token;
           $^{*}$: beyond 2\,SE, which is 0.8--1.8; negative is better).
           Bold: the paper's setting, 1{,}024 sequences for 5 epochs.
           Steps: sequences $\times$ epochs $/ 8$. $=$\,FO6: no tile is
           flipped, and the map is FourOverSix ($+9.7$ / $+9.0$ against the
           reference).}
  \label{tab:app-budget}
  \vskip 0.1in
  \begin{center}
  \begin{small}
  \begin{tabular}{@{}rccccc@{}}
    \toprule
     & \multicolumn{5}{c}{Epochs} \\
    \cmidrule(l){2-6}
    Sequences & 1 & 2 & 5 & 10 & 20 \\
    \midrule
    128 & = FO6 & = FO6 & $+2.3$$^{*}$ / $+1.6$$^{*}$ & $-0.2$ / $-0.6$ & $-2.3$$^{*}$ / $-2.9$$^{*}$ \\
    256 & = FO6 & $+5.5$$^{*}$ / $+4.0$$^{*}$ & 0 / 0 & $-2.8$$^{*}$ / $-2.5$$^{*}$ & $-2.9$$^{*}$ / $-3.7$$^{*}$ \\
    512 & $+4.9$$^{*}$ / $+4.3$$^{*}$ & $+1.4$$^{*}$ / $+0.2$ & $-2.7$$^{*}$ / $-1.7$$^{*}$ & $-3.1$$^{*}$ / $-2.8$$^{*}$ & $-2.7$$^{*}$ / $-2.7$$^{*}$ \\
    1{,}024 & $+1.0$ / $+0.5$ & $-1.7$$^{*}$ / $-2.0$$^{*}$ & $\mathbf{-3.1}$$^{*}$ / $\mathbf{-2.4}$$^{*}$ & $-3.5$$^{*}$ / $-3.8$$^{*}$ & $-3.4$$^{*}$ / $-4.6$$^{*}$ \\
    2{,}048 & $-1.3$$^{*}$ / $-1.8$$^{*}$ & $-3.2$$^{*}$ / $-2.7$$^{*}$ & $-4.5$$^{*}$ / $-3.7$$^{*}$ & $-5.7$$^{*}$ / $-4.4$$^{*}$ & $-4.0$$^{*}$ / $-4.6$$^{*}$ \\
    4{,}096 & $-3.3$$^{*}$ / $-2.6$$^{*}$ & $-3.6$$^{*}$ / $-3.8$$^{*}$ & $-3.5$$^{*}$ / $-3.6$$^{*}$ & $-5.1$$^{*}$ / $-4.3$$^{*}$ & $-5.0$$^{*}$ / $-3.8$$^{*}$ \\
    8{,}192 & $-3.0$$^{*}$ / $-2.8$$^{*}$ & $-2.9$$^{*}$ / $-2.6$$^{*}$ & $-4.9$$^{*}$ / $-4.3$$^{*}$ & $-4.0$$^{*}$ / $-2.9$$^{*}$ & $-3.3$$^{*}$ / $-3.9$$^{*}$ \\
    \bottomrule
  \end{tabular}
  \end{small}
  \end{center}
  \vskip -0.1in
\end{table*}

Every map in Table~\ref{tab:app-robust} is significantly better than
FourOverSix on both corpora. Other seeds and non-deterministic kernels give
maps whose E0M3 tiles overlap with the paper's by a Jaccard index of only
0.34--0.38, yet their $\Delta$NLL against the paper's map stays within two
standard errors on both corpora (at most $0.44 \times 10^{-3}$). The map is
thus not unique, but its quality is: many tiles have tile gradients close to
zero, and which of them cross the threshold depends on the order of the
updates. Generic web text and instruction data select almost disjoint sets of
tiles (Jaccard 0.05), yet lose little against math and code: web text is
within two standard errors on both corpora, and instruction data is worse by
$2.3 \times 10^{-3}$ on WikiText-2 only; they keep 82--103\% of its gain over
FourOverSix. The WikiText-2 training split
is better on WikiText-2 ($-6.1 \times 10^{-3}$) but not on C4, the in-domain
effect of calibrating on the evaluation corpus.

\paragraph{Calibration budget.}
We vary the number of calibration sequences (128 to 8{,}192) and of epochs (1
to 20) on a grid (Tables~\ref{tab:app-budget} and~\ref{tab:app-budget-e0m3});
a run takes $\text{sequences} \times \text{epochs} / 8$ steps, so the cells
on each anti-diagonal, such as 512 sequences for 20 epochs, 1{,}024 for 10,
and 2{,}048 for 5, have the same number of steps. Every set of sequences
contains the smaller ones; sets beyond 256 are drawn from the same two sources
by the same rule. Each map is compared with a reference map of 256 sequences
and 5 epochs (160 steps). The paper's setting, 1{,}024 sequences for 5
epochs (640 steps), is a cell of the grid, and its map matches the paper's
map in $\Delta$NLL and E0M3 share. The calibration time grows in proportion
to the steps: 8{,}192 sequences for 5 epochs take 24 times as long as the
reference and about 7 times as long as the paper's calibration.

\begin{table}[!htb]
  \caption{Tiles flipped to E0M3 (\%) by the maps of
           Table~\ref{tab:app-budget}; the paper's setting in bold, and the
           reference map is 256 sequences for 5 epochs.}
  \label{tab:app-budget-e0m3}
  \vskip 0.1in
  \begin{center}
  \begin{small}
  \setlength{\tabcolsep}{4pt}
  \begin{tabular}{@{}rrrrrr@{}}
    \toprule
     & \multicolumn{5}{c}{Epochs} \\
    \cmidrule(l){2-6}
    Sequences & 1 & 2 & 5 & 10 & 20 \\
    \midrule
    128 & 0.00 & 0.00 & 0.24 & 1.84 & 3.52 \\
    256 & 0.00 & 0.04 & 1.11 & 3.00 & 6.00 \\
    512 & 0.04 & 0.29 & 2.48 & 5.68 & 11.16 \\
    1{,}024 & 0.22 & 1.07 & \textbf{5.28} & 10.90 & 20.09 \\
    2{,}048 & 0.66 & 3.35 & 10.67 & 19.69 & 31.45 \\
    4{,}096 & 2.73 & 8.05 & 19.85 & 31.23 & 40.78 \\
    8{,}192 & 8.52 & 16.69 & 31.15 & 40.55 & 46.03 \\
    \bottomrule
  \end{tabular}
  \end{small}
  \end{center}
  \vskip -0.1in
\end{table}

The latent parameters start at $-1$ and Adam moves each by about
$\eta = 0.02$ per step, so no tile can flip before about 50 steps: with 40
steps or fewer, no tile flips and the map is FourOverSix, and with 64 and 80
steps at most 0.24\% of the tiles flip. Beyond the reference's 160 steps,
both more sequences and more epochs improve the map, by up to
$5.7 \times 10^{-3}$ on WikiText-2 and $4.6 \times 10^{-3}$ on C4
(Table~\ref{tab:app-budget}); the paper's setting reaches $3.1$ and
$2.4 \times 10^{-3}$, about half of these gains.

Three patterns hold across the grid. First, the best number of epochs falls
as the sequences grow: 128 and 256 sequences are best at 20 epochs and still
improving, 512 to 4{,}096 sequences at about 10, with 5 or 20 epochs within
two standard errors, and 8{,}192 sequences at 5. Second, at an equal number
of steps, more sequences help more than more epochs when the steps are few:
at 1{,}280 steps, 2{,}048 sequences for 5 epochs are better than 512
sequences for 20 by 1.8 and $1.1 \times 10^{-3}$ on WikiText-2 and C4. With
more steps the difference fades: at 2{,}560 steps, 2{,}048 sequences for 10
epochs are better than 4{,}096 for 5 by $2.1 \times 10^{-3}$ on WikiText-2
but not on C4, and at 5{,}120 steps the three settings agree within two
standard errors. Third, the gain saturates: every run of 1{,}280 steps or
more is 2.7 to $5.7 \times 10^{-3}$ better than the reference on
WikiText-2 and 2.6 to $4.6 \times 10^{-3}$ on C4, but these runs differ
among themselves mostly by 1 to $2 \times 10^{-3}$, several times the
seed-to-seed variation of Table~\ref{tab:app-robust} (at most
$0.44 \times 10^{-3}$), and no run beyond
5{,}120 steps is better than the best runs of 2{,}560 to 5{,}120 steps. The
best runs are better than the paper's setting by $2.6 \times 10^{-3}$ on
WikiText-2 and $2.2 \times 10^{-3}$ on C4, by the difference of their
entries in Table~\ref{tab:app-budget}.

The share of E0M3 tiles depends almost only on the number of steps
(Table~\ref{tab:app-budget-e0m3}): about 5--6\% at 640 steps, 11\% at
1{,}280, 20\% at 2{,}560, 31\% at 5{,}120, 41\% at 10{,}240, and 46\% at
20{,}480, mostly slightly more with more epochs at equal steps. Perplexity
stops improving once 10--30\% of the tiles are E0M3, while each further
E0M3 tile raises the cost at inference on SM120
(Appendix~\ref{app:exec:opt}): at 25\% and 50\% E0M3 tiles, the
$16 \times 64$ GEMMs take 1.2--7.1 and 1.9--10.7 points longer than with
none. The paper's setting, 640 steps with 5.3\% of the tiles E0M3, is
therefore a choice of calibration and inference cost close to, but short
of, saturation; the best maps of the grid take four times as many steps and
flip about 20\% of the tiles.

\clearpage
\section{Hardware-Native Execution}
\providecommand{\pend}[1]{\textcolor{red}{#1}} %
\label{app:exec}

Section~\ref{sec:execution} describes how the two Blackwell variants
execute a flip map and what it costs. This appendix gives the details: the
storage of the map (\S\ref{app:exec:mapping}) and the E0M3 encoding
(\S\ref{app:exec:e0m3}), common to both; on SM120, the dispatch per K-tile
(\S\ref{app:exec:dispatch}), the optimizations of our kernels and where their
remaining overhead lies (\S\ref{app:exec:opt}), and the GEMM time of a
forward pass (\S\ref{app:exec:gemm}); on SM100, the operand width and the
cost of the deployed maps (\S\ref{app:exec:sm100}); and how we check that native execution computes the
intended model (\S\ref{app:exec:correct}). Unless stated otherwise, SM120
measurements are taken on one RTX PRO 6000 at its 500\,W power limit,
without clock locking, and SM100 measurements on one GB200.

\subsection{Format Tags and Storage}
\label{app:exec:mapping}
\label{app:exec:packing}
\label{app:exec:platforms}

Table~\ref{tab:app-platforms} summarizes how a format tile is executed on the
two variants.

\begin{table}[!htbp]
  \caption{How a format tile is executed on the two Blackwell variants.
           Activations are E2M1 on both.}
  \label{tab:app-platforms}
  \vskip 0.1in
  \begin{center}
  \begin{footnotesize}
  \begin{tabular}{@{}p{0.19\columnwidth}p{0.37\columnwidth}p{0.31\columnwidth}@{}}
    \toprule
                     & SM120 (RTX PRO 6000, RTX 5090)                       & SM100 (B200, GB200) \\
    \midrule
    Instruction      & \texttt{mma.sync} \texttt{m16n8k64}, \texttt{mxf4nvf4} & \texttt{tcgen05.mma}, \texttt{mxf4nvf4} \\
    Format field     & opcode                                               & instruction descriptor, in a register \\
    Weight operand   & $A$ ($16{\times}64$, $256{\times}64$) or $B$ ($8{\times}64$) & $B$ \\
    Selecting E0M3   & branch per K-tile into an arm                        & register write per MMA \\
    Map storage      & bit 7 of the tile's scale bytes                      & one bit per tile \\
    Toolchain        & post-compile opcode patch                            & descriptor override \\
    \bottomrule
  \end{tabular}
  \end{footnotesize}
  \end{center}
  \vskip -0.1in
\end{table}

\paragraph{Weights keep the NVFP4 layout.}
Both formats use the NVFP4 container unchanged: two 4-bit sign-magnitude codes
per byte, one UE4M3 scale per 16 elements along $K$ in the layout the MMA
expects, and one FP32 scale per tensor. An E0M3 tile differs only in the
values written. Each 16-element block receives the scale
$s = \mathrm{E4M3}(\max|v|/7)$ of its values $v$ after division by the tensor
scale, and each element the code
$\mathrm{sign}(v) \cdot \min(7, \mathrm{round}(|v|/s))$. Every weight takes
4.5 bits, as in NVFP4.

\paragraph{The tag.}
On SM120 the tag of a tile is bit 7 of each of its UE4M3 scale bytes, which
is set exactly when the tile is E0M3. This bit is the sign bit of an E4M3
number; block scales are positive, so NVFP4 never sets it, and the tensor
core ignores it when it decodes the scale. The tag therefore takes no storage
and no bandwidth, and the kernel reads it from the scale stage that it has
already loaded into shared memory for the MMA, once per K-tile
(\S\ref{app:exec:dispatch}). Our exporter writes the tag into every scale
byte of the tile and rejects a map that is not uniform within a tile. On
SM100 the map is a separate bit array with one bit per tile: for each
256-channel output tile, $\lceil K/2048 \rceil$ 32-bit words, in which bit $j$
of word $w$ tags K-block $32w + j$. This is $1/16{,}384$ bits per weight, plus
the padding of each row to a whole word. A matrix without E0M3 tiles carries
no map.

\paragraph{Coarse and partial tiles.}
On SM120 a $256 \times 64$ map runs on the $16 \times 64$ instruction: the
kernel binds the two 16-row operand slices that a warp consumes per K-block,
64 rows apart in its 128-row block of weights, into one 32-row granule with
one tag, so a K-tile has two tags and four arms instead of 16
(\S\ref{app:exec:dispatch}). A $256 \times 64$ map is stored like a
$16 \times 64$ one, with equal tags on the 16 slices of each tile, and the
loader checks that the tags are uniform over every 128-row block. Some
output dimensions are not multiples of 256: 4{,}480, 15{,}680, and 22{,}656
in Nemotron-9B, and 48 in Qwen3.8-27B (Appendix~\ref{app:setup:models}). A
$256 \times 64$ map then has a partial last tile, which carries one format
like any other tile and is executed only over its real rows; every such
dimension is a multiple of 16.
On SM100 the kernel accepts any $N$ and any $K$ that is a multiple of 64: the
TMA zero-fills the partial last tile, and the epilogue masks its extra output
columns, so no weight is padded.

\subsection{The E0M3 Encoding}
\label{app:exec:e0m3}

E0M3 is not a documented format. PTX cannot express it, and on SM100 it is
value~0 of the operand-format field, which CUTLASS leaves unnamed. Its
semantics were established empirically by
others~\cite{qianqian2026int4,revollllt2026e0m3}: the tensor core decodes the
4-bit sign-magnitude code on the magnitudes $\{0, 1, \dots, 7\}$ instead of
the E2M1 magnitudes $\{0, 0.5, 1, 1.5, 2, 3, 4, 6\}$ and applies the same
UE4M3 block scale. Reading the weights back through the kernel (\S\ref{app:exec:correct})
confirms this decoding for every weight of every deployed map on SM120 and
SM100, and would
expose a change of the encoding in a future toolchain or GPU as a format
mismatch.

\paragraph{E0M3 costs no MMA throughput.}
On SM120, MMA loops whose instruction streams differ only in the patched
format bits run 0.4--0.6\% faster with E0M3 on either or both operands than
with E2M1 on both. On a GB200, a stock-style $8192^3$ GEMM with one format per
operand reaches 8{,}043.8--8{,}058.2\,TFLOP/s for all four combinations of
E2M1 and E0M3. What E0M3 tiles cost on SM120 is therefore not the MMA but the
arm they select (\S\ref{app:exec:opt}).

\subsection{SM120: From a Branch per MMA to a Branch per K-Tile}
\label{app:exec:dispatch}
\label{app:exec:ab}

\paragraph{Tiles, placement, and arms.}
The number of arms is $2^t$ for the $t$ tags a warp reads per K-tile. With
the weights on the $A$ operand, computing $D^{\top} = W X^{\top}$, the eight
warps form a $4 \times 2$ grid; a warp consumes two 16-row weight slices in
each of the two K-blocks, so $16 \times 64$ tiles give four tags and 16 arms,
and $256 \times 64$ tiles, whose two slices share a tag
(\S\ref{app:exec:mapping}), give two tags and four arms. An $8 \times 64$
tile requires the weights on the $B$ operand, computing $D = X W^{\top}$
instead; switching between the two computations swaps the operands without
moving data. In the $4 \times 2$ grid a warp would then span 64 weight
columns, eight tiles per K-block and $2^{16}$ arms; the kernel instead
arranges the warps $1 \times 8$, so that each spans 16 columns, two tiles per
K-block, and 16 arms. In this arrangement a warp reads 1.5 times the
shared-memory data of the stock arrangement, which is the largest remaining
cost of $8 \times 64$ (\S\ref{app:exec:opt}).

\paragraph{Tiles, placement, and arms.}
The number of arms is $2^t$ for the $t$ tags a warp reads per K-tile. With
the weights on the $A$ operand, the eight warps form a $4 \times 2$ grid; a
warp consumes two 16-row weight slices in each of the two K-blocks, so
$16 \times 64$ tiles give four tags and 16 arms, and $256 \times 64$ tiles,
whose two slices share a tag (\S\ref{app:exec:mapping}), give two tags and
four arms. An $8 \times 64$ tile requires the weights on the $B$ operand,
computing $D = X W^{\top}$ rather than the weights-on-$A$ computation
$D^{\top} = W X^{\top}$, which swaps the operands without moving data. In the
$4 \times 2$ grid a warp would then
span 64 weight columns, eight tiles per K-block and $2^{16}$ arms; the
kernel instead arranges the warps $1 \times 8$, so that each spans 16
columns, two tiles per K-block, and 16 arms. In this arrangement a warp reads
1.5 times the shared-memory data of the stock arrangement, which is the
largest remaining cost of $8 \times 64$ (\S\ref{app:exec:opt}).

\paragraph{One branch per K-tile.}
At each K-tile a warp reads its tags from the shared-memory scale stage,
combines them into an arm index, and branches to the arm, a copy of the
K-tile's mainloop body in which the format of every MMA is fixed. Each arm
contains the stage's shared-memory loads, its barrier, and the release of the
stage, so the compiler keeps the branch as a branch instead of
if-converting it. The kernel holds 1{,}024 MMA sites instead of the stock
kernel's 64, none predicated, and every path through a K-tile issues the same
32 MMAs per warp as the stock kernel.

\paragraph{A branch per MMA.}
Our first kernel chose the format with a branch before each MMA. The
compiler if-converts such a branch into a pair of predicated MMAs, one per
format (Figure~\ref{fig:sm120-dispatch}b), and a predicated-off MMA still
takes a tensor-pipe issue slot, so the kernel issues twice the tensor-pipe
instructions of the stock kernel. Writing the branch differently (as
warp-uniform branches, with predicates proven uniform, or around arms of up
to 32 MMAs) does not prevent the if-conversion. Only an indirect jump
through a table before every MMA keeps a real branch
(Figure~\ref{fig:sm120-dispatch}c): it issues as many tensor-pipe
instructions as the stock kernel, but each MMA pays a constant load, the
jump, and a warp synchronization, so it is no faster than the predicated
pairs.

To measure what the per-K-tile dispatch buys in our current kernels, we
built per-MMA variants of the deployed $16 \times 64$ and $8 \times 64$
kernels that differ from them only in the dispatch: a branch before every
MMA, which the compiler turns into predicated pairs
(Table~\ref{tab:app-ktile}). At $4096^3$ on the RTX PRO 6000, the per-MMA
branch makes the GEMM 55.8\% and 49.8\% slower than the stock kernel
($8 \times 64$ and $16 \times 64$), whatever the tags; the per-K-tile kernels
are 6.3\% and 1.2\% slower, and thus $1.47\times$ and $1.48\times$ faster than
the per-MMA branch. On the RTX 5090 the per-MMA branch costs 67.0\% and
61.1\%, the per-K-tile kernels 8.7\% and 2.6\%, so the per-K-tile kernels are
$1.54\times$ and $1.57\times$ faster. The penalty is below $2\times$
although the tensor-pipe instructions double, so at this shape and clock the
stock kernel is not purely bound by tensor-pipe issue. Every per-MMA
kernel's output equals that of its per-K-tile counterpart bit for bit.

\begin{table}[!htbp]
  \caption{Branch per MMA against branch per K-tile at $4096^3$, with the
           tags of a real layer. $\mu$s and TFLOP/s: GEMM time and throughput,
           isolated launches with cold weights. Iso.\ and B2B: extra time over
           the stock NVFP4 kernel, isolated and back-to-back launches. MMAs:
           tensor-pipe instructions per warp and K-tile. RTX PRO 6000: mean of
           10 repeats in the session of Table~\ref{tab:app-c2}, with its tags
           (Mistral-7B); RTX 5090: tags of a Llama-3.1-8B map calibrated on
           that GPU. With all tiles E2M1, the per-MMA overheads
           change by at most 0.2 points.}
  \label{tab:app-ktile}
  \vskip 0.1in
  \begin{center}
  \begin{footnotesize}
  \setlength{\tabcolsep}{2.5pt}
  \begin{tabular}{@{}llrrrrr@{}}
    \toprule
    Tile & Branch & $\mu$s & TFLOP/s & Iso. & B2B & MMAs \\
    \midrule
    \multicolumn{7}{@{}l}{\emph{RTX PRO 6000}} \\
    \multicolumn{2}{@{}l}{Stock NVFP4}      & 118.48 & 1160.0 & --       & --       & 32 \\
    $8{\times}64$  & per MMA     & 184.58 & 744.6  & +55.8\% & +69.9\% & 64 \\
                   & per K-tile  & 125.99 & 1090.9 & +6.3\%  & +7.8\%  & 32 \\
    $16{\times}64$ & per MMA     & 177.48 & 774.4  & +49.8\% & +64.3\% & 64 \\
                   & per K-tile  & 119.96 & 1145.7 & +1.2\%  & +3.8\%  & 32 \\
    \midrule
    \multicolumn{7}{@{}l}{\emph{RTX 5090}} \\
    \multicolumn{2}{@{}l}{Stock NVFP4}      & 124.95 & 1100.0 & --       & --       & 32 \\
    $8{\times}64$  & per MMA     & 208.66 & 658.7  & +67.0\% & +58.4\% & 64 \\
                   & per K-tile  & 135.81 & 1012.0 & +8.7\%  & +8.8\%  & 32 \\
    $16{\times}64$ & per MMA     & 201.30 & 682.8  & +61.1\% & +53.3\% & 64 \\
                   & per K-tile  & 128.22 & 1071.9 & +2.6\%  & +4.5\%  & 32 \\
    \bottomrule
  \end{tabular}
  \end{footnotesize}
  \end{center}
  \vskip -0.1in
\end{table}

\subsection{SM120 Kernel Optimizations and Overhead Breakdown}
\label{app:exec:opt}
\label{app:exec:path}

The per-K-tile kernels of \S\ref{app:exec:dispatch} still cost more than the
stock kernel. This section splits that cost into its sources, describes the
changes that reduced each source, and gives the cost of E0M3 tiles as a
function of their share.

\paragraph{Sources of the overhead.}
We split the overhead over the stock kernel at $4096^3$ into three parts,
each the ratio of two builds or tag sets that differ in one respect. The
\emph{tiles} part, a build without dispatch against the stock kernel, is the
cost of the warp arrangement and tile shapes alone. The \emph{dispatch} part,
all tiles E2M1 against the build without dispatch, is the cost of reading the
tags and branching, which a kernel pays even when no tile is E0M3. The
\emph{E0M3 tiles} part, the tags of a real layer against all tiles E2M1, is
the cost of the arms that E0M3 tiles select. Before the optimizations, a
fourth part, the \emph{tag copies}, is the cost of the instructions that
change (ii) below removes: the build with them against the same build
without them. Table~\ref{tab:app-c2} gives this split before and after the
optimizations, all against the deployed stock kernel. After them, $8 \times 64$ costs the most, and more than half of its 6.3\% is the tiles:
its $1 \times 8$ arrangement reads 1.5 times the shared-memory data of the
stock arrangement (\S\ref{app:exec:dispatch}), and we know no change that
would reduce this without changing the output. For $16 \times 64$ and
$256 \times 64$ the tiles cost almost nothing; the rest is the dispatch and,
for $16 \times 64$, the E0M3 tiles. The dispatch and E0M3 parts share the all-E2M1 measurement,
so their errors partly cancel in their sum.

\paragraph{The optimizations.}
Six changes, made one after another on our first per-K-tile kernels,
reduced this cost. Each was measured against the build just before it, and
each leaves every output bit unchanged.
Four act on the dispatch:
(i)~\emph{frequency-ordered dispatch}: the arm index is first compared with
the all-E2M1 pattern, which falls through, before the search over the other
arms; in the deployed maps, 37--91\% of the (warp, K-tile) patterns are
all-E2M1, weighted by FLOPs;
(ii)~\emph{no tag copies}: the patcher found the MMA sites through identity
byte permutes of the scale register, four instructions per K-tile and warp;
it now locates the first site by data-flow analysis, which removes these
instructions and the 0.9 and 0.6 points they cost $8 \times 64$ and
$16 \times 64$;
(iii)~\emph{a warp-uniform branch}: the arm index passes through a warp
reduction (\texttt{\_\_reduce\_or\_sync}), so the compiler emits a uniform
branch with no warp synchronization between the first and the last MMA of
the K-tile;
(iv)~\emph{pipelined tag reads} ($8 \times 64$): the tags of the next K-tile
are read while the current one computes.
Together, (i), (iii), and (iv) cut the dispatch part of $8 \times 64$ and
$16 \times 64$ to less than half, from 3.9 and 1.8 to 1.7 and 0.5 points.
(v)~\emph{A four-arm $256 \times 64$ kernel} with the 32-row granules of
\S\ref{app:exec:mapping} replaces the 16-arm $16 \times 64$ kernel that ran
$256 \times 64$ maps before, which makes $256 \times 64$ almost insensitive
to E0M3 tiles (below).
(vi)~\emph{Tiles}: narrow token tiles (16, 32, and 64 tokens) and a
$128 \times 64$ tile for the weights-on-$B$ kernel, which previously had only
a 128-token tile, a $64 \times 64$ epilogue tile, a tile-scheduler order,
and a re-tuned table of tile widths; we apply them to the stock kernel
alike. They matter little at $4096^3$ but at small $T$, where the narrow
token tiles make the $8 \times 64$ GEMMs up to 41\% faster.
Together, the six changes lower the overhead at $4096^3$ from 9.2\%,
3.9\%, and 4.5\% to 6.3\%, 1.2\%, and 1.0\% ($8 \times 64$,
$16 \times 64$, $256 \times 64$; Table~\ref{tab:app-c2}). The tile changes
of (vi) apply to the stock kernel as well; since every kernel is compared
with the deployed stock kernel, what they gain there counts in the overhead
before the optimizations.

\begin{table}[!htbp]
  \caption{Overhead over the deployed stock kernel at $4096^3$ (\%),
           split into its parts, before and after the optimizations of this
           section. Isolated launches with cold weights, as in
           Table~\ref{tab:app-ktile}: mean $\pm$ standard deviation over
           10 repeats; all kernels interleaved in one session on one RTX PRO
           6000.
           Tags of layer~0 \texttt{o\_proj} of Mistral-7B-Instruct-v0.3 from
           its maps of Table~\ref{tab:main-ppl}. The first $256 \times 64$ kernel is the 16-arm
           $16 \times 64$ kernel, whose tiles and tag copies it shares.}
  \label{tab:app-c2}
  \vskip 0.1in
  \begin{center}
  \begin{footnotesize}
  \setlength{\tabcolsep}{3pt}
  \begin{tabular}{@{}lccc@{}}
    \toprule
    Part & $8{\times}64$ & $16{\times}64$ & $256{\times}64$ \\
    \midrule
    \multicolumn{4}{@{}l}{\emph{Before optimization}} \\
    Tiles       & $+3.6 \pm 0.2$ & $+1.1 \pm 0.2$ & $+1.1 \pm 0.2$ \\
    Tag copies  & $+0.9 \pm 0.1$ & $+0.6 \pm 0.2$ & $+0.6 \pm 0.2$ \\
    Dispatch    & $+3.9 \pm 0.3$ & $+1.8 \pm 0.3$ & $+1.8 \pm 0.3$ \\
    E0M3 tiles  & $+0.6 \pm 0.6$ & $+0.3 \pm 0.4$ & $+0.9 \pm 0.4$ \\
    \cmidrule(l){2-4}
    Total       & $+9.2 \pm 0.5$ & $+3.9 \pm 0.4$ & $+4.5 \pm 0.3$ \\
    \midrule
    \multicolumn{4}{@{}l}{\emph{After optimization (deployed)}} \\
    Tiles       & $+3.6 \pm 0.2$ & $+0.1 \pm 0.4$ & $-0.0 \pm 0.4$ \\
    Tag copies  & -- & -- & -- \\
    Dispatch    & $+1.7 \pm 0.3$ & $+0.5 \pm 0.4$ & $+0.9 \pm 0.4$ \\
    E0M3 tiles  & $+0.9 \pm 0.5$ & $+0.7 \pm 0.3$ & $+0.1 \pm 0.5$ \\
    \cmidrule(l){2-4}
    Total       & $+6.3 \pm 0.4$ & $+1.2 \pm 0.3$ & $+1.0 \pm 0.5$ \\
    \bottomrule
  \end{tabular}
  \end{footnotesize}
  \end{center}
  \vskip -0.1in
\end{table}
\paragraph{The cost of E0M3 tiles.}
An E0M3 tile changes which arm a warp runs, not how many MMAs it issues. The
arms are laid out as a search tree whose all-E2M1 path falls through; an arm
deeper in the tree costs taken branches, about 1\% per taken branch per
K-tile at $4096^3$. Frequency-ordered dispatch (i) makes the common
all-E2M1 pattern the cheapest at the expense of the patterns with E0M3
tiles: at $4096^3$, an $8 \times 64$ layer with every tile E0M3 costs 6.6
points more than one with none after the optimizations, against 1.4 before
(the tags and kernels of Table~\ref{tab:app-c2}). Table~\ref{tab:app-e0m3-cost} gives the overhead as a function of the
tiles flipped to E0M3, at random positions. A map with every tile E0M3 costs
3.2--11.9 points more than one with none for $8 \times 64$ and 1.7--9.4
points for $16 \times 64$, but at most 1.3 points for the four-arm
$256 \times 64$ kernel, within the noise of about 1 point. The cost is not
proportional to the share: for $8 \times 64$ it grows fastest at low
shares, and at $T = 512$ the first 5\% of E0M3 tiles add 3.1 points and
the remaining 95\% another 8.8. The deployed maps flip 2.3--14.8\% of
the tiles at $8 \times 64$ and $16 \times 64$, and 7.1--39.7\% at
$256 \times 64$ (Appendix~\ref{app:eval:share}). At these shares, the E0M3
tiles add up to about 2--3 points for $8 \times 64$ (at $T = 512$; 5 points at
Llama-1B's 11\%), 1--1.6 points for $16 \times 64$ (3.7 at Llama-1B's 15\%),
and nothing beyond the noise for $256 \times 64$.
The maps of Mistral-7B-Instruct-v0.3 lie within about 1 point of the random
placement at their share: in 106 of 108 cells within 1 point, and the two
exceptions (\texttt{o\_proj} at $T = 2048$) are 1.1 points faster with the
real tags than with random ones.
The longest calibrations of
Appendix~\ref{app:cost:robust} flip up to 46\% of the $16 \times 64$
tiles (Table~\ref{tab:app-budget-e0m3}); at 25\% and 50\%, the E0M3 tiles
add 1.2--7.1 and 1.9--10.7 points.

\begin{table}[!htbp]
  \caption{Overhead over the stock kernel (\%) against the tiles flipped to
           E0M3 (\%), at random positions, for the deployed kernels. Mean
           over the \texttt{o\_proj} ($4096 \times 4096$),
           \texttt{gate\_proj} ($14336 \times 4096$) and \texttt{down\_proj}
           ($4096 \times 14336$) shapes of Mistral-7B-v0.3, with random
           weights; each cell is the median of 90 launches.}
  \label{tab:app-e0m3-cost}
  \vskip 0.1in
  \begin{center}
  \begin{footnotesize}
  \setlength{\tabcolsep}{2.5pt}
  \begin{tabular}{@{}lr*{8}{r}@{}}
    \toprule
    & & \multicolumn{8}{c}{Tiles flipped to E0M3 (\%)} \\
    \cmidrule(l){3-10}
    Tile & $T$ & \makebox[1.8em][r]{0} & \makebox[1.8em][r]{1} & \makebox[1.8em][r]{2} & \makebox[1.8em][r]{5} & \makebox[1.8em][r]{10} & \makebox[1.8em][r]{25} & \makebox[1.8em][r]{50} & \makebox[1.8em][r]{100} \\
    \midrule
    $8{\times}64$ & 16   & 2.1 & 2.2 & 2.4 & 3.1 & 3.9 & 4.5 & 5.7 & 5.8 \\
    & 128  & 1.9 & 1.8 & 2.0 & 2.5 & 2.9 & 3.5 & 4.1 & 5.1 \\
    & 512  & 8.6 & 9.4 & 10.4 & 11.7 & 13.4 & 17.2 & 20.7 & 20.5 \\
    & 2048 & 8.9 & 9.3 & 9.4 & 10.4 & 11.6 & 13.9 & 15.6 & 15.8 \\
    \midrule
    $16{\times}64$ & 16   & 0.7 & 1.0 & 0.8 & 1.3 & 1.9 & 1.9 & 2.6 & 2.4 \\
    & 128  & 2.8 & 3.0 & 3.1 & 3.5 & 4.0 & 4.7 & 5.0 & 5.1 \\
    & 512  & 1.1 & 1.5 & 2.0 & 2.6 & 3.2 & 8.2 & 11.8 & 10.5 \\
    & 2048 & 2.2 & 2.2 & 2.2 & 2.6 & 3.4 & 5.5 & 8.2 & 7.2 \\
    \midrule
    $256{\times}64$ & 16   & 0.5 & 0.5 & 0.3 & 0.4 & 0.5 & 0.5 & 0.6 & 0.6 \\
    & 128  & 0.8 & 0.7 & 0.7 & 0.8 & 1.0 & 1.1 & 1.3 & 0.7 \\
    & 512  & 1.7 & 1.8 & 1.9 & 2.0 & 2.0 & 2.1 & 2.3 & 1.8 \\
    & 2048 & 1.2 & 2.0 & 1.1 & 2.5 & 2.0 & 2.4 & 2.9 & 2.5 \\
    \bottomrule
  \end{tabular}
  \end{footnotesize}
  \end{center}
  \vskip -0.1in
\end{table}

\subsection{SM120: Per-Forward GEMM Time}
\label{app:exec:gemm}

\paragraph{Method.}
The end-to-end latency of Section~\ref{sec:eval:latency} mixes the GEMM with
every other operator, and the microbenchmarks of \S\ref{app:exec:dispatch}
and \S\ref{app:exec:opt} use a single shape. To measure the mixed-format
kernels on real models, we time the GEMM of every quantized linear layer at
$T$ tokens and sum over a forward pass, timing one module per projection and
multiplying by the number of modules. Each call runs on an idle GPU after a
512\,MiB read that evicts the weights from the L2 cache, on one of several
weight copies whose total exceeds four times the L2 cache, so the weights
are read from memory as in inference. The activation quantizer runs before
each call but is not timed. We record the CUPTI kernel time and report the
median of three rounds of 30 calls, with the kernels in a rotated order.
Each projection is timed with two sets of tags: those of its module with the
median share of E0M3 tiles, and those of its module with the most E0M3 tiles.
The reference is the stock NVFP4 kernel with the same tile changes
(\S\ref{app:exec:opt}) and the weights on $A$, the kernel that FourOverSix
deploys; it is also the reference of $8 \times 64$. All kernels, the stock
one included, use the same table of tile widths; shapes that the table does
not list (all of Llama-1B and Nemotron-9B) use the
same fallback widths for every kernel.

\begin{table}[!htbp]
  \caption{Overhead of the GEMMs of a forward pass of the deployed kernels
           over the stock kernel (\%), median over the six models with their
           maps of Table~\ref{tab:main-ppl}, on one RTX PRO 6000 (driver 595.71, 500\,W). Tags
           of each projection: from its module with the median E0M3 share,
           and from its module with the most E0M3 tiles in parentheses. Last
           row: the median over all (model, $T$) cells with $T \ge 128$.}
  \label{tab:app-gemm}
  \vskip 0.1in
  \begin{center}
  \begin{footnotesize}
  \setlength{\tabcolsep}{3pt}
  \begin{tabular}{@{}rccc@{}}
    \toprule
    $T$ & $8{\times}64$ & $16{\times}64$ & $256{\times}64$ \\
    \midrule
    1     & $+0.65$ ($+0.80$) & $+0.58$ ($+0.94$) & $+0.12$ ($+0.25$) \\
    4     & $+0.12$ ($+0.39$) & $+0.44$ ($+0.81$) & $+0.04$ ($+0.04$) \\
    16    & $+0.96$ ($+1.48$) & $+0.94$ ($+1.12$) & $+0.20$ ($+0.30$) \\
    32    & $+2.45$ ($+3.33$) & $+1.15$ ($+1.43$) & $+0.39$ ($+0.38$) \\
    64    & $+1.29$ ($+1.55$) & $+1.50$ ($+2.17$) & $+0.32$ ($+0.39$) \\
    128   & $+2.18$ ($+2.92$) & $+1.67$ ($+2.25$) & $+0.53$ ($+0.54$) \\
    256   & $+6.12$ ($+6.99$) & $+2.83$ ($+4.13$) & $+1.28$ ($+1.27$) \\
    512   & $+9.90$ ($+11.28$) & $+2.84$ ($+4.02$) & $+1.59$ ($+1.91$) \\
    1024  & $+6.33$ ($+7.46$) & $+1.29$ ($+2.17$) & $+1.05$ ($+1.20$) \\
    2048  & $+7.34$ ($+8.21$) & $+1.57$ ($+2.46$) & $+1.56$ ($+1.70$) \\
    4096  & $+7.61$ ($+8.58$) & $+1.80$ ($+2.81$) & $+1.45$ ($+1.70$) \\
    8192  & $+8.16$ ($+9.13$) & $+1.88$ ($+2.73$) & $+1.79$ ($+1.88$) \\
    \midrule
    $\ge 128$ & $+7.27$ ($+8.09$) & $+1.93$ ($+2.82$) & $+1.39$ ($+1.50$) \\
    \bottomrule
  \end{tabular}
  \end{footnotesize}
  \end{center}
  \vskip -0.1in
\end{table}

\paragraph{Result.}
At $T \ge 128$, the prefill regime, the GEMMs of a forward pass take 7.3\%,
1.9\%, and 1.4\% longer than with the stock kernel with $8 \times 64$,
$16 \times 64$, and $256 \times 64$ tiles (Table~\ref{tab:app-gemm}), and
8.1\%, 2.8\%, and 1.5\% longer with the tags of the modules with the most
E0M3 tiles. As in the split of Table~\ref{tab:app-c2}, $8 \times 64$ pays
for its warp arrangement, and the densest tags add the least to
$256 \times 64$, with only four arms. Below $T = 128$, in decoding and short
prompts, all three cost at most 2.5\% (3.3\% with the densest tags).
$8 \times 64$ costs the most at $T = 512$, 9.9\%, where its GEMMs fill only part of one wave of CTAs, so the
extra shared-memory reads are not hidden. The
end-to-end overheads of Section~\ref{sec:eval:latency} are smaller, because
the other operators of the model are unaffected.

\subsection{SM100 Execution}
\label{app:exec:sm100}

\paragraph{The kernel.}
Our SM100 GEMM is a CUTLASS~4.6 block-scaled NVFP4 GEMM with a
$256 \times 256 \times 256$ tile per pair of SMs and the weights as the $B$
operand, so one MMA covers one $256 \times 64$ format tile. Before issuing
each MMA, the issuing thread reads the tile's map bit and writes the
$B$-format field of the instruction descriptor. The stock kernel we compare
against is the same GEMM without a map.

\paragraph{Cost of the deployed maps.}
Measured as in \S\ref{app:exec:gemm} on a GB200, the deployed
$256 \times 64$ maps make the GEMMs of a forward pass 0.4--2.6\% slower than
the stock kernel at $T \ge 128$ (median 1.7\%) and at most 11\% slower at
$T < 128$.

\paragraph{Narrower format tiles.}
Figure~\ref{fig:overhead}b issues each 256-wide MMA as $256/N$ narrower
ones on an $8192^3$ GEMM. Two hardware rules change the kernel at narrow
widths: $N = 16$ and 32 need padded weight scales, and $N = 8$ runs on one
SM, since 2-SM MMAs need $N$ to be a multiple of 16. The width itself causes
most of the cost: within the padded kernel, $N = 16$ takes $4.09\times$ as
long as $N = 256$, and within the one-SM kernel, $N = 8$ takes $25.5\times$
as long; random and all-E2M1 maps take the same time.

\subsection{Correctness of Native Execution}
\label{app:exec:correct}
\paragraph{Reading the weights back.}
Native and simulated execution would differ silently if the kernel decoded
some tile in a format other than the one its map assigns. To rule this out,
we make the kernel return the weights as it decodes them: with a deployed
map installed, every quantized linear layer runs through the native kernel
on an identity activation, so each GEMM output is the weight matrix as the
tensor core decoded it, and we compare it element by element with the stored
weights under the map's formats. This checks the whole path from the map to
the hardware: the format tags, their placement on the tiles, and the
kernel's dispatch on them. All 181.9 billion weight elements of the 18 maps
on SM120 (six models, three tiles) and all 60.6 billion of the six
$256 \times 64$ maps on SM100 are exact, and on SM100, replacing a deployed
map with an all-E2M1 one is detected in every model.

\paragraph{Kernel checks.}
On SM120, every change of \S\ref{app:exec:opt} and the per-MMA variants of
\S\ref{app:exec:dispatch} were accepted only if their outputs equal those of
the kernel they replace bit for bit, over the projection shapes of four
models, real and synthetic tags, and $T$ from 1 to 8{,}192, and if the
logits of whole models are unchanged. On SM100, the per-tile kernel matches
a host reference that decodes every tile with its assigned codebook to a
relative error of about $1.7 \times 10^{-3}$, the BF16 rounding of the
output, at every operand width and over partial tiles; a reference that
decodes every tile with one codebook does not match.

\clearpage
\section{Downstream Accuracy}
\label{app:downstream}
Table~\ref{tab:main-acc} reports, per model, the mean accuracy over seven
multiple-choice tasks. This appendix gives the individual task accuracies
behind those means and the comparisons of Section~\ref{sec:eval:quality}.
The protocols are those of Appendix~\ref{app:setup:eval}.

\subsection{Multiple-Choice Tasks}
\label{app:downstream:mc}
Tables~\ref{tab:app-downstream-llama1b} to~\ref{tab:app-downstream-phi4}
give the seven tasks for each model and method.
For $16 \times 64$, the mean accuracy over the seven tasks changes by
$+0.24$ to $+1.53$ percentage points relative to FourOverSix. The reported
means weight the seven tasks equally (Table~\ref{tab:main-acc}).

\begin{table*}[!b]
  \caption{Multiple-choice accuracy (\%, higher is better) on Llama-1B and Llama-8B.
           PIQA, HellaSwag, and ARC (ARC-e, ARC-c): length-normalized accuracy;
           the other tasks: accuracy. MMLU 5-shot, the other tasks 0-shot
           (Appendix~\ref{app:setup:eval}). Avg.: mean over the seven tasks,
           the value of Table~\ref{tab:main-acc}. $^{\ddagger}$: GPTQ-quantized
           NVFP4 weights with FourOverSix activations; $^{\dagger}$: per-block
           format, simulated. Bold and underlined: best and second-best
           quantized result per column and model (BF16 excluded; ties share the
           mark).}
  \label{tab:app-downstream-llama1b}
  \vskip 0.1in
  \begin{center}
  \begin{footnotesize}
  \setlength{\tabcolsep}{4pt}
  \begin{tabular}{@{}llcccccccc@{}}
    \toprule
    Method & Tile & PIQA & HellaSwag & WinoGrande & ARC-e & ARC-c & LAMBADA & MMLU & Avg. \\
    \midrule
    \multicolumn{10}{c}{\emph{Llama-1B}} \\
    \midrule
    BF16 & -- & 75.14 & 61.67 & 61.48 & 63.80 & 37.97 & 59.46 & 46.12 & 57.95 \\
    \cmidrule(lr){1-10}
    NVFP4 & -- & 71.65 & 57.72 & 59.91 & 59.97 & 36.26 & 53.10 & 41.47 & 54.30 \\
    FourOverSix & -- & 71.38 & 57.74 & 58.17 & 60.19 & \underline{36.35} & 53.64 & 41.34 & 54.11 \\
    GPTQ$^{\ddagger}$ & -- & 72.09 & 57.79 & 59.67 & 60.19 & 35.58 & 53.93 & \textbf{42.37} & 54.52 \\
    IF4$^{\dagger}$ & $1{\times}16$ & 71.93 & 58.31 & \underline{60.77} & 60.52 & 36.18 & 54.86 & 41.24 & 54.83 \\
    MixFP4$^{\dagger}$ & $1{\times}16$ & \textbf{74.21} & 58.50 & 60.30 & 60.35 & 35.84 & 54.82 & 41.32 & 55.05 \\
    \method{} & $8{\times}64$ & 72.91 & \textbf{59.39} & 60.62 & 61.70 & 35.84 & \textbf{56.49} & 41.91 & 55.55 \\
    \method{} & $16{\times}64$ & \underline{73.39} & \underline{59.36} & 60.46 & \underline{62.08} & 36.26 & 56.05 & \underline{41.92} & \underline{55.65} \\
    \method{} & $256{\times}64$ & 72.31 & 58.75 & \textbf{62.51} & \textbf{62.12} & \textbf{36.86} & \underline{56.16} & 41.68 & \textbf{55.77} \\
    \midrule
    \multicolumn{10}{c}{\emph{Llama-8B}} \\
    \midrule
    BF16 & -- & 81.34 & 79.58 & 73.80 & 79.71 & 55.72 & 72.13 & 68.31 & 72.94 \\
    \cmidrule(lr){1-10}
    NVFP4 & -- & \underline{80.09} & 77.71 & 72.06 & 77.69 & 52.30 & \underline{71.22} & 64.83 & 70.84 \\
    FourOverSix & -- & \textbf{80.14} & 78.04 & 71.74 & 77.02 & 52.05 & \textbf{71.45} & \underline{65.73} & 70.88 \\
    GPTQ$^{\ddagger}$ & -- & 79.05 & 77.91 & 71.27 & 76.60 & \textbf{53.84} & \underline{71.22} & 65.44 & 70.76 \\
    IF4$^{\dagger}$ & $1{\times}16$ & 79.65 & \textbf{78.79} & 72.77 & 77.36 & \underline{53.67} & 70.85 & 65.58 & 71.24 \\
    MixFP4$^{\dagger}$ & $1{\times}16$ & 79.11 & \underline{78.61} & \underline{72.85} & 77.86 & 53.24 & 70.52 & 65.65 & 71.12 \\
    \method{} & $8{\times}64$ & 79.71 & 78.56 & 72.69 & \textbf{78.79} & 53.41 & 71.18 & 65.31 & \textbf{71.38} \\
    \method{} & $16{\times}64$ & 79.82 & 78.52 & \textbf{73.48} & \underline{78.28} & 52.56 & 71.03 & 65.72 & \underline{71.34} \\
    \method{} & $256{\times}64$ & 79.71 & 78.52 & 72.14 & 77.69 & 52.05 & 69.75 & \textbf{65.94} & 70.83 \\
    \bottomrule
  \end{tabular}
  \end{footnotesize}
  \end{center}
  \vskip -0.1in
\end{table*}

\begin{table*}[!b]
  \caption{Multiple-choice accuracy (\%, higher is better) on Mistral-7B and Nemotron-9B;
           metrics, Avg., and marks as in Table~\ref{tab:app-downstream-llama1b}.}
  \label{tab:app-downstream-mistral7b}
  \vskip 0.1in
  \begin{center}
  \begin{footnotesize}
  \setlength{\tabcolsep}{4pt}
  \begin{tabular}{@{}llcccccccc@{}}
    \toprule
    Method & Tile & PIQA & HellaSwag & WinoGrande & ARC-e & ARC-c & LAMBADA & MMLU & Avg. \\
    \midrule
    \multicolumn{10}{c}{\emph{Mistral-7B}} \\
    \midrule
    BF16 & -- & 83.51 & 83.32 & 74.59 & 81.73 & 59.90 & 73.53 & 61.84 & 74.06 \\
    \cmidrule(lr){1-10}
    NVFP4 & -- & 82.86 & 82.73 & 74.03 & \textbf{81.78} & \textbf{59.13} & 73.24 & 59.95 & 73.39 \\
    FourOverSix & -- & 82.75 & 82.86 & 74.82 & 80.64 & 58.36 & 73.12 & 59.87 & 73.20 \\
    GPTQ$^{\ddagger}$ & -- & \textbf{83.46} & 82.81 & 74.03 & \underline{81.44} & \underline{59.04} & 72.89 & 60.25 & 73.42 \\
    IF4$^{\dagger}$ & $1{\times}16$ & 83.19 & \underline{82.90} & \underline{75.37} & 80.98 & 58.11 & 72.77 & \underline{60.26} & 73.37 \\
    MixFP4$^{\dagger}$ & $1{\times}16$ & \underline{83.24} & \textbf{83.05} & 75.30 & \underline{81.44} & 58.53 & 73.04 & \textbf{60.38} & \textbf{73.57} \\
    \method{} & $8{\times}64$ & 82.86 & 82.65 & \textbf{75.53} & 80.93 & 57.51 & \underline{73.69} & 60.01 & 73.31 \\
    \method{} & $16{\times}64$ & \underline{83.24} & 82.82 & 74.98 & 80.85 & 58.45 & \textbf{73.80} & 60.13 & \underline{73.47} \\
    \method{} & $256{\times}64$ & 82.92 & 82.82 & 73.40 & 80.98 & 57.51 & 73.59 & 60.10 & 73.04 \\
    \midrule
    \multicolumn{10}{c}{\emph{Nemotron-9B}} \\
    \midrule
    BF16 & -- & 80.85 & 77.49 & 73.01 & 79.84 & 56.14 & 66.78 & 74.59 & 72.67 \\
    \cmidrule(lr){1-10}
    NVFP4 & -- & \underline{80.52} & 76.46 & 72.93 & \textbf{79.29} & \textbf{55.20} & 64.51 & 73.10 & \textbf{71.72} \\
    FourOverSix & -- & 79.98 & 76.55 & 71.51 & \underline{79.04} & 53.92 & 64.31 & 73.09 & 71.20 \\
    GPTQ$^{\ddagger}$ & -- & \textbf{80.79} & 76.56 & 71.67 & 78.66 & 53.24 & \textbf{66.00} & 72.77 & 71.39 \\
    IF4$^{\dagger}$ & $1{\times}16$ & 79.87 & 76.57 & 72.45 & 78.62 & \underline{54.27} & 63.98 & 73.31 & 71.30 \\
    MixFP4$^{\dagger}$ & $1{\times}16$ & 80.41 & \textbf{76.94} & 72.77 & 78.83 & 53.58 & 63.71 & 73.42 & 71.38 \\
    \method{} & $8{\times}64$ & 80.36 & \underline{76.85} & \underline{73.24} & 78.91 & 53.58 & \underline{64.80} & 73.29 & \underline{71.58} \\
    \method{} & $16{\times}64$ & \underline{80.52} & 76.60 & 72.77 & 78.58 & 53.33 & 64.76 & \textbf{73.54} & 71.44 \\
    \method{} & $256{\times}64$ & 80.36 & 76.72 & \textbf{73.80} & 78.49 & 53.33 & 64.74 & \underline{73.49} & 71.56 \\
    \bottomrule
  \end{tabular}
  \end{footnotesize}
  \end{center}
  \vskip -0.1in
\end{table*}

\begin{table*}[!t]
  \caption{Multiple-choice accuracy (\%, higher is better) on Phi-4 and Qwen3.8-27B;
           metrics, Avg., and marks as in Table~\ref{tab:app-downstream-llama1b}.}
  \label{tab:app-downstream-phi4}
  \vskip 0.1in
  \begin{center}
  \begin{footnotesize}
  \setlength{\tabcolsep}{4pt}
  \begin{tabular}{@{}llcccccccc@{}}
    \toprule
    Method & Tile & PIQA & HellaSwag & WinoGrande & ARC-e & ARC-c & LAMBADA & MMLU & Avg. \\
    \midrule
    \multicolumn{10}{c}{\emph{Phi-4}} \\
    \midrule
    BF16 & -- & 81.12 & 81.95 & 76.95 & 72.77 & 56.14 & 72.27 & 80.29 & 74.50 \\
    \cmidrule(lr){1-10}
    NVFP4 & -- & 80.85 & 80.86 & 73.64 & \underline{73.23} & 54.10 & 71.88 & 79.11 & 73.38 \\
    FourOverSix & -- & 80.69 & 81.11 & 74.27 & 71.89 & 54.86 & 71.55 & 79.23 & 73.37 \\
    GPTQ$^{\ddagger}$ & -- & 80.74 & \underline{81.17} & 73.95 & 72.90 & \underline{55.55} & \textbf{72.75} & 79.27 & \textbf{73.76} \\
    IF4$^{\dagger}$ & $1{\times}16$ & 81.01 & 81.15 & \textbf{74.82} & 72.52 & 55.38 & 71.49 & \textbf{79.45} & \underline{73.69} \\
    MixFP4$^{\dagger}$ & $1{\times}16$ & \underline{81.23} & 81.14 & 73.09 & \textbf{73.74} & \textbf{55.89} & 71.28 & \underline{79.39} & 73.68 \\
    \method{} & $8{\times}64$ & 80.41 & 80.99 & 74.11 & 72.05 & 55.29 & 71.74 & 79.20 & 73.40 \\
    \method{} & $16{\times}64$ & \textbf{81.39} & \textbf{81.18} & 74.43 & 72.69 & 55.20 & 71.76 & 79.18 & \underline{73.69} \\
    \method{} & $256{\times}64$ & 80.90 & 81.11 & \underline{74.74} & 71.93 & 55.03 & \underline{71.94} & 79.31 & 73.57 \\
    \midrule
    \multicolumn{10}{c}{\emph{Qwen3.8-27B}} \\
    \midrule
    BF16 & -- & 81.50 & 82.86 & 75.93 & 73.11 & 58.36 & 72.02 & 82.62 & 75.20 \\
    \cmidrule(lr){1-10}
    NVFP4 & -- & 80.90 & \underline{82.39} & 73.88 & \underline{76.22} & \underline{59.73} & 70.17 & 81.32 & 74.94 \\
    FourOverSix & -- & 81.07 & 82.35 & 75.14 & 74.12 & 58.96 & 70.56 & 81.80 & 74.86 \\
    GPTQ$^{\ddagger}$ & -- & 81.01 & 82.21 & 74.82 & 74.66 & 56.91 & 70.99 & 81.41 & 74.57 \\
    IF4$^{\dagger}$ & $1{\times}16$ & 80.85 & 82.32 & 74.74 & \textbf{78.28} & \textbf{60.75} & \underline{71.61} & 81.05 & \textbf{75.66} \\
    MixFP4$^{\dagger}$ & $1{\times}16$ & 80.63 & \textbf{82.48} & \underline{75.45} & 75.04 & 58.53 & \textbf{71.67} & 81.95 & 75.11 \\
    \method{} & $8{\times}64$ & \underline{81.18} & 82.27 & \underline{75.45} & 74.83 & 58.62 & 71.34 & 82.55 & 75.18 \\
    \method{} & $16{\times}64$ & \textbf{81.72} & 82.22 & \textbf{75.85} & 75.97 & 59.30 & 71.41 & \textbf{82.63} & \underline{75.59} \\
    \method{} & $256{\times}64$ & \underline{81.18} & 82.21 & 75.14 & 75.25 & 59.47 & 70.64 & \underline{82.60} & 75.21 \\
    \bottomrule
  \end{tabular}
  \end{footnotesize}
  \end{center}
  \vskip -0.1in
\end{table*}

\clearpage
\onecolumn\twocolumn
\section{Analysis}
\label{app:analysis}

\subsection{Does the Tile Gradient Rank Flips Correctly?}
\label{app:eval:gradcheck}
Section~\ref{sec:eval:analysis} relies on the tile gradient $g_u$ of
\S\ref{sec:method:gradient} to choose which tiles to flip. We test two claims
about it: that $g_u$ ranks tiles by the change their flip causes in the loss,
better than scores that look at one layer at a time; and that its first-order
prediction holds for few tiles but overestimates the effect of many, which is
why \method{} re-evaluates it during training instead of selecting once.

\paragraph{Setup.}
We use Mistral-7B at $16 \times 64$, which has $6.8 \times 10^6$ tiles, of
which the paper's map flips 359{,}840 (5.28\%). $g_u$ is computed in one pass
over the 1{,}024 calibration sequences at the all-E2M1 map, as in the first step
of calibration; it is the derivative of the mean bucketed KL divergence of
Eq.~\eqref{eq:objective} over these sequences, and that loss equals the
trainer's bit for bit. We then flip sets of tiles from the all-E2M1
map and measure the change in the same bucketed KL divergence from the BF16
teacher (its top-1{,}000 tokens plus one tail bucket) on two sets of
sequences of 512 tokens: \emph{held-out}, 256 sequences (128 math and 128
code) drawn by the rule of
Appendix~\ref{app:cost:robust} and disjoint from the calibration set, and
\emph{in-sample}, the 1{,}024 calibration sequences themselves, on which $g_u$ is
computed and the paper's map was calibrated. Each change is paired over the
sequences of its set and reported with 2\,SE. Below, ``KL'' refers to this bucketed
KL. The prediction for a set of tiles is the sum of their $g_u$.

\paragraph{Sets of tiles.}
A single flip among the 0.1\% of tiles with the most negative $g_u$ is
predicted to change the KL divergence by $4 \times 10^{-6}$ on average, 40 to
75 times less than the standard error of a change ($1.6 \times 10^{-4}$ over
the 1{,}024 in-sample and $3.0 \times 10^{-4}$ over the 256 held-out
sequences), so we flip groups. (a)~We
split the tiles into ten bins by the rank of $g_u$, finer at the negative
end (0--0.1, 0.1--0.3, 0.3--1, 1--3, 3--10, 10--50, 50--90, 90--97,
97--99.7, and 99.7--100\% of the tiles, from the most negative), and draw
from each bin three disjoint groups of 1{,}000 tiles, 30 groups in all,
each flipped together. On these groups we
compare $g_u$ with
two scores that use one layer only: the activation-weighted score of
Table~\ref{tab:ablation} and the weight error
$\lVert W - W^{(0)} \rVert^2 - \lVert W - W^{(1)} \rVert^2$ of the tile. (b)~We
flip the $k$ tiles with the most negative $g_u$ over the whole model, for $k$
from 10 to 359{,}840, the size of the paper's map, and, for comparison, the
paper's map itself.

\begin{figure*}[t]
  \centering
  \includegraphics[width=\textwidth]{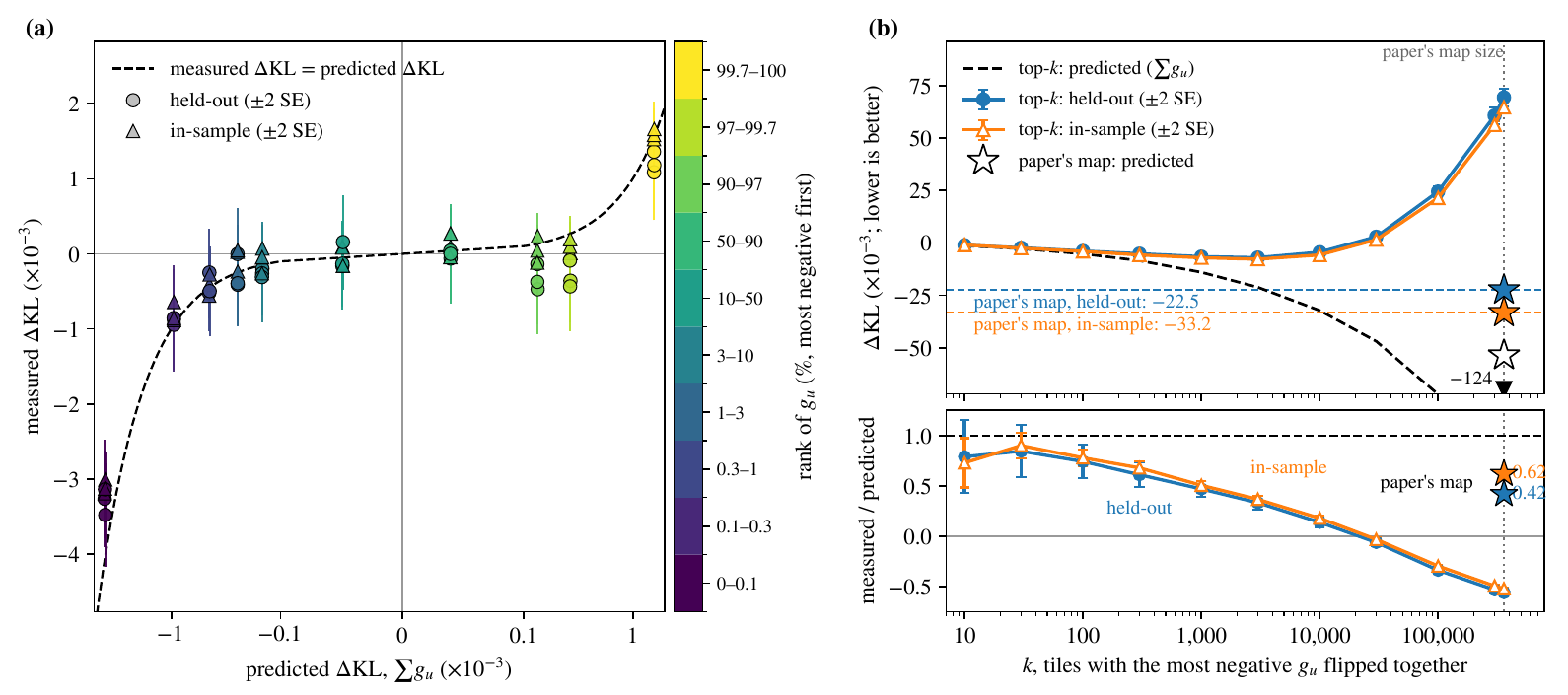}
  \caption{The tile gradient against the measured change in bucketed KL
           divergence
           on Mistral-7B at $16 \times 64$. Measured: the change in the KL
           divergence from the BF16 teacher when the tiles are flipped from
           the all-E2M1 map, paired over the sequences of each set; predicted: the sum of
           the tiles' $g_u$, its first-order estimate; negative means the KL
           divergence falls. (a)~The 30 groups of 1{,}000
           tiles: measured change ($\pm$2\,SE) against the predicted change
           $\sum g_u$, colored by the bin of $g_u$ each group is drawn from
           (dashed: measured $\Delta$KL equals predicted $\Delta$KL; the horizontal axis is linear
           between $-0.1$ and $0.1$ and logarithmic beyond). (b)~The $k$ tiles
           with the most negative $g_u$ flipped together. Top: measured
           ($\pm$2\,SE) and predicted change (dashed; the triangle marks its
           value at 359{,}840 tiles, which lies below the axis); stars: the
           paper's map, whose 359{,}840 tiles were chosen by calibration
           (dotted line: its size; horizontal dashed lines: its measured
           changes; hollow: its predicted change). Bottom: measured divided by
           predicted change (dashed: equal).}
  \label{fig:app-gradcheck}
\end{figure*}

\paragraph{Ranking.}
On the groups of (a), $g_u$ ranks them by their measured change, and the two one-layer
scores do not (Table~\ref{tab:app-gradcheck}): the Spearman correlation is
$+0.72$ on held-out and $+0.88$ on in-sample sequences for $g_u$, while the
activation-weighted score orders the groups the other way ($-0.43$ and
$-0.46$) and the weight error has intervals that include zero. Where a change is
measurable, $g_u$ predicts its sign: 10 of the 30 groups change the held-out
KL divergence by more than 2\,SE, and 11 the in-sample one, all in the
direction $g_u$ predicts and all from the bins of the 1\% most negative or
the 0.3\% most positive $g_u$. The other groups change the KL divergence by
less than 2\,SE.

\begin{table}[!hb]
  \caption{Ranking groups of tiles: Spearman correlation between each score
           and the measured decrease in bucketed KL divergence over the 30 groups of
           1{,}000 tiles (Figure~\ref{fig:app-gradcheck}a); 95\% bootstrap
           interval over groups in brackets. Act.-weighted: the
           activation-weighted score of Table~\ref{tab:ablation}.}
  \label{tab:app-gradcheck}
  \vskip 0.1in
  \begin{center}
  \begin{footnotesize}
  \setlength{\tabcolsep}{2.5pt}
  \begin{tabular}{@{}lcc@{}}
    \toprule
    Score & Held-out & In-sample \\
    \midrule
    $g_u$              & $+0.72$ [$+0.40$, $+0.90$] & $+0.88$ [$+0.69$, $+0.97$] \\
    Act.-weighted      & $-0.43$ [$-0.76$, $-0.01$] & $-0.46$ [$-0.78$, $-0.04$] \\
    Weight error       & $-0.27$ [$-0.69$, $+0.18$] & $-0.19$ [$-0.62$, $+0.27$] \\
    \bottomrule
  \end{tabular}
  \end{footnotesize}
  \end{center}
  \vskip -0.1in
\end{table}

\paragraph{Accuracy of the first-order prediction.}
For the top-$k$ sets of (b), the prediction holds approximately only for few
tiles: flipping the 10 to 100 most negative tiles changes the KL divergence by
0.73--0.90 times the prediction on the in-sample sequences, on which $g_u$ is
computed, and by 0.74--0.84 times on held-out ones
(Figure~\ref{fig:app-gradcheck}b). With more tiles the ratio falls, on
in-sample sequences to 0.68 at 300 tiles, 0.51 at 1{,}000 and 0.18 at $10^4$,
and on held-out ones to 0.61, 0.47 and 0.14. From $3 \times 10^4$ tiles on,
the flips raise the KL divergence; at 359{,}840, the size of the paper's map,
they raise it by $64.7 \pm 1.7 \times 10^{-3}$ on in-sample and
$69.4 \pm 4.3 \times 10^{-3}$ on held-out sequences, against a predicted
decrease of $124 \times 10^{-3}$. The tiles of one gradient
evaluation interact, so their first-order effects do not add up. The paper's
map flips the same number of tiles but was chosen by re-evaluating the
gradient during calibration; it lowers the KL divergence by
$22.5 \pm 1.7 \times 10^{-3}$ on held-out and $33.2 \pm 1.1 \times 10^{-3}$
on in-sample sequences, 0.42 and 0.62 times the sum of its tiles' $g_u$ at
the all-E2M1 map. The two sets of tiles differ: only 25\% of the paper's
tiles are among the 359{,}840 with the most negative $g_u$, half rank
between 10\% and 50\%, and 25{,}979 (7\%) have $g_u > 0$, which a selection
from one evaluation would never flip.
Selecting the same number of tiles from one evaluation
thus raises the KL divergence, while iterative selection keeps 0.42--0.62 of
the predicted gain; the one-shot map of Table~\ref{tab:ablation}, which selects once per
layer, is the corresponding test on perplexity.

\paragraph{Summary.}
The tile gradient ranks groups of tiles by their effect on the loss and
predicts its sign, which the one-layer scores do not; the effect is resolved
at the two ends of the range of $g_u$, the tiles that a selection flips first
or never. Its first-order prediction holds approximately (0.73--0.90 of it on
in-sample, 0.74--0.84 on held-out sequences) for up to 100 tiles and fails as
more tiles are flipped together, so a selection from one gradient evaluation
raises the KL divergence at the size of the paper's map, while re-evaluating
the gradient during calibration, as \method{} does, keeps 42\% of the
predicted gain on held-out and 62\% on in-sample sequences.

\subsection{Combining with Post-Training Quantization}
\label{app:eval:ptq}
\begin{table*}[!t]
  \caption{Combining \method{} ($16 \times 64$) with post-training
           quantization methods, native execution: WikiText-2 and C4
           perplexity (lower is better) and GSM8K accuracy (\%, higher is
           better). GPTQ compensates the quantization error with second-order
           updates, with activations quantized during calibration; its
           \method{} row trains the map on GPTQ-quantized candidates of both
           formats. Hadamard rotates weights and activations in blocks of 16
           before quantization, and its \method{} map is trained in the
           rotated basis. Format sets both the weights and the activations:
           NVFP4 rows use NVFP4 activations, the others FourOverSix
           activations. The GPTQ baseline of Table~\ref{tab:main-ppl} instead
           quantizes standard NVFP4 weights with GPTQ calibrated on BF16
           activations and uses FourOverSix activations. GSM8K follows
           lm-eval's \texttt{gsm8k\_llama} (8-shot chain of thought with the
           chat template, greedy decoding, at most 1024 new tokens, thinking
           off, strict match); its 2\,SE is 1.0--2.0 points. Bold: within each PTQ method, the lowest
           perplexity and the highest GSM8K accuracy per column; ties are all bold.}
  \label{tab:ptq}
  \vskip 0.1in
  \begin{center}
  \begin{small}
  \setlength{\tabcolsep}{6pt}
  \begin{tabular}{@{}llcccccc@{}}
    \toprule
     & & \multicolumn{3}{c}{Nemotron-9B} & \multicolumn{3}{c}{Qwen3.8-27B} \\
    \cmidrule(lr){3-5} \cmidrule(l){6-8}
    PTQ & Format & Wiki & C4 & GSM8K & Wiki & C4 & GSM8K \\
    \midrule
    \multicolumn{2}{@{}l}{BF16} & 8.09 & 11.15 & 87.4 & 7.05 & 9.89 & 95.7 \\
    \midrule
    RTN & NVFP4 & 8.45 & 11.54 & 87.3 & 7.57 & 10.22 & 95.4 \\
     & FourOverSix & 8.43 & 11.50 & \textbf{88.9} & 7.30 & 10.19 & 94.6 \\
     & \textbf{\method{}} & \textbf{8.33} & \textbf{11.44} & 88.6 & \textbf{7.08} & \textbf{10.12} & \textbf{95.9} \\
    \cmidrule(lr){1-8}
    GPTQ & NVFP4 & 8.43 & 11.50 & 84.5 & 7.44 & 10.17 & 95.9 \\
     & FourOverSix & 8.38 & 11.46 & \textbf{86.9} & 7.31 & 10.13 & \textbf{96.2} \\
     & \textbf{\method{}} & \textbf{8.32} & \textbf{11.43} & 85.0 & \textbf{7.11} & \textbf{10.10} & 95.8 \\
    \cmidrule(lr){1-8}
    Hadamard & NVFP4 & 8.52 & 11.64 & 84.9 & 7.61 & 10.34 & 95.8 \\
     & FourOverSix & 8.45 & 11.56 & \textbf{88.2} & 7.73 & 10.26 & \textbf{96.2} \\
     & \textbf{\method{}} & \textbf{8.34} & \textbf{11.46} & \textbf{88.2} & \textbf{7.21} & \textbf{10.15} & \textbf{96.2} \\
    \bottomrule
  \end{tabular}
  \end{small}
  \end{center}
  \vskip -0.1in
\end{table*}

\method{} changes only how each tile's codes are interpreted, so it should
compose with methods that change the values being quantized
(Section~\ref{sec:background}). Table~\ref{tab:ptq} combines it with GPTQ and
Hadamard rotation in the way IF4, MixFP4, and FourOverSix are combined with
such methods: the method is applied first, and the format choice is made on
its result. For GPTQ, both candidates are quantized with GPTQ and the map is
trained on them; for Hadamard rotation, the map is trained in the rotated
basis. \method{} improves both methods over their FourOverSix counterparts on
both models and corpora: by $7.1$ and $3.2 \times 10^{-3}$ nats per token
(WikiText-2, C4) on Nemotron-9B and by $27.7$ and $2.7 \times 10^{-3}$ on
Qwen3.8-27B for GPTQ, and by $14.0$ and $8.3$ and by $70.8$ and
$10.3 \times 10^{-3}$ for Hadamard rotation. Training the map on the method's
result matters: a map trained on round-to-nearest candidates and kept fixed
adds nothing to GPTQ on Nemotron-9B, and on the GPTQ candidates the retrained
map is $3.8$ and $2.1 \times 10^{-3}$ better than the fixed one. GPTQ improves
on \method{} alone only on C4, by $1.1$ and $2.0 \times 10^{-3}$, is worse on
Qwen3.8-27B's WikiText-2 ($+4.9 \times 10^{-3}$), and on Nemotron-9B loses
3.6 points of GSM8K; block-16 rotation degrades every format. The Hadamard
maps are calibrated on the fast recurrent kernels
(Table~\ref{tab:app-calib-settings}); on Nemotron-9B this training is not
reproducible bit for bit, so its row is one draw.

\end{document}